\documentclass[12pt]{article}
\usepackage{amsmath}
\usepackage{graphicx}
\usepackage{enumerate}
\usepackage{natbib}
\usepackage{url} % not crucial - just used below for the URL 
\usepackage[flushleft]{threeparttable}
\usepackage{multirow}
\usepackage{verbatim}
\newcommand{\blind}{1}

\usepackage{varwidth}
\usepackage{amsmath}
\usepackage{numprint}
\usepackage{algorithm}
\usepackage{algpseudocode}
\usepackage{multirow}

\algrenewcommand\algorithmicrequire{\textbf{Input:}}
\algrenewcommand\algorithmicensure{\textbf{Output:}}
\usepackage{amsmath,amssymb}

\usepackage{mathrsfs}
\usepackage{bm, xcolor}
\usepackage{mathrsfs}
\usepackage{verbatim}
\usepackage{longtable}
\usepackage{threeparttable}
\usepackage{color}
\usepackage{amsmath}

\usepackage{enumerate}
\usepackage{amssymb,bbm}
\usepackage{amsbsy}
\usepackage{amsmath}
\usepackage{amsthm}
\usepackage{amsfonts}
\usepackage{latexsym}
\usepackage{xifthen} % Allows us to put in optional arguments to questions
\usepackage{color}
\usepackage{manfnt}
\usepackage{ifthen}
\usepackage{bm}
\usepackage{caption}
\usepackage{subcaption}

\newcommand{\mR}{\mathbb{R}}

\newcommand{\A}{\mathcal{A}}

\newcommand{\E}{\mathbb{E}} % Expectation symbol
\providecommand{\argmin}{\mathop{\rm argmin}}

\makeatletter
\def\singlespace{\def\baselinestretch{1}\@normalsize}

\newtheorem{lemma}{Lemma}%[section]
\newtheorem{theorem}{Theorem}%[section]

\newtheorem{example}{Example}%[section]
\newtheorem{corollary}{Corollary}%[section]
\@addtoreset{equation}{section}
\newtheorem{definition}{Definition}
\newtheorem{assumption}{Assumption}
\newtheorem{proposition}{Proposition}

\renewcommand{\hat}{\widehat}

\def\singlespace{\def\baselinestretch{1}\@normalsize}

\makeatother

\def\newpage{\vfill\eject}

\newdimen\biblioindent    \biblioindent=30pt

 at 7truept

\def\beqr{\begin{eqnarray}}
	\def\eeqr{\end{eqnarray}}
\def\beqrs{\begin{eqnarray*}}
	\def\eeqrs{\end{eqnarray*}}

\def\beq{\begin{equation}}
\def\eeq{\end{equation}}
\def\beqn{\begin{eqnarray}}
\def\eeqn{\end{eqnarray}}
\def\beqnn{\begin{eqnarray*}}
\def\eeqnn{\end{eqnarray*}}

\def\A{{\bf A}}

\def\calP{\mathcal{P}}
\def\calL{\mathcal{L}}
\def\calD{\mathcal{D}}

\newcommand{\bA}{{\mathbf A}}
\newcommand{\bB}{{\mathbf B}}

\newcommand{\bE}{{\mathbf E}}
\newcommand{\bG}{{\mathbf G}}

\newcommand{\bI}{{\mathbf I}}

\newcommand{\bM}{{\mathbf M}}

\newcommand{\bS}{{\mathbf S}}

\newcommand{\bV}{{\mathbf V}}
\newcommand{\bW}{{\mathbf W}}

\newcommand{\bZ}{{\mathbf Z}}

\newcommand{\bu}{{\mathbf u}}

\newcommand{\bx}{{\mathbf x}}

\newcommand{\bz}{{\mathbf z}}

\newcommand{\bSigma}{\boldsymbol{\Sigma}}
\newcommand{\bDelta}{\boldsymbol{\Delta}}

\newcommand{\bphi} {\boldsymbol{\phi}}

\newcommand{\btheta} {\boldsymbol{\theta}}
\newcommand{\bxi} {\boldsymbol{\xi}}

\theoremstyle{definition}
\newtheorem{defi}{Definition}%[section]
\makeatletter

\newcommand{\Rmnum}[1]{\expandafter\@slowromancap\romannumeral #1@}
\makeatother

\usepackage{chngcntr}
\usepackage{apptools}
\AtAppendix{\counterwithin{lemma}{section}}

\begin{document}

	\def\spacingset#1{\renewcommand{\baselinestretch}%
		{#1}\small\normalsize} \spacingset{1}
	
	\if1\blind
	{
		\title{\bf    Statistical Inference for Differentially Private Stochastic Gradient Descent}
		\author{Xintao Xia\\
			Center for Data Science, Zhejiang University\\
			and \\
			Linjun Zhang\\
			Department of Statistics, Rutgers University\\
			and \\
			Zhanrui Cai\\
			Faculty of Business and Economics, The University of Hong Kong}
		\date{}
		\maketitle
	} \fi
	
	\if0\blind
	{
		\bigskip
		\bigskip
		\bigskip
		\begin{center}
			{\LARGE\bf Statistical Inference for Differentially Private Stochastic Gradient Descent}
		\end{center}
		\medskip
	} \fi
	
	\bigskip
	
	\begin{abstract}
		Privacy preservation in machine learning, particularly through Differentially Private Stochastic Gradient Descent (DP-SGD), is crucial for analyzing sensitive data. However, existing statistical inference methods for SGD predominantly focus on cyclic subsampling, while DP-SGD requires randomized subsampling. This paper first bridges this gap by establishing the asymptotic properties of SGD under the randomized rule and extending these results to DP-SGD. For DP-SGD, we show that the asymptotic variance decomposes into statistical, sampling, and privacy-induced components. We propose two methods for constructing valid confidence intervals: the plug-in method and the random scaling method. We also perform extensive numerical analysis showing that the proposed confidence intervals achieve nominal coverage rates while maintaining privacy. 
	\end{abstract}
	
	\noindent%
	\vfill
	
	\newpage
	\spacingset{1} % Single-spaced arXiv version.
	
	\section{Introduction}
	
	{Stochastic Gradient Descent} (SGD) is one of the most widely used optimization algorithms in modern statistics and machine learning due to its computational efficiency and low memory requirements \citep{goodfellow2016deep}. Consider a real-valued {convex} objective function $\calL(\btheta) := \E_Z l(\bZ;\btheta)$, where $l(\bZ;\btheta)$ is a convex loss function and $\bZ$ denotes one observation drawn from a probability distribution $\calP$. Let $\btheta^*\in\mR^p$ be the unique minimizer of $\calL(\btheta)$, $ \btheta^* = \argmin_{\theta} \calL(\btheta).$ In practice, we observe independent and identically distributed (i.i.d.) copies from $\calP$, denoted by $\calD = \{\bz_1, \bz_2,\dots, \bz_n\}$, and estimate the parameter by minimizing the empirical risk function: $n^{-1}\sum_{i=1}^{n} l(\bz_i,\btheta)$. Starting with an initial value, the SGD algorithm iteratively updates the parameter using the following rule:
	\begin{equation}\label{eq: update}
		\btheta^{(t)} = \btheta^{(t-1)} - \eta_t\cdot \frac{1}{|I_t|} \sum_{i\in I_t} \nabla l(\bz_i;\btheta^{(t-1)} ), 
	\end{equation}
	where $\eta_t$ is the learning rate and $I_t\subset\{1, 2, \dots, n\}$ is a subset of indices selected at iteration $t$. There are two typical rules to select $I_t$: the {\it randomized rule} and the {\it cyclic rule}. The {\it randomized rule} selects $I_t\subset\{1, 2, \dots, n\}$ uniformly at random in each iteration $t$, and is more common in practice. The {\it cyclic rule} selects $I_t$ by sequentially iterating through the index $\{1, 2, \dots, n\}$, and is particularly prevalent in online learning settings. Throughout this paper, we refer to the two variants as randomized SGD and cyclic SGD, respectively. 
	
	The \textit{averaged stochastic gradient descent} (ASGD) outputs the averaged iterates $\bar\btheta_T := T^{-1}\sum_{i=1}^{T}\btheta^{(i)}$. It was shown that the averaged iterates are asymptotically normally distributed with the full sample asymptotic variance \citep{ruppert1988efficient, polyak1992acceleration}. Specifically, let $A := \nabla^2 \calL(\btheta^*)$ be the Hessian matrix of the objective function evaluated at $\btheta^*$, and let $S := \E\{ \nabla l(\bZ;\btheta^*) \nabla l(\bZ;\btheta^*)^{\top}\}$ be the covariance matrix of the score function $\nabla l(\bZ;\btheta^*)$. Then, under mild conditions, $\sqrt{T}(\bar\btheta_T  - \btheta^*) \overset{d}{\to} N(0, A^{-1}SA^{-1}),$ where $\overset{d}{\to}$ stands for convergence in distribution. Recent developments in the statistics literature aim to understand the distributions of the estimates and provide valid inference tools in practice. Based on the sequential data, \cite{chen2020statistical} proposed the plug-in estimator and the batch-means estimator to estimate the asymptotic covariance of the average iterates, enabling the construction of corresponding confidence intervals. \cite{lee2022fast} constructed asymptotically pivotal statistics via random scaling based on tools from the functional central limit theorem. The method applies to online sequential data and can be implemented efficiently. The distribution of SGD was also studied in various estimation settings, including online learning \citep{su2023higrad}, contextual bandits \citep{chen2022online}, and debiased inference \citep{han2024online}. Nevertheless, the literature has yet to address how this framework might be adapted to a randomized stochastic gradient descent algorithm. In particular, the existing literature has exclusively focused on analyzing the distributional properties of cyclic SGD. The statistical inference framework for randomized SGD, a critical requirement for differentially private algorithms, remains unexplored. Theoretically, cyclic SGD benefits from processing independent data subsets in successive iterations, while randomized SGD encounters dependent data structures due to its stochastic sampling mechanism. This dependence adds significant complexity to analyzing statistical distributions and constructing confidence intervals.
	
	The DP-SGD algorithm (Algorithm \ref{alg:dp-sgd}) is widely used for preserving privacy in machine learning. It has been successfully integrated into popular deep learning frameworks such as PyTorch \citep{opacus} and TensorFlow \citep{tensorflow2020}, contributing to its widespread adoption in both academia and industry. \cite{bassily2014private} demonstrated that DP-SGD satisfies $(\varepsilon,\delta)$-DP and provided an upper error bound for the last iterate. \cite{abadi2016deep} further improved upon the results of \cite{bassily2014private} by introducing the \textit{moments accountant} method, which offers a tighter privacy analysis. \cite{chourasia2021differential} investigated the information leakage of iterative randomized learning algorithms and derived a tight bound under Rényi Differential Privacy (RDP). \cite{altschuler2022privacy} extended the work of \cite{chourasia2021differential} and showed that, when only the last iterate is reported, the variance of the added noise can be bounded. However, the final iteration of SGD often exhibits excessive variability, making it unsuitable for statistical inference. Under the Gaussian Differential Privacy (GDP) framework, \cite{dong2022gaussian} and \cite{bu2020deep} introduced a private central limit theorem and proved that the DP-SGD algorithm achieves $\mu$-GDP under mild conditions. This framework has also been used for adaptive multiple testing \citep{xia2023adaptive} and knockoff inference \citep{cai2026knockoff}. Another line of work has advanced understanding of DP-SGD's generalization behavior through the lens of algorithmic stability. For example, \cite{bassily2020stability} analyzes the uniform stability of DP-SGD for non-smooth convex losses and derives corresponding generalization bounds. Furthermore, \cite{wang2022differentially} establish generalization guarantees for DP-SGD with non-smooth convex losses in both bounded and unbounded domains under less restrictive assumptions. These developments demonstrate that DP-SGD remains robust well beyond smooth and convex regimes. Despite extensive work on DP-SGD, its asymptotic distribution and statistical inference have remained largely unexplored.
	
	\subsection{Contributions}
	
	We develop a statistical inference framework for DP-SGD under randomized subsampling. Our contributions are threefold:
	
	\begin{enumerate}
		\item Under the non-private setting, we derive the asymptotic distribution of the averaged iterates of SGD based on the {\it randomized rule}. When $T = kn$ and the batch size $|I_t| = m$, the asymptotic variance inflates by a factor of $\{1+1/(km)\}$ compared to cyclic SGD in the existing literature.
		\item We prove the asymptotic normality of the averaged DP-SGD algorithm iterates. Its asymptotic variance contains three terms, corresponding to the statistical variance, the randomization variance, and the differential privacy variance, respectively. The analysis broadly accommodates most commonly used privacy definitions.
		\item We propose two methods for constructing valid confidence intervals for the DP-SGD algorithm: the plug-in method and the random scaling pivotal method. The latter involves deriving the functional CLT based on the {\it randomized rule}, which is non-trivial and could be of independent interest.
	\end{enumerate}
	
	\subsection{Notation}
	For a vector $\boldsymbol{x}\in\mathbb{R}^p$, write $\|\boldsymbol{x}\|_2=(\sum_{i=1}^p x_i^2)^{1/2}$ and $\|\boldsymbol{x}\|_\infty=\max_{1\leq i\leq p}|x_i|$. For matrices, $\|\cdot\|_2$ and $\|\cdot\|_F$ denote the operator and Frobenius norms, respectively. For a positive deterministic sequence $a_n$, $\boldsymbol{b}_n=O_p(a_n)$ means that $\|\boldsymbol{b}_n\|_2/a_n$ is bounded in probability, and $\boldsymbol{b}_n=o_p(a_n)$ means that this ratio converges to zero in probability. We write $\stackrel{p}{\to}$ and $\stackrel{d}{\to}$ for convergence in probability and distribution, respectively. The symbols $\nabla f(Z,\boldsymbol{\theta})$ and $\nabla^2 f(Z,\boldsymbol{\theta})$ denote the gradient vector and Hessian matrix with respect to $\boldsymbol{\theta}$, respectively.
	
	\section{Randomized Stochastic Gradient Descent}
	
	Existing inference results for stochastic gradient descent \citep{chen2020statistical, lee2022fast} apply only when the index $I_t$ in update (\ref{eq: update}) is selected according to the {\it cyclic rule}. However, it is necessary to use the {\it randomized rule} in differentially private optimizations. Specifically, the index $I_t$ must be chosen randomly from $\{1, 2, \dots, n\}$ so that the {\it privacy amplification via sampling} \citep{beimel2014bounds} can be applied. This follows from the definition of differential privacy (DP) and the randomized subsampling scheme. Intuitively, if an adversary tries to infer whether someone (for example, John) was present in the dataset, they face uncertainty due to random subsampling. There is a possibility that John's data was removed during the sampling step. This uncertainty dilutes the adversary’s confidence and enhances privacy protection. See \cite{balle2018privacy} for formal details.
	
	As a result, before we study the statistical inference for DP-SGD, a natural question arises: \textit{What is the asymptotic distribution for non-private stochastic gradient descent with the randomized rule}? Next, we begin with a simple example to illustrate the difference between SGD using the {\it cyclic rule} and the {\it randomized rule}.
	
	\begin{example}
		\label{ex:ran.vs.seq}
		Consider the loss function $l(x;\theta)=(x-\theta)^2$ and $\{x_i\}_{i=1}^{n}$ are i.i.d. samples from $\mathcal{N}(\mu, \sigma^2)$. We use SGD to estimate the parameter $\mu$ with initial value $\theta^{(0)}=0$ and stepsize $\eta_t=1/2$. For illustration purposes, we set the batch size to $1$. For the {\it cyclic rule}, we can verify that $\bar{\theta}_{cyc}=\sum_{i=1}^{n}x_i/n$ with asymptotic distribution that satisfies $\sqrt{n}(\bar{\theta}_{cyc}-\mu)\overset{d}{\to} N(0,\sigma^2)$. For the {\it randomized rule} with $T=n$ iteration, it follows that $\bar{\theta}_{ran}=\sum_{i=1}^{n}x_{(i)}/n$, where $\{x_{(i)}\}_{i=1}^{n}$ are simple random samples with replacement from the dataset $\{x_i\}_{i=1}^{n}$. We can verify that the following asymptotic result holds: $\sqrt{n}(\bar{\theta}_{ran}-\mu)\stackrel{d}{\to}N(0,2\sigma^2)$. Furthermore, when each iteration uses a batch size $m$, and the procedure iterates $T$ times, where $T=k\cdot n$ and $k$ is a fixed constant. We can verify that $\sqrt{n}(\bar{\theta}_{ran}-\mu)\stackrel{d}{\to}N(0,\{1+1/(km)\}\sigma^2)$. In the limiting regime $k\to\infty$, the optimization-induced error becomes \textit{asymptotically negligible}, yielding $\sqrt{n}(\bar{\theta}_{ran}-\mu)\stackrel{d}{\to}N(0,\sigma^2)$. A detailed theoretical derivation is provided in the Supplementary Material (SM).
	\end{example}
	
	As shown in Example \ref{ex:ran.vs.seq}, unlike the cyclic rule, the asymptotic distribution of the estimator using the randomized rule depends on the iteration $T$. When the iteration satisfies $T=k\cdot n$ for a constant $k$, the asymptotic variance of the mean estimator using random draws depends on $k$. Especially when $k\to\infty$, the asymptotic distribution of $\bar{\theta}_{ran}$ satisfies $\sqrt{n}(\bar{\theta}_{ran}-\mu)\stackrel{d}{\to}N(0,\sigma^2)$ and is equal to the asymptotic distribution of the full sample estimator. We formulate the following regularity conditions.
	
	\begin{assumption}[Strong convexity and Hessian]\label{ass:strong_convex}
		The loss function $\mathcal{L}(\cdot)$ is strongly convex,
		\begin{equation}
			\label{eq:strong_convex}
			\mathcal{L}(\btheta^{\prime})\geq \mathcal{L}(\btheta)+\langle\nabla \mathcal{L}(\btheta),\btheta^{\prime}-\btheta\rangle+(\lambda/2)\|\btheta-\btheta^{\prime}\|^2,
		\end{equation}
		for any $\btheta,\btheta^{\prime}\in\Theta$ and some $\lambda>0$, where $\Theta$ is a compact neighborhood of $\btheta^*$ with $\btheta^*\in\operatorname{int}(\Theta)$. The Hessian matrix $\nabla^2 \mathcal{L}(\btheta^*)=\boldsymbol{A}$ exists.
	\end{assumption}
	
	\begin{assumption}\label{ass:innovation}
		The following properties hold for the sequence $\boldsymbol{\xi}(\btheta)=\nabla \mathcal{L}(\boldsymbol{\theta})-\nabla l(\boldsymbol{z};\boldsymbol{\theta})$, with parts (ii)--(iii) imposed for $\btheta\in\Theta$:
		\begin{enumerate}[(i)]
			\item The function $l(\boldsymbol{Z};\boldsymbol{\theta})$ is convex and continuously differentiable in $\boldsymbol{\theta}\in\mathbb{R}^p$ for any $\bZ$. There exists a nonnegative random variable $B(\bZ)$ such that $\|\nabla l(\bZ;\btheta)\|_2\leq B(\bZ)(1+\|\btheta\|_2)$ for all $\btheta\in\mathbb{R}^p$ and $\mathbb{E}\{B(\bZ)^2\}<\infty$. Thus the gradient is integrable.
			\item The conditional covariance of $\bxi$ has an expression around $\btheta^*$ such that for $\bDelta=\btheta-\btheta^*$, we have $\mathbb{E}\bxi(\btheta)\bxi(\btheta)^{\top}=\bS+\bSigma(\bDelta)$, where $\|\bSigma(\bDelta)\|\leq\Sigma_1\|\bDelta\|_2+\Sigma_2\|\bDelta\|_2^2$ and $|\text{tr}\{\bSigma(\bDelta)\}|\leq\Sigma_1\|\bDelta\|_2+\Sigma_2\|\bDelta\|_2^2$ for two constants $\Sigma_1,\Sigma_2$.
			\item There exist constants $\Sigma_3,\Sigma_4$ such that the fourth conditional moment of $\boldsymbol{\xi}(\btheta)$ is bounded by $\Sigma_3+\Sigma_4\|\bDelta\|_2^4$.
		\end{enumerate}
	\end{assumption}
	
	The strong-convexity and innovation-moment conditions are standard in \cite{chen2020statistical, polyak1992acceleration, ruppert1988efficient}. Strong convexity and the existence of a Hessian matrix are essential for the asymptotic normality of the full-sample estimator. See, for example, Chapter 5 of \cite{van2000asymptotic}. The growth and smoothness conditions hold for the linear and logistic models under the bounded-support conditions. Establishing the limiting distribution for SGD under the {\it randomized rule} is more challenging than for previous results based on the {\it cyclic rule}, as the random sampling scheme creates a complicated dependence structure in each iterate. Consequently, the conventional method of stochastic process approximation \citep{polyak1992acceleration} cannot be directly applied. Instead, we employ techniques from survey sampling \citep{fuller2011sampling}.
	
	\begin{theorem}
		\label{thm:ran_sgd_normal}
		Suppose that Assumptions \ref{ass:strong_convex}, \ref{ass:innovation}, and \ref{ass:smooth_var} hold. If the step size satisfies $\eta_t=\eta\cdot t^{-\alpha}$ for $\alpha\in(1/2,1)$, the batch size $m$ is bounded and the iteration satisfies $T=k\cdot n$ for $k\geq c_0$,
		\begin{equation}
			\label{eq:normality}
			\{1+1/(km)\}^{-1/2}\sqrt{n}(\boldsymbol{\bar{\theta}}_{T}-\boldsymbol{\theta}^*)\stackrel{d}{\to}N(\boldsymbol{0},\A^{-1}\bS\A^{-1}),
		\end{equation}
		where the Hessian matrix is $\bA = \nabla^2 \calL(\btheta^*)$ and $\bS = \E\{ \nabla l(\btheta^*, Z) \nabla l(\btheta^*, Z)^{\top}\}$.
	\end{theorem}
	The lower bound $c_0$ ensures that the SGD estimator $\bar{\boldsymbol{\theta}}_T$ attains the same $n^{-1/2}$ convergence rate as the full-sample estimator. In previous works, $T$ is set to $n$ due to the algorithm's cyclic nature. Here, $T=k\cdot n$ and is more suitable for real applications. For example, $k$ can be regarded as similar to the concept of {\it epoch} in modern deep learning algorithms. It refers to the average number of times the entire dataset is used to train the model. When $k$ is a bounded real number, the asymptotic variance inflates $\boldsymbol{A}^{-1}\boldsymbol{S}\boldsymbol{A}^{-1}/(km)$ compared to that of the full sample estimator. As demonstrated in Example \ref{ex:ran.vs.seq}, the variance increase can be attributed to the inefficient use of data during subsampling. When $k\to\infty$, the SGD based on the {\it randomized rule} becomes as efficient as that based on the {\it cyclic rule}.
	
	Theorem \ref{thm:ran_sgd_normal} also highlights the impact of batch size, an aspect less frequently discussed in current studies on SGD, which often focus on the online setting or assume a batch size of $m=1$. In practice, however, the choice of batch size plays a critical role, especially in training large language models, where the batch size varies significantly based on model size, available hardware, and optimization goals \citep{vaswani2017attention}. As the batch size $m$ increases, the additional term in the asymptotic variance of the estimator decreases in proportion to $1/m$.
	
	\section{Limiting Distribution for DP-SGD}
	
	In this section, we present the privacy guarantee and the asymptotic distribution of the differentially private stochastic gradient descent (DP-SGD) algorithm, as summarized in Algorithm \ref{alg:dp-sgd}. The core principle of differential privacy is to add appropriately scaled random noise to the algorithm's output, limiting the extent to which an individual's information can be inferred from the released results. The formal definitions of DP are in the SM. To formalize this, we introduce the concept of {\it sensitivity}, which quantifies the maximum change in the algorithm's output when a single entry in the dataset is modified.
	
	\begin{defi}[$\ell_2$-Sensitivity]
		\label{def:sensitivity}
		For a vector-valued deterministic algorithm $\mathcal{T}(\cdot):\mathcal{D}\to\mathbb{R}^{p}$, the $\ell_2$ sensitivity of $\mathcal{T}(\cdot)$ is defined as
		\begin{equation}
			\sup_{\calD,\calD'}\|\mathcal{T}(\calD)-\mathcal{T}(\calD')\|_2,
		\end{equation}
		where the two datasets $\calD$ and $\calD'$ only differ in one single entry.
	\end{defi}
	
	To establish the theoretical properties of DP-SGD, we introduce the following regularity assumptions on the gradient, which are commonly adopted in the DP-SGD literature.
	
	\begin{assumption}
		\label{ass:privacy_theta}
		The $\ell_2$ sensitivity of $\nabla l(\bZ;\btheta)$ is uniformly bounded by a constant $\Delta_g>0$.
	\end{assumption}
	
	The assumption of bounded gradient sensitivity is widely used in the DP literature; see \cite{song2013stochastic,avella2021differentially}. In Algorithm \ref{alg:dp-sgd}, the noise parameter $\sigma_1$ is proportional to $\Delta_g$; further details are provided in Corollary \ref{coro:rd_thm1} and Corollary \ref{coro:gd_thm1}. The following theorem establishes the general convergence properties of DP-SGD.
	
	\begin{algorithm}[h]
		\caption{Differentially Private Stochastic Gradient Descent Algorithm}
		\label{alg:dp-sgd}
		\begin{algorithmic}[1]
			\Require Data set $\{\boldsymbol{z}_i\}_{i=1}^{n}$, loss function $l(\cdot,\cdot)$, iteration $T$, batch size $m$, noise scale $\sigma_1$, stepsize $\{\eta_t\}_{t=1}^{T}$;
			\For{$t=1\dots T$}
			\State Randomly sample: Take a uniformly random subsample $I_t\subset[n]$ with batch size $m$;
			\State Compute gradient: $\boldsymbol{g}^{(t)}=\sum_{i\in I_t}\nabla l(\boldsymbol{z}_i;\btheta^{(t-1)})/m$;
			\State Descent and perturb: $\btheta^{(t)}=\btheta^{(t-1)}-\eta_t(\boldsymbol{g}^{(t)}+\boldsymbol{\xi}_t)$, where $\boldsymbol{\xi}_t$ is an independent draw from a multivariate normal distribution $N(\boldsymbol{0},\sigma_1^2\boldsymbol{I})$;
			\EndFor
			\Ensure $\bar{\btheta}_T=\sum_{t=1}^{T}\btheta^{(t)}/T$;
		\end{algorithmic}
	\end{algorithm}
	
	\begin{theorem}
		\label{thm:dp_sgd}
		Suppose that Assumptions \ref{ass:strong_convex}, \ref{ass:innovation}, \ref{ass:privacy_theta}, and \ref{ass:smooth_var} hold. If the step size satisfies $\eta_t=\eta\cdot t^{-\alpha}$ for $\alpha\in(1/2,1)$, the batch size $m$ is bounded, the iteration satisfies $T=k\cdot n$ for $k\geq c_0>0$, and $\sigma_1=O(1)$, then the output of Algorithm \ref{alg:dp-sgd} satisfies
		\begin{equation}
			\label{eq:normality_sgd}
			\sqrt{n}(\boldsymbol{\bar{\theta}}_T-\boldsymbol{\theta}^*) \stackrel{d}{\to}\boldsymbol{\phi}_{stat}+\boldsymbol{\phi}_{sam}+\boldsymbol{\phi}_{privacy},%+\bb,
		\end{equation} 
		The three terms on the right side of (\ref{eq:normality_sgd}) are independent and distributed as  $\boldsymbol{\phi}_{stat}\stackrel{d}{=}N(\boldsymbol{0},\boldsymbol{A}^{-1}\boldsymbol{S}\boldsymbol{A}^{-1})$, $\boldsymbol{\phi}_{sam}\stackrel{d}{=}N(\boldsymbol{0},\boldsymbol{A}^{-1}\boldsymbol{S}\boldsymbol{A}^{-1}/(km))$ and $\boldsymbol{\phi}_{privacy}\stackrel{d}{=}N(\boldsymbol{0},\sigma_1^2\boldsymbol{A}^{-2}/k)$.
	\end{theorem}
	
	In Theorem \ref{thm:dp_sgd}, the right side of (\ref{eq:normality_sgd}) contains three terms: $\boldsymbol{\phi}_{stat}$, $\boldsymbol{\phi}_{sam}$, and $\boldsymbol{\phi}_{privacy}$. The first term, $\boldsymbol{\phi}_{stat}$, represents the statistical limiting distribution of the full-sample estimator $\widehat{\boldsymbol{\theta}}$, which is the same as the distribution in \cite{polyak1992acceleration}. The second term $\boldsymbol{\phi}_{sam}$ represents the sampling error due to the {\it randomized rule} used in the SGD algorithm. The sum of the first two terms, $\boldsymbol{\phi}_{stat}+\boldsymbol{\phi}_{sam}$, corresponds to the asymptotic distribution of the non-private SGD under the randomized rule, as established in Theorem \ref{thm:ran_sgd_normal}. The last term $\boldsymbol{\phi}_{privacy}$ accounts for the randomization due to privacy protection.
	
	When choosing the random subsample $I_t$ in the SGD algorithm, two approaches are common in the literature: simple random sampling without replacement (SRSWOR) and Poisson subsampling. SRSWOR selects a minibatch by uniformly sampling a fixed number of data points from the dataset without replacement. Each minibatch, therefore, has the same size. Poisson subsampling selects each data point independently with a fixed probability, leading to a random minibatch size. For Poisson subsampling, gradients are divided by the expected batch size $m$, with an empty minibatch contributing zero. SRSWOR can be more statistically efficient than Poisson subsampling. The detailed explanations are provided in Section B of the SM. Poisson subsampling introduces additional randomness and requires its corresponding privacy accounting. Nevertheless, both schemes are standard in DP-SGD and are fully compatible with our theoretical assumptions. We demonstrate through the following three corollaries that by choosing $\sigma_1^2$ carefully, we can ensure not only that the output of Algorithm \ref{alg:dp-sgd} satisfies DP, but also that the asymptotic variance of the DP-SGD estimator $\bar{\btheta}_T$ matches that of the non-private full sample estimator. The definitions of the various DP notions are provided in Section A of the SM.
	
	\begin{corollary}
		\label{coro:ed_thm1}
		$(\varepsilon,\delta)$-DP: Suppose the conditions in Theorem \ref{thm:dp_sgd} hold. Fix privacy parameters $\varepsilon>0$ and $\delta\in(0,1)$ and a positive integer batch size $m$, and let $T=\lceil n^{l}\rceil$ for $1<l<2$. For a suitable choice of the noise standard deviation $\sigma_1=O(1)$, the DP-SGD estimator $\bar{\btheta}_T$ in Algorithm \ref{alg:dp-sgd} using SRSWOR is $(\varepsilon,\delta)$-DP for all sufficiently large $n$, and satisfies
		$$\sqrt{n}(\boldsymbol{\bar{\theta}}_T-\boldsymbol{\theta}^*)\stackrel{d}{\to}N(\boldsymbol{0},\boldsymbol{A}^{-1}\boldsymbol{S}\boldsymbol{A}^{-1}).$$
	\end{corollary}
	
	\begin{corollary}
		\label{coro:rd_thm1}
		$(\gamma,\varepsilon)$-RDP: Suppose the conditions in Theorem \ref{thm:dp_sgd} hold. Fix a R\'enyi order $\gamma\geq 2$. For $0<\varepsilon<c_1m^2T/n^2$, let $T=\lceil n^{l}\rceil$ for $1<l<2$, and the standard deviation $\sigma_1=c_2\Delta_gm/n\sqrt{T/\varepsilon}$, where $c_1>0$ and $c_2$ is a sufficiently large positive constant that may depend on $\gamma$ and $c_1$. The DP-SGD estimator $\bar{\btheta}_T$ using SRSWOR is $(\gamma,\varepsilon)$-RDP for all sufficiently large $n$. Furthermore, when $\sqrt{T/\varepsilon}/n=O(1)$, $\bar{\btheta}_T$ satisfies
		$$\sqrt{n}(\boldsymbol{\bar{\theta}}_T-\boldsymbol{\theta}^*)\stackrel{d}{\to}N(\boldsymbol{0},\boldsymbol{A}^{-1}\boldsymbol{S}\boldsymbol{A}^{-1}).$$
	\end{corollary}
	
	\begin{corollary}
		\label{coro:gd_thm1}
		$\mu$-GDP: Suppose the conditions in Theorem \ref{thm:dp_sgd} hold and $m\sqrt{T}/n\to c$ as $n\to\infty$ for $c>0$. For the privacy parameter $\mu>0$, there exists $\sigma>0$ such that
		$$\mu=\sqrt{2}c\cdot\sqrt{e^{\sigma^{-2}}\Phi(1.5\sigma^{-1})+3\Phi(-0.5\sigma^{-1})-2},$$
		where $\Phi(\cdot)$ denotes the cumulative distribution function of the standard normal distribution.
		The DP-SGD estimator $\bar{\btheta}_T$ using SRSWOR is asymptotically $\mu$-GDP, if the noise parameter satisfies $\sigma_1=2\sigma\Delta_g/m$. Furthermore, when $\mu$ is bounded away from $0$, then
		$$\sqrt{n}(\boldsymbol{\bar{\theta}}_T-\boldsymbol{\theta}^*)\stackrel{d}{\to}N(\boldsymbol{0},\boldsymbol{A}^{-1}\boldsymbol{S}\boldsymbol{A}^{-1}).$$
	\end{corollary}
	
	Statistical inference for DP-SGD has clear advantages over the \textit{Differentially Private Gradient Descent} (DP-GD) algorithm studied in \cite{avella2021differentially}. Under the $\mu$-GDP framework, the privacy error of the DP-GD algorithm is of order
	$O_p(\sqrt{T_{gd}}/(n\mu))$, where $T_{gd}$ denotes the number of iterations and satisfies $T_{gd}=O(\log(n))$. For the DP-SGD algorithm, Theorem \ref{thm:dp_sgd} and the proof of Corollary \ref{coro:gd_thm1} establish that the error introduced by the privacy guarantee in the estimation $\btheta$ is $O_p(1/(n\mu))$, which is unrelated to the number of iterations and much smaller compared to the privacy error of DP-GD. Under the $(\varepsilon,\delta)$-DP framework of Corollary \ref{coro:ed_thm1}, a suitable bounded noise standard deviation gives a leading privacy-noise component of order $O_p(\sigma_1/\sqrt{T})=O_p(T^{-1/2})$. Since $T=\lceil n^l\rceil$ with $1<l<2$, this component is $o_p(n^{-1/2})$ and does not affect the full sample asymptotic variance.
	
\begin{figure}[ht]
		\centering
		\includegraphics[width=0.6\linewidth]{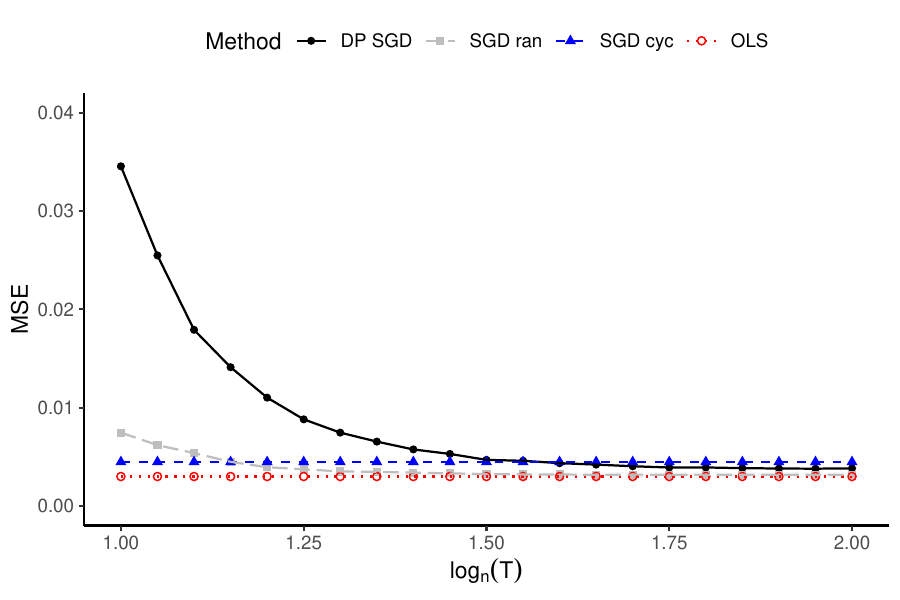}
		\caption{Mean square error (MSE) for linear regression with $n=1000$ samples and $p=3$ covariates, averaged over $1000$ repetitions. The number of iterations $T$ increases from $n$ to $n^2$. The four algorithms are DP-SGD, randomized SGD (SGD-ran), cyclic SGD (SGD-cyc), and the oracle estimator (OLS).}
		\label{fig:1}
	\end{figure}

	We conduct a numerical study in Figure \ref{fig:1} to further illustrate the relation between algorithm convergence and the number of iterations $T$. We plot the MSE against $T$ increasing from $n$ to $n^2$. The DP-GD algorithm is not shown in the figure because it cannot be displayed on the plot when $T = O(\log n)$. We use the \text{ordinary least squares} (OLS) estimator in red as the benchmark. Note that the OLS does not depend on $T$, and the cyclic SGD no longer changes when $T>n$. Those two algorithms are shown as horizontal lines in the figure. When $T=n$, randomized SGD has around twice the MSE of OLS and is less efficient than cyclic SGD. Those observations are consistent with our derived theories. However, when the iterations $T$ reach $n^{1.3}$, randomized SGD achieves an MSE similar to OLS and outperforms cyclic SGD. The DP-SGD algorithm is shown to match the performance of randomized SGD closely.
	
	\section{Confidence Interval of the DP-SGD Algorithm}
	
	In this section, we propose two methods for constructing valid confidence intervals for the DP-SGD estimates. The first method involves plugging the estimated parameter into the Hessian matrix and the score covariance function. The second method is based on random-scaled pivotal statistics and involves deriving the functional Central Limit Theorem. 
	
	\subsection{Plug-in Covariance Matrix Estimation}
	
	In this section, we introduce a consistent estimator for the covariance matrix of $\bar{\btheta}_T$. By the normality result established in Theorem \ref{thm:dp_sgd}, we construct an asymptotic confidence region using the estimated covariance matrix. We propose to estimate the asymptotic covariance matrix $\bV=\bA^{-1}\bS\bA^{-1}$ by applying the plug-in method separately to estimate the matrices $\bA$ and $\bS$. Note that the nonprivate full-sample estimators of $\bA$ and $\bS$ are
	\begin{equation*}
		\widehat{\bA}=\frac{1}{n}\sum_{i=1}^{n}\nabla^2 l(\bz_i;\bar{\btheta}_T)\quad\text{and}\quad\widehat{\bS}=\frac{1}{n}\sum_{i=1}^{n}\nabla l(\bz_i;\bar{\btheta}_T)\nabla l(\bz_i;\bar{\btheta}_T)^{\top},
	\end{equation*}
	respectively. To ensure privacy, we propose to estimate the covariance matrix by
	\begin{equation*}
		\widetilde{\bA}=\widehat{\bA}+\bxi^{\prime}\quad\text{and}\quad\widetilde{\bS}=\widehat{\bS}+\bxi^{\prime\prime},
	\end{equation*}
	where components of $\bxi_t^{\prime}$ and $\bxi_t^{\prime\prime}$ are i.i.d. draws from normal distribution $N(0,\sigma_2^2)$ and $N(0,\sigma_3^2)$, respectively. We allocate an additional privacy budget to the joint covariance release in addition to the budget for DP-SGD. The following proposition shows that the variance estimations satisfy privacy guarantees under appropriate conditions.
	
	\begin{assumption}
		\label{ass:privacy_var}
		The $l_2$ sensitivities of the matrices $\nabla^2 l(\bZ;\btheta)$ and $\nabla l(\bZ;\btheta)\nabla l(\bZ;\btheta)^{\top}$ are bounded by the constants $\Delta_A$ and $\Delta_S$. Matrix sensitivity is measured after vectorization.
	\end{assumption}
	
	\begin{proposition}
		\label{prop:privacy}
		Suppose Assumption \ref{ass:privacy_var} holds. Then, the estimates \( \widetilde{\bA} \) and \( \widetilde{\bS} \) satisfy the following privacy guarantees:
		\begin{itemize}
			\item $(\varepsilon,\delta)$-DP for $0<\varepsilon<2$ and $0<\delta<1$ when $\sigma_2\geq 2(\Delta_A/n)\sqrt{2\log(2.5/\delta)}/\varepsilon$ and $\sigma_3\geq 2(\Delta_S/n)\sqrt{2\log(2.5/\delta)}/\varepsilon$;
			\item $(\gamma,\varepsilon)$-RDP when $\sigma_2\geq(\Delta_A/n)(\gamma/\varepsilon)^{1/2}$ and $\sigma_3\geq(\Delta_S/n)(\gamma/\varepsilon)^{1/2}$;
			\item $\mu$-GDP when $\sigma_2\geq(\Delta_A/n)(\sqrt{2}/\mu)$ and $\sigma_3\geq(\Delta_S/n)(\sqrt{2}/\mu)$.
		\end{itemize}
	\end{proposition}
	
	The bounded sensitivity conditions are frequently assumed in the DP literature; see \cite{avella2021differentially}. The privacy guarantee follows directly from the Gaussian mechanism as presented in the SM. Since the variance estimators depend on the Hessian matrix of the loss function, we use the following smoothness assumption, also invoked in Theorems \ref{thm:ran_sgd_normal} and \ref{thm:dp_sgd}. Given standard deviations $\sigma_2,\sigma_3$ of additional random errors $\bxi^{\prime},\bxi^{\prime\prime}$, Proposition \ref{prop:plug_in} provides the error bound of the plug-in variance estimator.
	
	\begin{assumption}
		\label{ass:smooth_var}
		There are constants $L_2$ and $L_4$ such that for all $\btheta\in\Theta$,
		\begin{equation*}
			\|\nabla^2 l(\bZ;\btheta)-\nabla^2 l(\bZ;\btheta^*)\|_2\leq L_2\|\btheta-\btheta^*\|_2,
		\end{equation*}
		\begin{equation*}
			\mathbb{E}\|\nabla^2 l(\bZ;\btheta^*)-\bA\|_F^2\leq L_4.
		\end{equation*}
	\end{assumption}
	
	\begin{proposition}
		\label{prop:plug_in}
		Suppose conditions in Theorem \ref{thm:dp_sgd} and Assumption \ref{ass:smooth_var} hold. The private estimators $\widetilde{\bA}$ and $\widetilde{\bS}$ satisfy
		\begin{equation*}
			\|\widetilde{\bA}-\bA\|_2=O_p(n^{-1/2}+\sigma_1T^{-1/2}+\sigma_2),
		\end{equation*}
		\begin{equation*}
			\|\widetilde{\bS}-\bS\|_2= O_p(n^{-1/2}+\sigma_1T^{-1/2}+\sigma_3),
		\end{equation*}
		for a constant $C$ depending on $L_2,L_4$. Thus, the estimated variance $\widetilde{\bV}=\widetilde{\bA}^{-1}\widetilde{\bS}\widetilde{\bA}^{-1}$ satisfies
		\begin{equation*}
			\|\widetilde{\bV}-\bA^{-1}\bS\bA^{-1}\|_2=O_p(n^{-1/2}+\sigma_1T^{-1/2}+\sigma_2+\sigma_3).
		\end{equation*}
	\end{proposition}
	
	Before forming the covariance estimate or applying spectral thresholding, we symmetrize each noisy matrix $\bM$ as $(\bM+\bM^{\top})/2$ and retain the same notation. This postprocessing preserves privacy and the error rates in Proposition \ref{prop:plug_in}. Because the estimated variance is given by $\widetilde{\bA}^{-1}\widetilde{\bS}\widetilde{\bA}^{-1}$, it is required to accurately estimate $\widetilde{\bA}^{-1}$. To avoid potential singularity due to statistical randomness, we can employ the same thresholding estimator as in \cite{chen2020statistical}. For the consistent covariance estimates in Theorem \ref{theorem:cover_plugin}, a positive eigenvalue floor tending to zero guarantees invertibility and is asymptotically inactive. The bounds in Proposition \ref{prop:plug_in} apply to the unthresholded symmetrized estimates; in this consistent regime, they also hold after thresholding. However, such thresholding is typically unnecessary in practice, as $\widetilde{\bA}$ converges to $\bA$ with high probability. Indeed, in our experiments, we observed that $\widetilde{\bA}$ remained invertible without eigenvalue thresholding.
	
	Using the estimated variance $\widetilde{\bV}$, the asymptotic confidence interval for $\theta_j$ at the $1-\alpha$ level, where $j=1,\dots,p$, is given by:
	\begin{equation}
		\label{eq:ci-plug}
		\text{CI}_{\alpha}(\theta_j):=\big(\bar{\theta}_{T,j}+c_{\alpha/2}\sqrt{(\widetilde{\bV})_{j,j}/n},\bar{\theta}_{T,j}+c_{1-\alpha/2}\sqrt{(\widetilde{\bV})_{jj}/n}\big)\quad\text{for}\quad j=1,\dots,p,
	\end{equation}
	where $\bar{\theta}_{T,j}$ is the $j$-th component of $\bar{\btheta}_T$, and $c_{\alpha/2},c_{1-\alpha/2}$ are $\alpha/2,1-\alpha/2$ quantiles of standard normal distribution, respectively. The following theorem shows that the constructed confidence intervals are asymptotically valid.
	
	\begin{theorem}
		\label{theorem:cover_plugin}
		Suppose the conditions in Theorem \ref{thm:dp_sgd} and Proposition \ref{prop:plug_in} hold. Furthermore, under the DP (RDP or GDP) framework, suppose the conditions in Proposition \ref{prop:privacy} and Corollary \ref{coro:ed_thm1} (Corollary \ref{coro:rd_thm1} or Corollary \ref{coro:gd_thm1}) are satisfied. In addition, assume that $\sigma_2=o(1)$, $\sigma_3=o(1)$, and $V_{j,j}>0$ for every $j=1,\dots,p$, where $\bV=\bA^{-1}\bS\bA^{-1}$. Then, for any $j=1,\dots,p$ and $\alpha\in(0,1)$, we have
		\begin{equation*}
			\lim_{n\to\infty}\mathbb{P}\{\theta_j^*\in\text{CI}_{\alpha}(\theta_j)\}=1-\alpha.
		\end{equation*}
	\end{theorem}
	
	Theorem \ref{theorem:cover_plugin} demonstrates that the proposed confidence interval satisfies DP and achieves an asymptotic coverage probability of $1-\alpha$. The conditions $\sigma_2=o(1)$ and $\sigma_3=o(1)$ are mild and readily satisfied by choosing the noise scales at the lower bounds in Proposition \ref{prop:privacy}. Besides, Theorem \ref{theorem:cover_plugin} also implies that $$\sqrt{n}(\bar{\theta}_{T,j}-\theta^*_j)\big\{(\widetilde{\bV})_{j,j}\big\}^{-1/2}\stackrel{d}{\to}N(0, 1).$$ 
	As illustrated in Figure \ref{fig:1}, the additional randomness of the DP-SGD estimator due to privacy, though small, may not be negligible in small samples. To improve the finite-sample performance of the inference procedure, we propose to apply a correction to the variance by considering the extra noise added by DP: 
	\begin{equation}
		\label{eq:plug_correct}
		\begin{split}
			\text{CI}_{\alpha}^c(\theta_j):=\big(&\bar{\theta}_{T,j}+c_{\alpha/2}\sqrt{(\widetilde{\bV})_{j,j}/n+\sigma_1^2(\widetilde{\bA}^{-2})_{j,j}/(kn)},\\
			&\bar{\theta}_{T,j}+c_{1-\alpha/2}\sqrt{(\widetilde{\bV})_{j,j}/n+\sigma_1^2(\widetilde{\bA}^{-2})_{j,j}/(kn)}\big),
		\end{split}
	\end{equation}
	for $j=1,\dots,p$. The correction considers the variance of the statistical error $\bphi_{stat}$ and the privacy error $\bphi_{privacy}$ in Theorem \ref{thm:dp_sgd}. It can be shown that the finite-sample corrected confidence interval is asymptotically valid: $\lim_{n\to\infty}\mathbb{P}\{\theta_j^*\in\text{CI}_{\alpha}^c(\theta_j)\}=1-\alpha$, and is asymptotically as efficient as the confidence interval $\text{CI}_{\alpha}(\theta_j)$ defined in \eqref{eq:ci-plug}. A similar procedure was also considered in \cite{avella2021differentially} and \cite{cai2023private} to improve finite-sample performance.

	\subsection{Asymptotically Pivotal Statistics after Random Scaling}
	
	This section proposes using the random scaling method to construct asymptotically pivotal statistics, enabling the construction of confidence intervals without estimating the covariance matrix. A common approach in statistics is to estimate the asymptotic variance and then plug the estimated variance into Wald-type confidence intervals. However, this plug-in method has a significant drawback: from a data privacy perspective, the additional release of the estimated matrices $\bA$ and $\bS$ requires an additional privacy budget, which in turn increases the privacy-induced error in parameter estimation. \cite{lee2022fast} introduced a novel method for online inference by constructing asymptotically pivotal statistics using random scaling. The key idea is to leverage the information contained in the sequence  $\{\btheta^{(t)}\}_{t=1}^{T}$ to estimate the variance of these asymptotically pivotal statistics. A major advantage of this approach is that it circumvents the need to estimate the covariance matrix separately. However, the confidence intervals constructed using pivotal statistics may be wider than those obtained via the Wald method.
	
	We begin by developing a functional central limit theorem for the sample path of the DP-SGD estimator described in Algorithm \ref{alg:dp-sgd}, which generalizes Theorem \ref{thm:dp_sgd}.
	
	\begin{proposition}
		\label{prop:funcation_CLT}
		Suppose the conditions in Theorem \ref{thm:dp_sgd} hold. Then the partial-sum process of Algorithm \ref{alg:dp-sgd} satisfies, in 
		$D([0,1],\mathbb{R}^p)$,
		\begin{equation}
			\label{eq:functional_CLT}
			\frac{\sqrt{n}}{T}\sum_{t=1}^{\lfloor rT\rfloor}(\btheta^{(t)}-\btheta^*)\stackrel{d}{\to}r\bV^{1/2}\bphi+\bV^{1/2}\bW_1(r)(km)^{-1/2}+\sigma_1\bA^{-1}\bW_2(r)k^{-1/2},
		\end{equation}
		where $r\in[0,1]$ and $\lfloor rT\rfloor$ is the integer part of $rT$. The vector $\bphi$ and the processes $\bW_1,\bW_2$ are mutually independent. The vector $\bphi$ follows a $p$-dimensional standard normal distribution, and $\bW_1,\bW_2$ are standard $p$-dimensional Wiener processes \citep{durrett2019probability}.
	\end{proposition}
	
	The asymptotic distribution of the partial sum of DP-SGD based on the randomized rule differs significantly from that of SGD based on the cyclic rule. As demonstrated in Theorem 1 of \cite{lee2022fast}, the partial sum of SGD estimates under the cyclic rule converges to a single Wiener process. In contrast, for DP-SGD under the randomized rule, the asymptotic distribution in \eqref{eq:functional_CLT} contains three independent components. The first term represents the statistical error of the full-sample estimator, with the scale factor  $r$  arising from the scaling $\lfloor rT\rfloor/T$. The second term captures the additional random error introduced by the random subsampling used in each SGD update. The third term corresponds to the additional random noise injected to ensure privacy.
	
	This difference in asymptotic distributions can be attributed to the distinct ways in which information is integrated during the SGD procedure. Under the cyclic rule, the analyst sequentially acquires new information at each iteration, causing the partial sum of updates to converge to a single Wiener process. In contrast, under the randomized rule, the analyst has the potential to access information from the entire dataset at every step, which contributes to the first term in \eqref{eq:functional_CLT}. The second term in \eqref{eq:functional_CLT} emerges because information is integrated through random subsamples in each iteration, introducing additional variability. Consider the random scaling matrix
	\begin{equation}
		\label{eq:random_scaling}
		\widehat{\bV}:=\frac{m}{n}\sum_{t=1}^{T}\bigg\{\frac{\sqrt{n}}{T}\sum_{s=1}^{t}(\btheta^{(s)}-\bar{\btheta}^{(T)})\bigg\}\bigg\{\frac{\sqrt{n}}{T}\sum_{s=1}^{t}(\btheta^{(s)}-\bar{\btheta}^{(T)})\bigg\}^{\top}.
	\end{equation}
	For the regimes in Corollaries \ref{coro:ed_thm1}--\ref{coro:gd_thm1}, the uniform linearization lemma in the SM gives
	\begin{equation*}
		\widehat{\bV}\stackrel{d}{\to}\int_{0}^{1}[\bV^{1/2}\{\bW_1(r)-r\bW_1(1)\}+\bA^{-1}\sigma_1m^{1/2}\{\bW_2(r)-r\bW_2(1)\}]^{\otimes 2}\text{d}r,
	\end{equation*}
	where for a vector $\bu$, we write $\bu^{\otimes 2}:=\bu\bu^{\top}$. The following theorem gives the limiting distribution of $\sqrt{n}(\bar{\theta}_{T,j}-\theta^*_j)\widehat{V}_{j,j}^{-1/2}$ and establishes asymptotic pivotality and valid confidence intervals when $\sigma_1=o(1)$.
	
	\begin{theorem}
		\label{thm:random_scale}
		Suppose the conditions in Theorem \ref{thm:dp_sgd} and one of Corollaries \ref{coro:ed_thm1}--\ref{coro:gd_thm1} hold. Then, for any $j=1,\dots,p$ such that $V_{j,j}>0$, we have
		\begin{equation}
			\label{eq:random_scale}
			\sqrt{n}(\bar{\theta}_{T,j}-\theta^*_j)\widehat{V}_{j,j}^{-1/2}\stackrel{d}{\to}Z\cdot\bigg[\int_{0}^{1}\{W(r)-rW(1)\}^2\text{d}r\bigg]^{-1/2}\cdot\bigg\{\frac{V_{j,j}}{V_{j,j}+m\sigma_1^2(A^{-2})_{j,j}}\bigg\}^{1/2},
		\end{equation}
		where $(A^{-2})_{j,j}$ denotes the corresponding diagonal entry, and $Z\sim N(0,1)$ is independent of the standard Wiener process $W(\cdot)$. Furthermore, suppose that $\sigma_1=o(1)$, and for any $\alpha\in(0,1)$, let $c_{1-\alpha/2}$ be the $1-\alpha/2$ quantile of $Z\cdot[\int_{0}^{1}\{W(r)-rW(1)\}^2\text{d}r]^{-1/2}$. The DP confidence interval
		\begin{equation}
			\label{eq:pi_con}
			\text{CI}_{\alpha}(\theta_j):=\big(\bar{\theta}_{T,j}-c_{1-\alpha/2}(\widehat{V}_{j,j}/n)^{1/2},\bar{\theta}_{T,j}+c_{1-\alpha/2}(\widehat{V}_{j,j}/n)^{1/2}\big)
		\end{equation}
		satisfies: $
		\lim_{n\to\infty}\mathbb{P}\{\theta_j^*\in\text{CI}_{\alpha}(\theta_j)\}=1-\alpha.$
	\end{theorem}
	
	In Theorem \ref{thm:random_scale}, we first establish that the proposed statistic $\sqrt{n}(\bar{\theta}_{T,j}-\theta^*_j)\widehat{V}_{j,j}^{-1/2}$ converges to a pivotal quantity scaled by a multiplicative factor. Furthermore, under the condition that the standard deviation of each component of $\bxi_t$ in Algorithm \ref{alg:dp-sgd} satisfies $\lim_{n\to\infty}\sigma_1=0$, the statistic converges to $Z\cdot[\int_{0}^{1}\{W(r)-rW(1)\}^2\text{d}r]^{-1/2}$. The asymptotic distribution of this statistic is symmetric, and its corresponding critical values are provided in \cite{abadir1997two}. For nonvanishing privacy noise, we use the corrected interval in \eqref{eq:rs_correct} with consistent covariance estimates, as ensured by the vanishing covariance-noise conditions in Theorem \ref{theorem:cover_plugin}. 
	
	The multiplicative factor in \eqref{eq:random_scale} is at most one. Consequently, the confidence interval constructed in \eqref{eq:pi_con} can lead to over-coverage. To address this issue for small sample sizes or when $\sigma_1$ is not sufficiently small, we propose obtaining private estimates of $V_{j,j}$ and $(A^{-2})_{j,j}$, denoted as $\widetilde{V}_{j,j}$ and $(\widetilde{A}^{-2})_{j,j}$, respectively, as outlined in Proposition \ref{prop:privacy}. The finite-sample-corrected confidence interval is then given by:
	\begin{equation}
		\label{eq:rs_correct}
		\begin{split}
			\text{CI}_{\alpha}(\theta_j):=\bigg(&\bar{\theta}_{T,j}-c_{1-\alpha/2}(\widehat{V}_{j,j}/n)^{1/2}\bigg\{\frac{\widetilde{V}_{j,j}}{\widetilde{V}_{j,j}+m\sigma_1^2(\widetilde A^{-2})_{j,j}}\bigg\}^{1/2},\\
			&\bar{\theta}_{T,j}+c_{1-\alpha/2}(\widehat{V}_{j,j}/n)^{1/2}\bigg\{\frac{\widetilde{V}_{j,j}}{\widetilde{V}_{j,j}+m\sigma_1^2(\widetilde A^{-2})_{j,j}}\bigg\}^{1/2}\bigg).
		\end{split}
	\end{equation}
	
	\section{Gradient Clipping and Examples}
	\label{sec:app}
	
	Gradient clipping is a widely adopted modification to standard stochastic gradient descent that enhances the stability of the training process. At the $t$-th iteration in Algorithm \ref{alg:dp-sgd}, gradient clipping ensures that the gradient norm is constrained to be less than or equal to a predefined threshold $\tau>0$. Specifically, the clipped gradient $\boldsymbol{g}^{(t)}$ is computed as:
	\begin{equation}
		\label{eq:clipped}
		\boldsymbol{g}^{(t)}=\frac{1}{|I_t|}\sum_{i\in I_t}\nabla l(\boldsymbol{z}_i;\btheta^{(t-1)})\cdot\min\bigg(1,\frac{\tau}{\|\nabla l(\boldsymbol{z}_i;\btheta^{(t-1)})\|_2}\bigg).
	\end{equation}
	where $I_t$  represents the mini-batch at iteration $t$, and $\nabla l(\boldsymbol{z}_i;\btheta^{(t-1)})$ denotes the gradient of the loss function with respect to the model parameters. This technique has been successfully applied in various domains, including convex optimization \citep{alber1998projected} and deep learning applications \citep{goodfellow2016deep}, to mitigate challenges arising from large gradient magnitudes \citep{gurbuzbalaban2021heavy}.
	
	Besides improving training stability, gradient clipping plays a critical role in ensuring differential privacy \citep{abadi2016deep,dong2022gaussian, xia2026privateSIR}. By constraining the gradient norm, the sensitivity of the clipped gradient is bounded by $2\tau$. Under Assumption \ref{ass:privacy_theta}, it is also bounded by $\Delta_g$, since clipping is nonexpansive; here $\Delta_g$ denotes the sensitivity bound on the original gradients. This bounded sensitivity is essential for achieving differential privacy guarantees. Next, we demonstrate that the clipped DP-SGD estimator, obtained by replacing Step 3 in Algorithm \ref{alg:dp-sgd} with the clipped gradient defined in \eqref{eq:clipped}, follows the same asymptotic distribution as established in Theorem \ref{thm:dp_sgd}.

	\begin{assumption}
		\label{ass:bound_innovation}
		The clipping threshold satisfies: 
		$$\mathbb{P}\big(\sup_{\btheta\in\Theta,1\leq i\leq n}\|\nabla l(\boldsymbol{z}_i;\btheta)-\nabla \mathcal{L}(\btheta)\|_2\leq\tau/2\big)\to 1\qquad\text{as }n\to\infty.$$
	\end{assumption}
	
	Assumption \ref{ass:bound_innovation} assumes that the stochastic noise can be bounded with probability approaching 1. This condition has been employed in analyses of clipped SGD in non-private settings, as demonstrated by \cite{zhang2019gradient}, \cite{qian2021understanding}, and \cite{koloskova2023revisiting}. Under Assumption \ref{ass:privacy_theta}, the population-centered gradient deviations are uniformly bounded by $\Delta_g$. Thus, Assumption \ref{ass:bound_innovation} holds for any fixed $\tau\ge2\Delta_g$; the sufficiently large fixed threshold in Theorem \ref{thm:clip} also ensures stability of the unprojected iterations.
	
	To further clarify Assumption \ref{ass:bound_innovation}, we note that
	$\nabla\mathcal{L}(\btheta^*)=0$ at $\btheta=\btheta^*$. On the event in Assumption \ref{ass:bound_innovation}, continuity of $\nabla\mathcal L$ therefore implies that clipping is inactive in a sufficiently small neighborhood of $\btheta^*$. As shown by \cite{koloskova2023revisiting}, such conditions are essential to ensure that the bias introduced by gradient clipping diminishes asymptotically, which is required for valid statistical inference. When $\btheta^{(t)}$ lies outside this neighborhood, clipping constrains the gradient magnitude; stability under the threshold choice below is established in the proof.

	\begin{theorem}
		\label{thm:clip}
		Suppose the conditions in Theorem \ref{thm:dp_sgd} and Assumption \ref{ass:bound_innovation} hold. For a sufficiently large fixed clipping threshold $\tau$, the output of clipped DP-SGD satisfies
		\begin{equation*}
			\sqrt{n}(\boldsymbol{\bar{\theta}}_T-\boldsymbol{\theta}^*) \stackrel{d}{\to}\boldsymbol{\phi}_{stat}+\boldsymbol{\phi}_{sam}+\boldsymbol{\phi}_{privacy},%+\bb,
		\end{equation*}
		where the three terms on the right side are defined in Theorem \ref{thm:dp_sgd}.
	\end{theorem}
	
	Theorem \ref{thm:clip} establishes that the asymptotic distribution of the clipped DP-SGD estimator coincides with that of the standard DP-SGD estimator. As a result, the findings presented in Corollaries \ref{coro:ed_thm1}–\ref{coro:gd_thm1} can be derived analogously. To illustrate the practical implications of these results, we provide two specific inference examples: ordinary linear regression and logistic regression, both of which are implemented using DP-SGD.
	
	\subsection{Ordinary Linear Regression}
	\label{ex:as-linear}
	
	We begin by examining ordinary linear regression, one of the most fundamental and widely used statistical models. In this framework, the samples, denoted by $\{\bz_i\}_{i=1}^{n}=\{(\bx_i,y_i)\}_{i=1}^{n}$, are independent and identically distributed (i.i.d.) random vectors generated from the linear model: $y_i=\bx_i^{\top}\btheta^*+e_i$, where $e_i$ represents the random error term, independent of $\bx_i$, with mean $0$ and finite variance $\sigma^2$. Here, $\btheta^*$ is the unknown parameter of interest.
	We assume that the covariate vector $\bx_i$ follows a distribution with mean 
	$\boldsymbol{0}$ and a positive definite covariance matrix $\bSigma$. For this setup, we consider the squared error loss function: $l(\bz_i;\btheta):=(y_i-\bx_i^{\top}\btheta)^2/2$. The corresponding population-level objective function is given by:
	$$\mathcal{L}(\btheta)=\sigma^2/2 + (\btheta-\btheta^*)^{\top}\bSigma(\btheta-\btheta^*)/2,$$ 
	which is strongly convex with a convexity parameter $\lambda=\lambda_{min}(\bSigma)$, and $\lambda_{min}(\cdot)$ denotes the smallest eigenvalue of a matrix. The Hessian matrix of $\mathcal{L}(\btheta)$ is $\bSigma$. Furthermore, the innovation sequence is defined as:
	$$\bxi(\btheta)=(\bSigma-\bx_i\bx_i^{\top})(\btheta-\btheta^*)+\bx_ie_i$$ satisfies Assumption \ref{ass:innovation} under the bounded-support conditions below. In particular, Assumption \ref{ass:innovation}(i) holds with $B(\bZ)=\|\bx\|_2^2+|y|\|\bx\|_2$, and Assumption \ref{ass:smooth_var} holds because $\nabla^2l(\bZ;\btheta)=\bx\bx^{\top}$ does not depend on $\btheta$.
	
	Here we provide a detailed discussion of the DP-SGD estimator for $\btheta^*$. In Algorithm \ref{alg:dp-sgd}, the gradient function at the $t$-th iteration is given by $
	\boldsymbol{g}^{(t)}=-\sum_{i\in I_t}(y_i-\bx_i^{\top}\btheta^{(t-1)})\bx_i/|I_t|.$ For variance estimation and statistical inference, the full sample estimator $\widehat{\btheta}_n$ has an asymptotic variance of $\bV=\sigma^2\bSigma^{-1}$. Building on Theorem \ref{thm:dp_sgd} and leveraging the specific structure of linear regression, we propose estimating $\bSigma$ and $\sigma^2$ as follows:
	\begin{equation*}
		\widehat{\bSigma}=\frac{1}{n}\sum_{i=1}^{n}\bx_i\bx_i^{\top}\quad\text{and}\quad\widehat{\sigma}^2=\frac{1}{n}\sum_{i=1}^{n}(y_i-\bx_i^{\top}\bar{\btheta}_T)^2,
	\end{equation*}
	respectively. The estimated variance is then constructed by $\widehat{\bV}=\widehat{\bSigma}^{-1}\widehat{\sigma}^2$.
	
	\begin{assumption}[Design conditions for ordinary linear regression]
		\label{ass:linear_regression}
		The random variables $\bx_i$ and $y_i$ have bounded supports, $\|\bx_i\|_{\infty}\leq C_x$ and $|y_i|\leq C_y$, respectively. Assume that $\btheta^*\in\Theta$ and that $\Theta$ is $\ell_2$-bounded, i.e., $\sup_{\btheta\in\Theta}\|\btheta\|_2\le C_0$.
	\end{assumption}
	
	The conditions outlined in Assumption \ref{ass:linear_regression} are commonly adopted in the differential privacy literature; see, for example, \cite{cai2021cost, sheffet2017differentially}. The bounded support conditions in Assumption \ref{ass:linear_regression} can be relaxed to sub-Gaussian distribution assumptions. Specifically, when $\bx$ follows sub-Gaussian distributions, the large deviation theorem implies that for $C_x=C_y=\sqrt{C\log(n)}$, where $C$ is a positive constant, the conditions in Assumption \ref{ass:linear_regression} hold with probability approaching $1$.
	
	Next, we examine the sensitivity parameters of the proposed method, which are central to ensuring differential privacy. We assume that the variables $\{(\bz_i,y_i)\}_{i=1}^{n}$ satisfy the design conditions specified in Assumption \ref{ass:linear_regression}. For gradient evaluations at parameters satisfying $\|\btheta\|_2\le C_0$, the $\ell_2$ sensitivity is bounded by $\Delta_g=2(C_y+\sqrt{p}C_xC_0)\sqrt{p}C_x$. For variance estimation at a fixed parameter satisfying the same bound, the $\ell_2$ sensitivities of $\widehat{\bSigma}$ and $\widehat{\sigma}^2$ are bounded by $2pC_x^2/n$ and $2(C_y+\sqrt{p}C_xC_0)^2/n$, respectively. Applying these bounded-domain sensitivity bounds requires a mechanism that enforces the parameter bound; for example, for projected linear-regression updates.
	
	\subsection{Logistic Regression}
	In this section, we consider the problem of estimating parameters in logistic regression for binary classification. The binary response $y\in\{0,1\}$, given the covariate $\bx$, follows the probabilistic model:
	\begin{equation*}
		\Pr(y=1\mid\bx)=\{1+\exp(-\bx^{\top}\btheta)\}^{-1}\quad\text{and}\quad\Pr(y=0\mid\bx)=1-\Pr(y=1\mid\bx), 
	\end{equation*}
	where $\btheta\in\mathbb{R}^p$ is the unknown parameter. The sample loss function of maximum likelihood estimator of $\btheta$ with $n$ i.i.d. samples $\{\bz_i\}_{i=1}^{n}=\{\bx_i,y_i\}_{i=1}^{n}$ is $l(\bz_i;\btheta)=-[y_i\bx_i^{\top}\btheta-\log\{1+\exp(\bx^{\top}\btheta)\}]$. The corresponding population function $\mathcal{L}(\btheta)=\mathbb{E}\{l(\bz_i;\btheta)\}$ satisfies the strong convexity condition by assuming the smallest eigenvalue of the Hessian matrix $\mathbb{E}[\exp(\bx^{\top}\btheta)\{1+\exp(\bx^{\top}\btheta)\}^{-2}\bx\bx^{\top}]$ is positive uniformly over $\btheta\in\Theta$.
	
	For the DP-SGD estimation of $\btheta^*$, the gradient function in Algorithm \ref{alg:dp-sgd} at $t$-th iteration is $\boldsymbol{g}^{(t)}=\sum_{i\in I_t}[\{1+\exp(-\bx_i^{\top}\btheta^{(t-1)})\}^{-1}-y_i]\bx_i/|I_t|$. The full sample estimator $\widehat{\btheta}_n$ has the asymptotic variance $\bV=\{\nabla^2 \mathcal{L}(\btheta^*)\}^{-1}$, which is equal to the asymptotic variance of the DP-SGD estimator $\bar{\btheta}_{T}$ under the conditions of Theorem \ref{thm:dp_sgd} when $T/n\to\infty$ and $\sigma_1=O(1)$. Following Proposition \ref{prop:privacy}, it suffices to estimate the matrix $\bA=\bV^{-1}$ by adding appropriate noise to the sample estimator $\widehat{\bA}=\sum_{i=1}^{n}\exp(\bx_i^{\top}\bar{\btheta}_T)\{1+\exp(\bx_i^{\top}\bar{\btheta}_T)\}^{-2}\bx_i\bx_i^{\top}/n$. 
	
	We assume the covariates $\bx_i\in\mathbb{R}^p$ have bounded support, $\|\bx_i\|_{\infty}\leq C_x$. The loss is globally convex and satisfies Assumption \ref{ass:innovation}(i) with $B(\bZ)=\|\bx\|_2$. Its Hessian has a bounded derivative under this support condition, so Assumption \ref{ass:smooth_var} also holds. For DP-SGD, the $\ell_2$ sensitivity of the gradient is $\Delta_g=2\sqrt{p}C_x$. For variance estimation, the $\ell_2$ sensitivity of $\widehat{\bA}$ is bounded by $pC_x^2/(2n)$. By Proposition \ref{prop:privacy} and Theorem \ref{theorem:cover_plugin}, the proposed method provides valid confidence intervals within the differentially private framework.
	
	\section{Numerical Studies}
	
	\subsection{Simulations}
	This section illustrates the empirical performance of the plug-in and the random scaling method for DP-SGD inference through numerical simulations. We consider linear and logistic regression, where each component of the true parameter ($\btheta^*$) is independently sampled from a uniform distribution with support $[0,1]$ for the linear regression and $[0,1/2]$ for the logistic regression. For both models, the covariates ($\bx\in\mathbb{R}^3$) are i.i.d. samples from $\mathcal{N}(\boldsymbol{0}, \bSigma)$. We consider two covariance structures: the identity structure, where $\bSigma = \bI_d$, and the Toeplitz structure, where $\Sigma_{i,j} = 0.5^{|i-j|}$. For linear regression, the noise term $e_i$ is assumed to be independent and follows $\mathcal{N}(0, 1)$. 
	
	For all experimental settings, the total number of iterations is fixed at $T=n^2$. The step size parameter $\alpha$ is set to $0.501$, which is consistent with the choice in \cite{chen2020statistical}. The performance of the estimators is evaluated based on two metrics: the average coverage rate and the average interval length for each coefficient. The nominal coverage probability is set to $95\%$. We adopt the $\mu$-GDP framework and fix the privacy parameter at $\mu=2$ for estimation. We allocate an additional $2$-GDP privacy budget to the joint covariance release for variance estimation. 
	
	We compare the DP-SGD estimator using: (1) the plug-in inference method with finite-sample correction (\ref{eq:plug_correct}), denoted as plug-in, and (2) the random scaling method with finite-sample correction (\ref{eq:rs_correct}), denoted as Random-Scaling. In addition, we include the oracle full-sample estimator as a benchmark to quantify the efficiency loss due to privacy protection. All results are averaged over 1000 independent trials.
	
	\begin{figure}[ht]
		\centering
		\includegraphics[width=\linewidth]{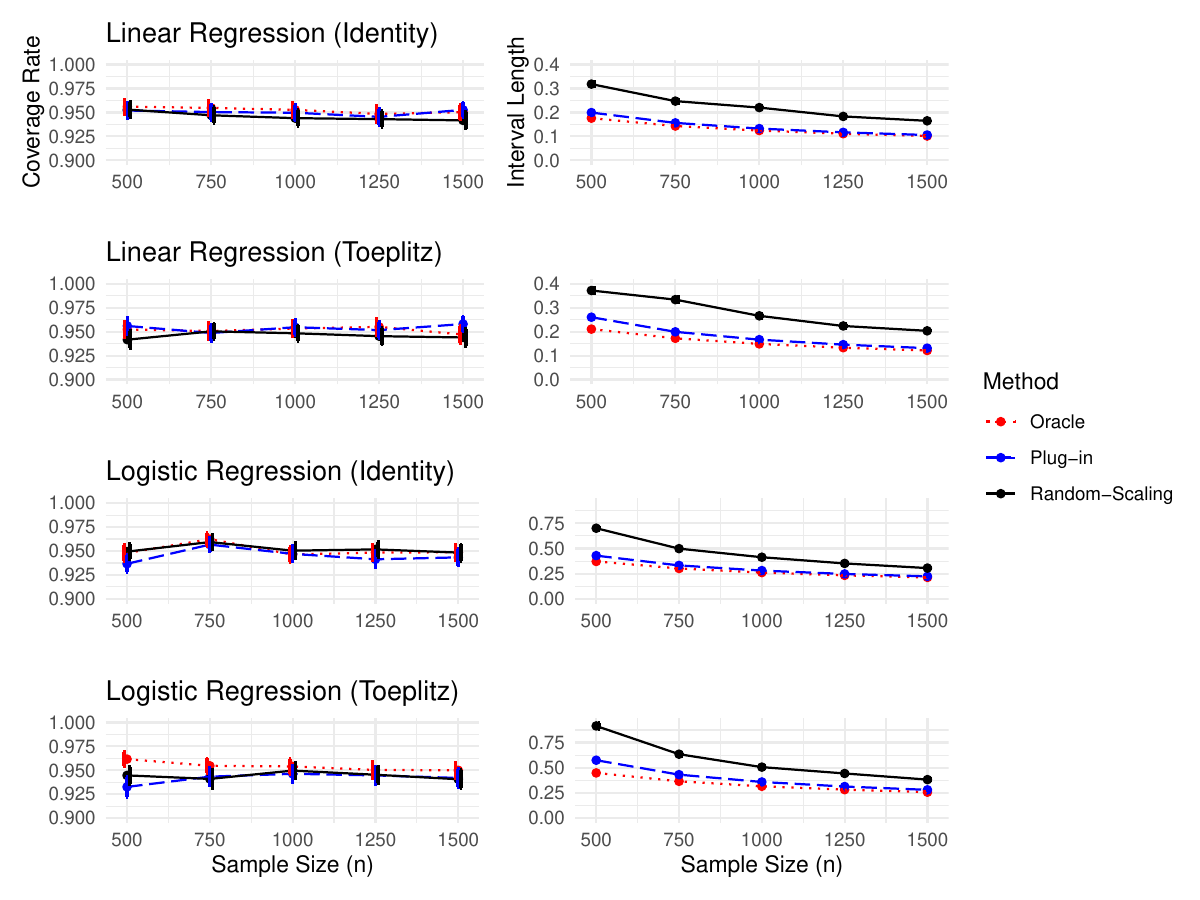}
		\caption{Average coverage rates (left column) and interval lengths (right column) for three methods across four regression models. The oracle full-sample estimator, plug-in method, and random scaling method are compared with increasing sample sizes from 500 to 1500. }
		\label{fig:simulation} 
	\end{figure}
	
	From Figure \ref{fig:simulation}, we observe that both the plug-in method and the random scaling method perform well. The average interval length of the plug-in method closely matches that of the full-sample estimator; however, its average coverage rate falls slightly below the nominal level for small sample sizes. In contrast, the random-scaling method achieves nominal coverage rates for small sample sizes, albeit with much wider confidence intervals compared to the full-sample estimator. As the sample size increases, the difference between the plug-in and oracle methods becomes negligible. The random scaling method also achieves nominal coverage rates as the sample size increases and yields slightly wider confidence intervals than the plug-in method. This phenomenon is consistent with theoretical expectations and aligns with the numerical findings reported in \cite{lee2022fast}.
	
	\subsection{Comparison between DP-SGD and DP-GD}
	
\begin{figure}[ht]
		\centering
		\includegraphics[width=\linewidth]{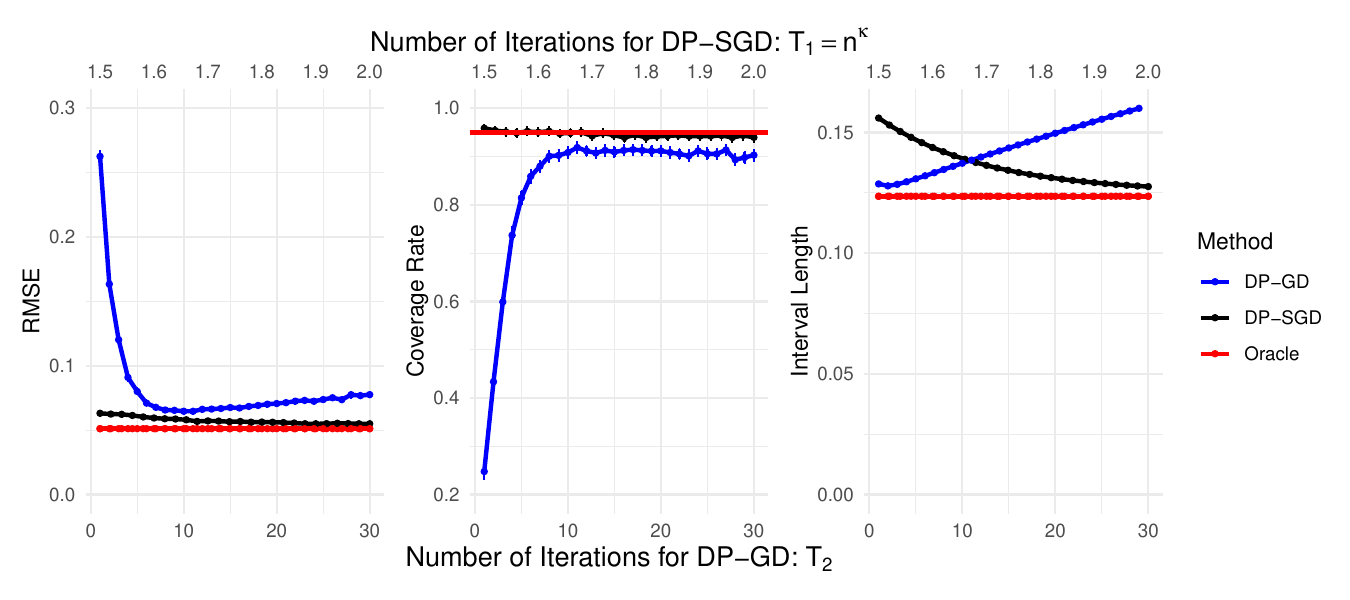}
		\caption{The mean Euclidean error, coverage rate, and confidence interval length of DP-SGD and DP-GD when the number of iterations $T_1$ and $T_2$ increases. Because $T_1=n^{\kappa}$ and $T_2$ differ in order, we use the top axis for $\kappa$ in $T_1$ and use the bottom axis for $T_2$.}
		\label{fig:gd.vs.sgd}
	\end{figure}

	In this section, we compare the numerical performance between DP-SGD and DP-GD, highlighting the stability of DP-SGD in terms of both estimation error and the coverage rate of the plug-in method with finite-sample correction. We use a linear regression model with a sample size of $1000$ as the illustrative example. The results are presented in Figure \ref{fig:gd.vs.sgd}. Since the number of iterations for DP-GD and DP-SGD differ in magnitude, we use dual horizontal axes to plot both methods on the same graph. The top axis represents $\kappa$ for the number of DP-SGD iterations, defined as $T_1 = n^\kappa$, where $\kappa$ ranges from 1.5 to 2. The bottom axis indicates the number of DP-GD iterations, $T_2$, ranging from 0 to 30.
	
	In terms of mean Euclidean error, DP-SGD outperforms DP-GD, exhibiting a lower mean Euclidean error that is less sensitive to the number of iterations. The observed decrease in DP-SGD's mean Euclidean error is driven by the tightest privacy bound under $\mu$-GDP when $T = n^2$. Regarding empirical coverage rates, DP-GD consistently falls well below the nominal rate across all iterations, whereas DP-SGD maintains valid empirical coverage. This disparity stems from the finite-sample correction in DP-SGD, which incorporates a theoretical guarantee accounting for all noise introduced by privacy mechanisms throughout the optimization process. In contrast, DP-GD's finite-sample correction only considers noise added at the final iteration, leading to significant deviations from the nominal coverage rate when the added noise is not negligible. Additionally, the confidence interval length of DP-SGD decreases as $T_1$ increases, while the interval length of DP-GD widens as $T_2$ increases. All three figures consistently illustrate the superior performance of DP-SGD over DP-GD.

	\subsection{Real Data Example}
	In this section, we apply the proposed methods to analyze data from the Student Achievement and Retention (STAR) Project, a randomized trial conducted by \cite{angrist2009incentives} to evaluate the impact of academic support services and incentives on the academic performance of first-year college students. This trial was conducted at a satellite campus of a large Canadian university to investigate the effects of academic support services and financial incentives on the performance of first-year students. The study targeted incoming students in September 2005, excluding those in the top quartile of high school GPA.
	
	Students in the treatment group were offered academic support services (SSP), financial incentives (SFP), and a combination of services and incentives (SFSP). Among $n=1399$ students, participation required signing a consent form; students who did not consent were ineligible for the services or incentives. The covariates include a three-dimensional indicator variable for assignment to the SSP, SFP, and SFSP treatment groups, respectively, as well as sex and high school group as auxiliary variables. The outcome variable is the first-year GPA. Following the approach suggested by \cite{angrist2009incentives}, we consider a linear model and focus on estimating the effects of the three treatments.
	
	The results are presented in Table \ref{tab:real}. The standard error of the random-scaling method is omitted in the table because the random-scaling method directly yields a confidence interval but does not provide a consistent estimator of the standard error. The interval length of the plug-in method is slightly wider than that of the oracle estimator, reflecting the finite-sample correction. Both the plug-in and random scaling methods indicate that the combination of services and financial incentives offered under SFSP had a greater impact on academic performance than services or fellowships alone. This finding is consistent with the results of the oracle estimator and the conclusions drawn by \cite{angrist2009incentives}.

	\begin{table}[ht]
		\centering
		\setlength{\extrarowheight}{0pt} 
		\begin{tabular}{|l|l|l|l|l|}
			\hline
			Variable              & Method    & Estimate          & Standard Error & $p$-value \\ \hline
			\multirow{3}{*}{SSP}  & Oracle    &  0.008& 0.063&0.895 \\ \cline{2-5} 
			& Plug-in & \multirow{2}{*}{0.025} &  0.088 &    0.773     \\ \cline{2-2} \cline{4-5} 
			& RS      &                   &      \texttimes          &   0.656      \\ \hline
			\multirow{3}{*}{SFP}  & Oracle    &      -0.040             &     0.062           &    0.516     \\ \cline{2-5} 
			& Plug-in & \multirow{2}{*}{-0.034} &  0.0851              &   0.686      \\ \cline{2-2} \cline{4-5} 
			& RS      &                   &    \texttimes           &    0.370     \\ \hline
			\multirow{3}{*}{SFSP} & Oracle    &     0.143              &    0.080            &   0.076      \\ \cline{2-5} 
			& Plug-in & \multirow{2}{*}{0.196} &     0.118           &     0.095    \\ \cline{2-2} \cline{4-5} 
			& RS      &                   &     \texttimes          &     0.015    \\ \hline
		\end{tabular}
		\caption{Regression estimates of SSP, SFP, and SFSP on first-year GPA using the oracle method and the two DP-SGD inference methods.}% The $p$-values correspond to the two-sided hypothesis test evaluating whether the coefficients are different from zero.}
	\label{tab:real}
\end{table}

\section{Conclusion and Discussion}

We establish asymptotic normality for SGD under randomized subsampling, with variance inflation determined by batch size and iteration count. For DP-SGD, the limiting distribution separates into independent statistical, subsampling, and privacy components. We develop plug-in and random scaling confidence intervals with asymptotically valid coverage under privacy constraints and evaluate their finite-sample performance numerically. Future work includes extending the analysis beyond smooth losses and developing inference for differentially private zeroth-order methods, such as the Kiefer--Wolfowitz algorithm. Non-convex losses and high-dimensional covariates pose further challenges: multiple local minima and large parameter spaces complicate the characterization of the limiting distribution and generalization error of DP-SGD.

\section*{Data availability statement}

The data in Section 6.3 are openly available in the
Angrist Data Archive at
\url{https://economics.mit.edu/people/faculty/josh-angrist/angrist-data-archive}.

\section*{Disclosure Statement}

The authors report there are no competing interests to declare. 

\bibliographystyle{refstyle}
\bibliography{reference}

@article{polyak1992acceleration,
  title={Acceleration of stochastic approximation by averaging},
  author={Polyak, Boris T and Juditsky, Anatoli B},
  journal={SIAM journal on control and optimization},
  volume={30},
  number={4},
  pages={838--855},
  year={1992},
  publisher={SIAM}
}

@article{opacus,
	title={Opacus: {U}ser-Friendly Differential Privacy Library in {PyTorch}},
	author={Ashkan Yousefpour and Igor Shilov and Alexandre Sablayrolles and Davide Testuggine and Karthik Prasad and Mani Malek and John Nguyen and Sayan Ghosh and Akash Bharadwaj and Jessica Zhao and Graham Cormode and Ilya Mironov},
	journal={arXiv preprint arXiv:2109.12298},
	year={2021}
}

@misc{tensorflow2020,
	author = {TensorFlow},
	title = {TensorFlow (Version 2.0.0)},
	year = {2020},
	publisher = {Google},
	note = {\url{https://www.tensorflow.org/}}
}

@article{altschuler2022privacy,
  title={Privacy of noisy stochastic gradient descent: More iterations without more privacy loss},
  author={Altschuler, Jason and Talwar, Kunal},
  journal={Advances in Neural Information Processing Systems},
  volume={35},
  pages={3788--3800},
  year={2022}
}

@techreport{ruppert1988efficient,
	title={Efficient estimations from a slowly convergent Robbins-Monro process},
	author={Ruppert, David},
	year={1988},
	institution={Cornell University Operations Research and Industrial Engineering}
}

@article{han2024online,
	title={Online inference with debiased stochastic gradient descent},
	author={Han, Ruijian and Luo, Lan and Lin, Yuanyuan and Huang, Jian},
	journal={Biometrika},
	volume={111},
	number={1},
	pages={93--108},
	year={2024},
	publisher={Oxford University Press}
}

@article{chen2020statistical,
  title={Statistical inference for model parameters in stochastic gradient descent},
  author={Chen, Xi and Lee, Jason D and Tong, Xin T and Zhang, Yichen},
  journal={Annals of Statistics},
  volume={48},
  number={1},
  pages={251--273},
  year={2020},
  publisher={Institute of Mathematical Statistics}
}

@book{fuller2011sampling,
  title={Sampling statistics},
  author={Fuller, Wayne A},
  year={2011},
  publisher={John Wiley \& Sons}
}

@book{goodfellow2016deep,
  title={Deep learning},
  author={Goodfellow, Ian and Bengio, Yoshua and Courville, Aaron and Bengio, Yoshua},
  volume={1},
  number={2},
  year={2016},
  publisher={MIT press Cambridge}
}

@article{chen2022online,
  title={Online statistical inference for contextual bandits via stochastic gradient descent},
  author={Chen, Xi and Lai, Zehua and Li, He and Zhang, Yichen},
  journal={arXiv preprint arXiv:2212.14883},
  year={2022}
}

@inproceedings{lee2022fast,
  title={Fast and robust online inference with stochastic gradient descent via random scaling},
  author={Lee, Sokbae and Liao, Yuan and Seo, Myung Hwan and Shin, Youngki},
  booktitle={Proceedings of the AAAI Conference on Artificial Intelligence},
  volume={36},
  pages={7381--7389},
  year={2022}
}

@inproceedings{abadi2016deep,
  title={Deep learning with differential privacy},
  author={Abadi, Martin and Chu, Andy and Goodfellow, Ian and McMahan, H Brendan and Mironov, Ilya and Talwar, Kunal and Zhang, Li},
  booktitle={Proceedings of the 2016 ACM SIGSAC conference on computer and communications security},
  pages={308--318},
  year={2016}
}

@article{beimel2014bounds,
  title={Bounds on the sample complexity for private learning and private data release},
  author={Beimel, Amos and Brenner, Hai and Kasiviswanathan, Shiva Prasad and Nissim, Kobbi},
  journal={Machine learning},
  volume={94},
  pages={401--437},
  year={2014},
  publisher={Springer}
}

@inproceedings{bassily2014private,
  title={Private empirical risk minimization: Efficient algorithms and tight error bounds},
  author={Bassily, Raef and Smith, Adam and Thakurta, Abhradeep},
  booktitle={2014 IEEE 55th annual symposium on foundations of computer science},
  pages={464--473},
  year={2014},
  organization={IEEE}
}

@article{dong2022gaussian,
  title={Gaussian differential privacy},
  author={Dong, Jinshuo and Roth, Aaron and Su, Weijie J},
  journal={Journal of the Royal Statistical Society Series B: Statistical Methodology},
  volume={84},
  number={1},
  pages={3--37},
  year={2022},
  publisher={Oxford University Press}
}

@article{bu2020deep,
  title={Deep learning with Gaussian differential privacy},
  author={Bu, Zhiqi and Dong, Jinshuo and Long, Qi and Su, Weijie J},
  journal={Harvard data science review},
  volume={2020},
  number={23},
  pages={10--1162},
  year={2020},
  publisher={NIH Public Access}
}

@article{mcleish1978extended,
  title={An extended martingale invariance principle},
  author={McLeish, DL},
  journal={The Annals of Probability},
  pages={144--150},
  year={1978},
  publisher={JSTOR}
}

@article{abadir1997two,
  title={Two mixed normal densities from cointegration analysis},
  author={Abadir, Karim M and Paruolo, Paolo},
  journal={Econometrica: Journal of the Econometric Society},
  pages={671--680},
  year={1997},
  publisher={JSTOR}
}

@article{cai2023private,
  title={Differentially Private Estimation and Inference in High-Dimensional Regression with {FDR} Control},
  author={Cai, Zhanrui and Li, Sai and Xia, Xintao and Zhang, Linjun},
  journal={Journal of Machine Learning Research},
  volume={27},
  number={81},
  pages={1--54},
  year={2026}
}

@article{cai2021cost,
  title={The cost of privacy: Optimal rates of convergence for parameter estimation with differential privacy},
  author={Cai, T Tony and Wang, Yichen and Zhang, Linjun},
  journal={The Annals of Statistics},
  volume={49},
  number={5},
  pages={2825--2850},
  year={2021},
  publisher={Institute of Mathematical Statistics}
}

@article{avella2021differentially,
	title={Differentially private inference via noisy optimization},
	author={Avella-Medina, Marco and Bradshaw, Casey and Loh, Po-Ling},
	journal={The Annals of Statistics},
	note = {forthcoming},
	year={2023}
}

@inproceedings{dwork2006calibrating,
  title={Calibrating noise to sensitivity in private data analysis},
  author={Dwork, Cynthia and McSherry, Frank and Nissim, Kobbi and Smith, Adam},
  booktitle={Theory of Cryptography: Third Theory of Cryptography Conference, TCC 2006, New York, NY, USA, March 4-7, 2006. Proceedings 3},
  pages={265--284},
  year={2006},
  organization={Springer}
}

@article{dwork2014algorithmic,
  title={The algorithmic foundations of differential privacy},
  author={Dwork, Cynthia and Roth, Aaron and others},
  journal={Foundations and Trends{\textregistered} in Theoretical Computer Science},
  volume={9},
  number={3--4},
  pages={211--407},
  year={2014},
  publisher={Now Publishers, Inc.}
}

@article{xia2023adaptive,
  title={Adaptive False Discovery Rate Control with Privacy Guarantee},
  author={Xia, Xintao and Cai, Zhanrui},
  journal={Journal of Machine Learning Research},
  year={2023},
  volume={24},
  number={252},
  pages={1--35},
}

@article{cai2026knockoff,
  title={Knockoff inference under privacy constraints},
  author={Cai, Zhanrui and Fan, Yingying and Gao, Lan},
  journal={Journal of the Royal Statistical Society Series B: Statistical Methodology},
  year={2026},
  eid={qkag120},
  doi={10.1093/jrsssb/qkag120},
}

@article{xia2026privateSIR,
  title={Differentially Private Sliced Inverse Regression: Minimax Optimality and Algorithm},
  author={Xia, Xintao and Zhang, Linjun and Cai, Zhanrui},
  journal={Journal of the American Statistical Association},
  year={2026},
  volume={121},
  number={554},
  pages={1116--1127},
  doi={10.1080/01621459.2025.2555059},
}

@article{chourasia2021differential,
  title={Differential privacy dynamics of langevin diffusion and noisy gradient descent},
  author={Chourasia, Rishav and Ye, Jiayuan and Shokri, Reza},
  journal={Advances in Neural Information Processing Systems},
  volume={34},
  pages={14771--14781},
  year={2021}
}

@article{balle2018privacy,
	title={Privacy amplification by subsampling: Tight analyses via couplings and divergences},
	author={Balle, Borja and Barthe, Gilles and Gaboardi, Marco},
	journal={Advances in neural information processing systems},
	volume={31},
	year={2018}
}

@book{van2000asymptotic,
  title={Asymptotic statistics},
  author={Van der Vaart, Aad W},
  volume={3},
  year={2000},
  publisher={Cambridge university press}
}

@article{horvitz1952generalization,
  title={A generalization of sampling without replacement from a finite universe},
  author={Horvitz, Daniel G and Thompson, Donovan J},
  journal={Journal of the American statistical Association},
  volume={47},
  number={260},
  pages={663--685},
  year={1952},
  publisher={Taylor \& Francis}}

@article{vaswani2017attention,
  title={Attention is all you need},
  author={Vaswani, Ashish and Shazeer, Noam and Parmar, Niki and Uszkoreit, Jakob and Jones, Llion and Gomez, Aidan N and Kaiser, {\L}ukasz and Polosukhin, Illia},
  journal={Advances in neural information processing systems},
  volume={30},
  year={2017}
}

@inproceedings{song2013stochastic,
  title={Stochastic gradient descent with differentially private updates},
  author={Song, Shuang and Chaudhuri, Kamalika and Sarwate, Anand D},
  booktitle={2013 IEEE global conference on signal and information processing},
  pages={245--248},
  year={2013},
  organization={IEEE}
}

@inproceedings{mironov2017renyi,
  title={R{\'e}nyi differential privacy},
  author={Mironov, Ilya},
  booktitle={2017 IEEE 30th computer security foundations symposium (CSF)},
  pages={263--275},
  year={2017},
  organization={IEEE}
}

@inproceedings{wang2019subsampled,
  title={Subsampled r{\'e}nyi differential privacy and analytical moments accountant},
  author={Wang, Yu-Xiang and Balle, Borja and Kasiviswanathan, Shiva Prasad},
  booktitle={The 22nd international conference on artificial intelligence and statistics},
  pages={1226--1235},
  year={2019},
  organization={PMLR}
}

@article{birrell2024differentially,
  title={Differentially Private Stochastic Gradient Descent with Fixed-Size Minibatches: Tighter RDP Guarantees with or without Replacement},
  author={Birrell, Jeremiah and Ebrahimi, Reza and Behnia, Rouzbeh and Pacheco, Jason},
  journal={arXiv preprint arXiv:2408.10456},
  year={2024}
}

@inproceedings{balle2020hypothesis,
  title={Hypothesis testing interpretations and renyi differential privacy},
  author={Balle, Borja and Barthe, Gilles and Gaboardi, Marco and Hsu, Justin and Sato, Tetsuya},
  booktitle={International Conference on Artificial Intelligence and Statistics},
  pages={2496--2506},
  year={2020},
  organization={PMLR}
}

@inproceedings{sheffet2017differentially,
  title={Differentially private ordinary least squares},
  author={Sheffet, Or},
  booktitle={International Conference on Machine Learning},
  pages={3105--3114},
  year={2017},
  organization={PMLR}
}

@article{su2023higrad,
  title={Higrad: Uncertainty quantification for online learning and stochastic approximation},
  author={Su, Weijie J and Zhu, Yuancheng},
  journal={Journal of Machine Learning Research},
  volume={24},
  number={124},
  pages={1--53},
  year={2023}
}

@article{alber1998projected,
  title={On the projected subgradient method for nonsmooth convex optimization in a Hilbert space},
  author={Alber, Ya I and Iusem, Alfredo N and Solodov, Mikhail V},
  journal={Mathematical Programming},
  volume={81},
  pages={23--35},
  year={1998},
  publisher={Springer}
}

@inproceedings{gurbuzbalaban2021heavy,
  title={The heavy-tail phenomenon in SGD},
  author={Gurbuzbalaban, Mert and Simsekli, Umut and Zhu, Lingjiong},
  booktitle={International Conference on Machine Learning},
  pages={3964--3975},
  year={2021},
  organization={PMLR}
}

@inproceedings{qian2021understanding,
  title={Understanding gradient clipping in incremental gradient methods},
  author={Qian, Jiang and Wu, Yuren and Zhuang, Bojin and Wang, Shaojun and Xiao, Jing},
  booktitle={International Conference on Artificial Intelligence and Statistics},
  pages={1504--1512},
  year={2021},
  organization={PMLR}
}

@inproceedings{zhang2019gradient,
  title={Why Gradient Clipping Accelerates Training: A Theoretical Justification for Adaptivity},
  author={Zhang, Jingzhao and He, Tianxing and Sra, Suvrit and Jadbabaie, Ali},
  booktitle={International Conference on Learning Representations},
  year={2020}
}

@inproceedings{koloskova2023revisiting,
  title={Revisiting Gradient Clipping: Stochastic bias and tight convergence guarantees},
  author={Koloskova, Anastasia and Hendrikx, Hadrien and Stich, Sebastian U},
  booktitle={International Conference on Machine Learning},
  pages={17343--17363},
  year={2023},
  organization={PMLR}
}

@article{angrist2009incentives,
  title={Incentives and services for college achievement: Evidence from a randomized trial},
  author={Angrist, Joshua and Lang, Daniel and Oreopoulos, Philip},
  journal={American Economic Journal: Applied Economics},
  volume={1},
  number={1},
  pages={136--163},
  year={2009},
  publisher={American Economic Association}
}

@book{durrett2019probability,
  title={Probability: theory and examples},
  author={Durrett, Rick},
  volume={49},
  year={2019},
  publisher={Cambridge university press}
}

@article{wang2022differentially,
  title={Differentially private SGD with non-smooth losses},
  author={Wang, Puyu and Lei, Yunwen and Ying, Yiming and Zhang, Hai},
  journal={Applied and Computational Harmonic Analysis},
  volume={56},
  pages={306--336},
  year={2022},
  publisher={Elsevier}
}

@article{bassily2020stability,
  title={Stability of stochastic gradient descent on nonsmooth convex losses},
  author={Bassily, Raef and Feldman, Vitaly and Guzm{\'a}n, Crist{\'o}bal and Talwar, Kunal},
  journal={Advances in Neural Information Processing Systems},
  volume={33},
  pages={4381--4391},
  year={2020}
}

% Supplementary material (merged from supp.tex).
\clearpage
\appendix
\spacingset{1}
\setcounter{figure}{0}
\renewcommand{\thefigure}{S\arabic{figure}}
\setcounter{table}{0}
\renewcommand{\thetable}{S\arabic{table}}
\counterwithin{definition}{section}
\counterwithin{theorem}{section}
\counterwithin{algorithm}{section}
\section*{Supplementary Material}

\section{Definitions of Differential Privacy}

	First proposed by \cite{dwork2006calibrating}, differential privacy (DP) provides a rigorous mathematical foundation to preserve privacy. It enables meaningful data analysis while protecting individual privacy. Here we give the formal definition.
	
	\begin{definition}[$(\varepsilon,\delta)$-Differential Privacy \citep{dwork2006calibrating}]
		\label{def:dp}
		A randomized algorithm $M:\calD\to\mR$ is $(\varepsilon,\delta)$-DP for $\varepsilon,\delta>0$, if for every pair of neighboring data sets $\calD,\calD'$ that differ by one individual datum and every measurable set $\mathcal{S}\subset\mR$ with respect to $M(\cdot)$,
		\begin{equation*}
			\mathbb{P}[M(\calD)\in\mathcal{S}]\leq e^{\varepsilon}\mathbb{P}[M(\calD')\in\mathcal{S}]+\delta,
		\end{equation*}
		where the probability measure $\mathbb{P}$ is induced by the randomness of $M$ only.
	\end{definition}
	
	%In addition to the classic definition, other popular frameworks for privacy are also proposed, such as R\'enyi Differential Privacy (RDP) \citep{mironov2017renyi} and Gaussian Differential Privacy (GDP) \citep{dong2022gaussian}. 
	
	The Rényi Differential Privacy (RDP) framework proposed by \cite{mironov2017renyi} is more amenable to DP-SGD analysis and can be easily converted to DP guarantees. To define RDP, we first define the Rényi divergence.
	
	\begin{definition}[R\'enyi Divergence]
		\label{def:rd}
		The R\'enyi divergence between two probability measures $\mu$ and $\nu$ of order $\alpha \in (1, \infty)$ is
		\[
		\mathcal{D}_\alpha (\mu \parallel \nu) =
		\frac{1}{\alpha - 1}
		\log \int \left(
		\frac{\mu(x)}{\nu(x)}
		\right)^\alpha \nu(x) \, dx,
		\]
		if $\mu \ll \nu$. The R\'enyi divergences of order $\alpha \in \{1, \infty\}$ are defined by continuity.
		
	\end{definition}
	
	\begin{definition}[R\'enyi Differential Privacy \citep{mironov2017renyi}]
		
		A randomized algorithm $M:\calD\to\mR$ is $(\alpha,\varepsilon)$-RDP for $\alpha\geq 1$ and $\varepsilon>0$, if for every pair of neighboring data sets $\calD$ and $\calD'$, the probability measure, induced by the randomness of $M$, satisfies
		\begin{equation*}
			\mathcal{D}_\alpha (M(\calD) \parallel M(\calD'))\leq\varepsilon.
		\end{equation*}
	\end{definition}
	
	\textit{Gaussian Differential Privacy} (GDP) is another novel framework for differential privacy proposed by \cite{dong2022gaussian}. It utilizes trade-off functions from hypothesis testing to quantify the indistinguishability between two neighboring data sets. Consider two neighboring data sets $\calD$ and $\calD'$. Let $\phi$ represent a rejection rule for the hypothesis test, where the null hypothesis $H_0$ asserts that the underlying dataset is $\calD$, and the alternative hypothesis $H_1$ asserts that the underlying dataset is $\calD'$. The type I error rate and the type II error rate associated with $\phi$ are denoted by $\alpha_{\phi}$ and $\beta_{\phi}$, respectively.
	\begin{definition}[Gaussian Differential Privacy \citep{dong2022gaussian}]
		A randomized algorithm $M:\calD\to\mR$ is $\mu$-GDP for $\mu>0$, if for every pair of neighboring data sets $\calD$ and $\calD'$,
		\begin{equation*}
			\inf_{\phi}\{\beta_{\phi}:\alpha_{\phi}\leq\alpha\}\geq\Phi(\Phi^{-1}(1-\alpha)-\mu)
		\end{equation*}
		for all $0<\alpha<1$, where the infimum is taken over all rejection rules based on $M(\cdot)$. 
	\end{definition}

	%\section{Theoretical Details of Examples}   
	
	\section{Proof of Example 1 and Further Discussion of Sampling Schemes}
	
	\begin{proof}
		We theoretically prove the statements in Example 1. 
		
		We first consider the SGD estimator with cyclic draws. Note that for the loss function $l(\theta,x)=(x-\theta)^2$ and stepsize $\eta_i=1/2$, the gradient of the loss function is $\nabla l(\theta,x)=-2(x-\theta)$ and the update procedure given data point $x_{i+1}$ and the current estimator $\theta^{(i)}$ can be written as 
		$$\theta^{(i+1)}=\theta^{(i)}-\tfrac12\,2(\theta^{(i)}-x_{i+1})=x_{i+1}.$$
		We show that the expression can be simplified by mathematical induction. Given the initial value $\theta^{(0)}=0$, the first iteration yields $\theta^{(1)}=x_1$. Thus, we conclude that $\bar{\theta}_{cyc}=\sum_{i=1}^{n}x_i/n$ by mathematical induction. By the assumption that $x_1,\dots,x_n$ are i.i.d. and follow normal distributions, we have $\sqrt{n}(\bar{\theta}_{cyc}-\mu)\sim N(0,\sigma^2)$ by the property of Gaussian distributions.
		
		Then, we show the estimator based on the randomized rule. We employ certain techniques in \textit{Survey Sampling} to achieve the desired results corresponding to random draws. We first introduce some notation. Let $\mathcal{F}_n=\{x_i\}_{i=1}^{n}$ denote the finite population and $x_{(i)}$ be the selected sample in $i$-th draw for $i=1,\dots,n$. Note that we have $\bar{\theta}_{ran}=\sum_{i=1}^{n}x_{(i)}/n$ by similar arguments in the SGD with cyclic draws, where $\{x_{(i)}\}_{i=1}^{n}$ are i.i.d. \textit{simple random samples with replacement} from the finite population $\mathcal{F}_n$. Conditional on the finite population $\mathcal{F}_n$, we have 
		$$\mathbb{E}(x_{(i)}\mid\mathcal{F}_n)=\sum_{i=1}^{n}x_i/n=\bar{\theta}_{cyc}$$ 
		and 
		$$\mathbb{V}(x_{(i)}\mid\mathcal{F}_n)=\sum_{i=1}^{n}(x_i-\bar{\theta}_{cyc})^2/n.$$
		By the CLT in survey sampling \citep{fuller2011sampling}, 
		$$\sqrt{n}(\bar{\theta}_{ran}-\bar{\theta}_{cyc})\mid\mathcal{F}_n\stackrel{d}{\to} N(0,\sum_{i=1}^{n}(x_i-\bar{\theta}_{cyc})^2/n)\stackrel{d}{\to} N(0,\sigma^2),$$
		where we use Slutsky's theorem in the last expression. By the previous results for $\bar{\theta}_{cyc}$ and Theorem 1.3.6 in \cite{fuller2011sampling}, we have 
		$$\sqrt{n}(\bar{\theta}_{ran}-\mu)=\sqrt{n}(\bar{\theta}_{ran}-\bar{\theta}_{cyc})+\sqrt{n}(\bar{\theta}_{cyc}-\mu)\stackrel{d}{\to}N(0,2\sigma^2).$$
		
		The result of random draws can be further generalized to the case where a random draw of batch size $m$ is used, and the total number of iterations $T$ satisfies $T/n=k$ for a positive constant $k>0$. We introduce some additional notation. We use $\{x_{(i,j)}\}_{j=1}^{m}$ to denote the selected samples in $i$-th draw, where $\{x_{(i,j)}\}_{j=1}^{m}$ is a \textit{simple random sample without replacement} of size $m$ from the population $\mathcal{F}_n=\{x_i\}_{i=1}^{n}$. Note that in this case, given the data points $\{x_{(i+1,j)}\}_{j=1}^{m}$ and the current estimator $\theta^{(i)}$, the averaged gradient of the loss function is 
		$$\frac{1}{m}\sum_{j=1}^{m}\nabla l(\theta^{(i)},x_{(i+1,j)})=-\frac{2}{m}\sum_{j=1}^{m}(x_{(i+1,j)}-\theta^{(i)})=-2(\frac{1}{m}\sum_{j=1}^{m}x_{(i+1,j)}-\theta^{(i)}),$$
		and the update procedure can be written as 
		$$\theta^{(i+1)}=\theta^{(i)}-\frac12\frac2m\sum_{j=1}^{m}(\theta^{(i)}-x_{(i+1,j)})=\frac1m\sum_{j=1}^{m}x_{(i+1,j)}.$$
		Averaging these iterates, we obtain $$\bar{\theta}_{ran}=\frac{1}{T}\sum_{i=1}^{T}\frac{1}{m}\sum_{j=1}^{m}x_{(i,j)},$$ 
		where the samples $\{\{x_{(i,j)}\}_{j=1}^{m}\}_{i=1}^{T}$ can be viewed as $T$ i.i.d. replications. Thus, it remains to consider the samples in each iteration. Note that in each iteration, the value $m^{-1}\sum_{j=1}^{m}x_{(i,j)}$ is the \textit{Horvitz–Thompson estimator} \citep{horvitz1952generalization} of the population mean. By Corollary 1.2.2.1 of \cite{fuller2011sampling}, for each iteration, we have 
		$$\mathbb{E}\bigg(\frac{1}{m}\sum_{j=1}^{m}x_{(i,j)}\mid\mathcal{F}_n\bigg)=\frac{1}{n}\sum_{i=1}^{n}x_i=\bar{\theta}_{cyc}$$
		and
		$$\mathbb{V}\bigg(\frac{1}{m}\sum_{j=1}^{m}x_{(i,j)}\mid\mathcal{F}_n\bigg)=\bigg(1-\frac{m}{n}\bigg)\frac{\sum_{i=1}^{n}(x_i-\bar{\theta}_{cyc})^2}{(n-1)m},$$
		where $1-m/n$ is called the \textit{finite sample correction} and $\sum_{i=1}^{n}(x_i-\bar{\theta}_{cyc})^2/n$ is the finite population variance. When $m$ is bounded by a constant, by the CLT in survey sampling, we have
		$$\sqrt{n}(\bar{\theta}_{ran}-\bar{\theta}_{cyc})\mid\mathcal{F}_n\stackrel{d}{\to} N\bigg(0,\frac{1}{k m}\cdot\sum_{i=1}^{n}(x_i-\bar{\theta}_{cyc})^2/n\bigg).$$
		By the previous results for $\bar{\theta}_{cyc}$ and Theorem 1.3.6 in \cite{fuller2011sampling}, we have 
		$$\sqrt{n}(\bar{\theta}_{ran}-\mu)=\sqrt{n}(\bar{\theta}_{ran}-\bar{\theta}_{cyc})+\sqrt{n}(\bar{\theta}_{cyc}-\mu)\stackrel{d}{\to}N(0,\{1+1/(km)\}\sigma^2).$$
		
		Furthermore, when $k\to\infty$, we have
		\begin{equation*}
			\mathbb{E}[\bar{\theta}_{ran}\mid\mathcal{F}_n]=\frac{1}{n}\sum_{i=1}^{n}x_i
		\end{equation*}
		and
		\begin{equation*}
			\mathbb{V}[\bar{\theta}_{ran}\mid\mathcal{F}_n]=\frac{n-m}{n-1}\frac{1}{nmk}\bigg\{\frac{1}{n}\sum_{i=1}^{n}(x_i-\frac{1}{n}\sum_{i=1}^{n}x_i)^2\bigg\}.
		\end{equation*}
		Then, by Markov's inequality,
		\begin{equation*}
			\begin{split}
				\mathbb{P}\bigg(|\bar{\theta}_{ran}-\frac{1}{n}\sum_{i=1}^{n}x_i|\geq\frac{\epsilon}{n^{1/2}}\bigg)=&\mathbb{E}\bigg[\mathbb{P}\bigg(|\bar{\theta}_{ran}-\frac{1}{n}\sum_{i=1}^{n}x_i|\geq\frac{\epsilon}{n^{1/2}}\bigg)\bigg|\mathcal{F}_n\bigg]\\
				\leq&\frac{n}{\epsilon^2}\mathbb{E}\bigg[\frac{1}{nmk}\bigg\{\frac{1}{n}\sum_{i=1}^{n}(x_i-\frac{1}{n}\sum_{i=1}^{n}x_i)^2\bigg\}\bigg]=o(1).
			\end{split}
		\end{equation*}
		Thus, by Slutsky's theorem,
		\begin{equation*}
			\begin{split}
				\sqrt{n}(\bar{\theta}_{ran}-\mu)&=\sqrt{n}\bigg\{\bigg(\bar{\theta}_{ran}-\frac{1}{n}\sum_{i=1}^{n}x_i\bigg)+\bigg(\frac{1}{n}\sum_{i=1}^{n}x_i-\mu\bigg)\bigg\}\\
				&\stackrel{d}{\to}\sqrt{n}\bigg(\frac{1}{n}\sum_{i=1}^{n}x_i-\mu\bigg)\stackrel{d}{\to}N(0,\sigma^2).
			\end{split}
		\end{equation*}
		\hfill \( \blacksquare \)
	\end{proof}

	\subsection{Sampling Schemes}
	
	The random subsample $I_t$ in each iteration of DP-SGD can be generated using two different sampling procedures: simple random sampling without replacement (SRSWOR) or Poisson subsampling. Poisson subsampling is utilized by \cite{abadi2016deep}. To align with standard practices in SGD, we adopt SRSWOR for both theoretical analysis and experiments. For Poisson subsampling, gradients are divided by the expected batch size $m$, with an empty minibatch contributing zero. Compared to Poisson subsampling, SRSWOR can be more efficient. In terms of mean estimation, the difference in efficiency is asymptotically negligible when the population means converge to zero. Thus, the conclusions of Theorems 1 and 2 hold for Poisson subsampling. 
	
	To illustrate the idea, we theoretically analyze the efficiency gain of simple random sampling without replacement (SRSWOR) compared to Poisson sampling following Example 1. Let $Z$ be a random variable with mean $\mu$ and variance $\sigma^2$ and $\mathcal{F}_n=\{Z_i\}_{i=1}^{n}$ be i.i.d. samples from the distribution $Z$. 
	
	For SRSWOR, we consider the estimator
	\begin{equation*}
		\hat{\mu}_{SRSWOR}=\sum_{i=1}^{n}z_iI_i/m,
	\end{equation*}
	where a sample of size $m$ is selected by SRSWOR, the sample indicator $I_i=1$ if element-$i$ is selected and $I_i=0$ otherwise. By Corollary 1.2.2.1 of \cite{fuller2011sampling}, we have
	\begin{equation*}
		\mathbb{E}(\hat{\mu}_{SRSWOR}\mid\mathcal{F}_n)=\sum_{i=1}^{n}z_i/n:=\mu_n,
	\end{equation*}
	where $\mu_n$ is the finite sample mean of $\{z_i\}_{i=1}^{n}$, and
	\begin{equation*}
		\begin{split}
			\mathbb{V}(\hat{\mu}_{SRSWOR}\mid\mathcal{F}_n)&=\bigg(1-\frac{m}{n}\bigg)\frac{1}{m}\sum_{i=1}^{n}\frac{(z_i-\mu_n)^2}{n-1}=\frac{n-m}{n-1}\frac{\sigma^2_n}{m}\\
			&\stackrel{p}{\to}\bigg(1-\frac{m}{n}\bigg)\frac{\sigma^2}{m},
		\end{split}
	\end{equation*}
	where $\sigma^2_n=n^{-1}\sum_{i=1}^n(z_i-\mu_n)^2$ is the finite population variance.
	
	For Poisson sampling, we consider the estimator
	\begin{equation*}
		\hat{\mu}_{Poi}=\sum_{i=1}^{n}z_iI_i/m,
	\end{equation*}
	where $\{I_i\}_{i=1}^{n}$ are i.i.d. samples of a Bernoulli distribution with $\mathbb{P}(I_i=1)=m/n$. The expected sample size of Poisson sampling is 
	\begin{equation*}
		\mathbb{E}(\sum_{i=1}^{n}I_i)=n\cdot\frac{m}{n}=m.
	\end{equation*}
	Furthermore, we have
	\begin{equation*}
		\mathbb{E}(\hat{\mu}_{Poi}\mid\mathcal{F}_n)=\sum_{i=1}^{n}z_i/n:=\mu_n,
	\end{equation*}
	and
	\begin{equation*}
		\begin{split}
			\mathbb{V}(\hat{\mu}_{Poi}\mid\mathcal{F}_n)&=\frac{1}{m^2}\sum_{i=1}^{n}\mathbb{V}(I_i)z_i^2=\frac{1}{m^2}\sum_{i=1}^{n}\frac{m}{n}\bigg(1-\frac{m}{n}\bigg)z_i^2\\
			&=\bigg(1-\frac{m}{n}\bigg)\frac{1}{m}\sum_{i=1}^{n}\frac{z_i^2}{n}=\bigg(1-\frac{m}{n}\bigg)\frac{1}{m}\sum_{i=1}^{n}\bigg\{\frac{(z_i-\mu_n)^2}{n}+\frac{\mu_n^2}{n}\bigg\}\\
			&=\bigg(1-\frac{m}{n}\bigg)\frac{\sigma_n^2+\mu_n^2}{m}\stackrel{p}{\to}\bigg(1-\frac{m}{n}\bigg)\frac{\sigma^2+\mu^2}{m}.
		\end{split}
	\end{equation*}
	
	For $m<n$ and $\sigma_n^2>0$, the relative efficiency of the estimator $\hat{\mu}_{SRSWOR}$ compared to the estimator $\hat{\mu}_{Poi}$ is
	\begin{equation*}
		\frac{\mathbb{V}(\hat{\mu}_{Poi}\mid\mathcal{F}_n)}{\mathbb{V}(\hat{\mu}_{SRSWOR}\mid\mathcal{F}_n)}=\frac{n-1}{n}\frac{\sigma_n^2+\mu_n^2}{\sigma_n^2}\stackrel{p}{\to}1+\frac{\mu^2}{\sigma^2}.
	\end{equation*}
	When $\mu\neq 0$, the estimator $\hat{\mu}_{SRSWOR}$ is asymptotically more efficient than the estimator $\hat{\mu}_{Poi}$. When $\mu= 0$, the estimator $\hat{\mu}_{SRSWOR}$ is asymptotically as efficient as the estimator $\hat{\mu}_{Poi}$. In the analysis of SGD, realizations of random variables, $\{z_i\}_{i=1}^{n}$, represent the gradients of the loss function evaluated at the true parameter $\btheta^*$. According to the first-order optimality condition, the samples $\{z_i\}_{i=1}^{n}$ can be viewed as being generated from a distribution with $\mu=0$. Thus, it can be shown that the asymptotic distribution of SGD under Poisson sampling is equal to that of SGD under SRSWOR.
	
	\section{Proofs of Theorems}
	\subsection{Proof of Theorem 1}
	\begin{proof}
		The proof focuses on the random sampling nature of each iteration. We first consider the problem of solving a linear equation, then extend it to the general convex minimization problem by conditioning on the observed data. Throughout, the initial value is fixed, and averages are taken over iterates $1,\ldots,T$, as in the main paper.
		
		First, let $\bA$ be positive definite and $\bA\btheta^*=\boldsymbol b$. We use the innovation sign convention of the general-loss argument below:
		\begin{equation}\label{eq:sgd_update}
			\btheta^{(t)}=\btheta^{(t-1)}-\eta_t(\bA\btheta^{(t-1)}-\boldsymbol b)+\eta_t\bar{\bxi}^{(t)},
		\end{equation}
		where $\bar{\bxi}^{(t)}=m^{-1}\sum_{i=1}^n I_i^{(t)}\bxi_{(i)}$, the $\bxi_{(i)}$ are i.i.d. mean-zero vectors with covariance $\bS$, and the size-$m$ SRSWOR samples are independent across iterations conditional on $\mathcal F_n=\{\bxi_{(i)}\}_{i=1}^n$. Reversing the innovation sign leaves the covariance calculation unchanged. Write $\bDelta^{(t)}=\btheta^{(t)}-\btheta^*$. To make the averaging representation exact, define
		\[
		\begin{aligned}
			P_{t,j}(\bA)&=\prod_{s=j+1}^{t}(\bI-\eta_s\bA),\qquad P_{j,j}(\bA)=\bI,\\
			B_{j,T}(\bA)&=\eta_j\sum_{t=j}^{T}P_{t,j}(\bA),\qquad
			C_T(\bA)=\frac1T\sum_{t=1}^{T}P_{t,0}(\bA).
		\end{aligned}
		\]
		Unrolling the recursion gives
		\begin{equation}\label{eq:linear_pr}
			\sqrt n\,\bar{\bDelta}^{(T)}=
			\underbrace{\sqrt n C_T(\bA)\bDelta^{(0)}}_{I^{(1)}}+
			\underbrace{\frac{\sqrt n}{T}\sum_{t=1}^{T}\bA^{-1}\bar{\bxi}^{(t)}}_{I^{(2)}}+
			\underbrace{\frac{\sqrt n}{T}\sum_{t=1}^{T}\boldsymbol w_t^T\bar{\bxi}^{(t)}}_{I^{(3)}},
		\end{equation}
		where $\boldsymbol w_t^T=B_{t,T}(\bA)-\bA^{-1}$. The averaging-weight bounds in Lemma 2 of \cite{polyak1992acceleration} and Lemma \ref{lem:linear_algebra} give
		\[
		\|C_T(\bA)\|=O(T^{-1}),\qquad
		\max_{t\le T}\|\boldsymbol w_t^T\|\le K,\qquad
		\frac1T\sum_{t=1}^{T}\|\boldsymbol w_t^T\|=O(T^{\alpha-1}).
		\]
		These bounds also hold uniformly when the eigenvalues of $\bA$ range over a fixed compact subset of $(0,\infty)$. Thus $I^{(1)}=O(\sqrt n/T)=o(1)$ for $T/n\ge c_0>0$.
		
		The draws are conditionally independent, but are not independent unconditionally. Accordingly, with $\bar{\bxi}=n^{-1}\sum_{i=1}^n\bxi_{(i)}$, split the third term as
		\[
		I^{(3)}=\frac{\sqrt n}{T}\sum_{t=1}^{T}\boldsymbol w_t^T(\bar{\bxi}^{(t)}-\bar{\bxi})
		+\frac{\sqrt n}{T}\left(\sum_{t=1}^{T}\boldsymbol w_t^T\right)\bar{\bxi}.
		\]
		Conditional second moments bound the first term, while $\sqrt n\bar{\bxi}=O_p(1)$ bounds the second. Consequently,
		\[
		I^{(3)}=O_p\!\left(\sqrt{n/T}\,T^{(\alpha-1)/2}+T^{\alpha-1}\right)=o_p(1).
		\]
		For the leading term,
		\[
		I^{(2)}=\sqrt n\,\bA^{-1}\bar{\bxi}
		+\frac{\sqrt{n/T}}{\sqrt T}\sum_{t=1}^{T}\bA^{-1}(\bar{\bxi}^{(t)}-\bar{\bxi}).
		\]
		If $\bS_n=n^{-1}\sum_{i=1}^n(\bxi_{(i)}-\bar{\bxi})(\bxi_{(i)}-\bar{\bxi})^{\top}$, the exact SRSWOR covariance is
		\[
		\operatorname{Var}(\bar{\bxi}^{(t)}\mid\mathcal F_n)=\frac{n-m}{m(n-1)}\bS_n.
		\]
		The conditional triangular-array CLT therefore gives
		\[
		T^{-1/2}\sum_{t=1}^{T}\bA^{-1}(\bar{\bxi}^{(t)}-\bar{\bxi})
		\ \Big|\ \mathcal F_n\ \stackrel d\longrightarrow\ N(0,\bA^{-1}\bS\bA^{-1}/m).
		\]
		The ordinary CLT gives covariance $\bA^{-1}\bS\bA^{-1}$ for the empirical-mean component $\sqrt n\bA^{-1}\bar{\bxi}$. The conditional characteristic functions converge in probability to the nonrandom Gaussian characteristic function. Hence the two limiting components are independent. This proves the claimed covariance $\{1+(km)^{-1}\}\bA^{-1}\bS\bA^{-1}$ for the linear recursion.
		
		We now consider the general loss and condition on $\mathcal F_n=\{\bz_i\}_{i=1}^n$. Write $\mathcal L_n(\btheta)=n^{-1}\sum_i l(\bz_i;\btheta)$ and $H_n(\btheta)=\nabla^2\mathcal L_n(\btheta)$. Assumption 5 and the law of large numbers imply
		\[
		H_n(\btheta^*)\stackrel p\longrightarrow\bA,\qquad
		\sup_{\|\btheta-\btheta^*\|\le r}\|H_n(\btheta)-\bA\|
		\le\|H_n(\btheta^*)-\bA\|+L_2r.
		\]
		Choose $r>0$ sufficiently small that this neighborhood lies in the interior of $\Theta$. Since $\bA$ is positive definite, with probability tending to one the empirical loss has uniformly positive curvature on this neighborhood. Moreover, $\nabla\mathcal L_n(\btheta^*)=O_p(n^{-1/2})$. Convexity then gives a unique interior empirical minimizer $\btheta_n^*$ in this neighborhood, and Taylor expansion gives
		\begin{equation}\label{eq:empirical_root_expansion}
			\sqrt n(\btheta_n^*-\btheta^*)=-\bA^{-1}\frac1{\sqrt n}\sum_{i=1}^n\nabla l(\bz_i;\btheta^*)+o_p(1),
			\qquad \widehat{\bA}:=H_n(\btheta_n^*)\stackrel p\longrightarrow\bA.
		\end{equation}
		In particular, $\btheta_n^*-\btheta^*=O_p(n^{-1/2})$. All conditional bounds below are established on data events of probability tending to one on which the empirical averages of $B(\bz_i)^2$ and $\|\nabla^2l(\bz_i;\btheta^*)\|^2$, the norm of $\btheta_n^*$, and the relevant local curvature constants are bounded by deterministic constants.
		
		Set $\bDelta^{(t)}=\btheta^{(t)}-\btheta_n^*$ and define
		\[
		\bxi_{(i)}'(\bDelta)=\nabla\mathcal L_n(\btheta_n^*+\bDelta)-\nabla l(\bz_i;\btheta_n^*+\bDelta),
		\qquad \bar{\bxi}^{(t)\prime}=\frac1m\sum_i I_i^{(t)}\bxi_{(i)}'(\bDelta^{(t-1)}).
		\]
		Then $\bar{\bxi}^{(t)\prime}$ is conditionally centered given the past, and
		\[
		\bDelta^{(t)}=\bDelta^{(t-1)}-\eta_t\nabla\mathcal L_n(\btheta_n^*+\bDelta^{(t-1)})+\eta_t\bar{\bxi}^{(t)\prime}.
		\]
		The global linear-growth condition in Assumption 2(i) and convexity yield, for $X_t=\|\bDelta^{(t)}\|^2$,
		\begin{equation}\label{eq:conditional_distance_drift}
			E(X_t\mid\mathscr F_{t-1},\mathcal F_n)
			\le (1+K\eta_t^2)X_{t-1}-2\eta_t\langle\bDelta^{(t-1)},\nabla\mathcal L_n(\btheta^{(t-1)})\rangle+K\eta_t^2
			\le (1+K\eta_t^2)X_{t-1}+K\eta_t^2.
		\end{equation}
		Thus $(X_t+1)/\prod_{s=1}^t(1+K\eta_s^2)$ is a nonnegative conditional supermartingale. Its maximal inequality gives
		\begin{equation}\label{eq:uniform_localization}
			P\left(\sup_{t\ge0}\|\bDelta^{(t)}\|>R\mid\mathcal F_n\right)
			\le \frac{\exp(K\sum_{t\ge1}\eta_t^2)(X_0+1)}{R^2+1}.
		\end{equation}
		This bound is uniform in $n$ on the stated data events and hence applies to the growing horizon $T=T_n$.
		
		For fixed $R$, local empirical strong convexity and global convexity imply
		\[
		\langle\bDelta,\nabla\mathcal L_n(\btheta_n^*+\bDelta)\rangle\ge c_R\|\bDelta\|^2,\qquad \|\bDelta\|\le R,
		\]
		for a deterministic $c_R>0$. Indeed, along any ray from $\btheta_n^*$ the directional derivative is nondecreasing; curvature on a fixed small ball therefore supplies this inequality on any fixed larger ball. Define $\tau_R=\inf\{t\ge0:\|\bDelta^{(t)}\|>R\}$ and $u_t=E[X_t1\{\tau_R>t\}\mid\mathcal F_n]$. By \eqref{eq:conditional_distance_drift},
		\[
		u_t\le(1-2c_R\eta_t+K\eta_t^2)u_{t-1}+K\eta_t^2.
		\]
		For $\eta_t=\eta t^{-\alpha}$, $\alpha\in(1/2,1)$, this recursion gives $u_t\le K_R\eta_t$ after finitely many initial steps. This is a stopped conditional moment bound; no unconditional moment claim is made for the empirical minimizer.
		
		Retain the linear comparison process
		\[
		\bDelta_1^{(0)}=\bDelta^{(0)},\qquad
		\bDelta_1^{(t)}=(\bI-\eta_t\widehat{\bA})\bDelta_1^{(t-1)}+\eta_t\bar{\bxi}^{(t)\prime}.
		\]
		Put $r_n(\bDelta)=\nabla\mathcal L_n(\btheta_n^*+\bDelta)-\widehat{\bA}\bDelta$. Assumption 5 is anchored at $\btheta^*$, so in its neighborhood the appropriate Taylor bound is
		\[
		\|r_n(\bDelta)\|\le\frac{L_2}{2}\|\bDelta\|^2+2L_2\|\btheta_n^*-\btheta^*\|\,\|\bDelta\|.
		\]
		On the rest of a fixed localization ball, the linear-growth bound supplies the same type of bound with a larger constant $K_R$, because there $\|\bDelta\|$ is bounded away from zero. The initial values of the two processes agree, so their difference has no initial-condition term. The exact averaging representation above gives
		\[
		\sqrt n(\bar{\bDelta}_1^{(T)}-\bar{\bDelta}^{(T)})
		=\frac{\sqrt n}{T}\sum_{t=1}^{T}B_{t,T}(\widehat{\bA})r_n(\bDelta^{(t-1)}).
		\]
		Using the stopped moment bound, conditional Markov's inequality, and \eqref{eq:empirical_root_expansion}, the norm of this expression, on $\{\tau_R>T\}$, is
		\[
		O_p\left(\sqrt n\,T^{-\alpha}+T^{-\alpha/2}+\sqrt n/T\right)=o_p(1).
		\]
		Here finite initial terms contribute $O_p(\sqrt n/T)$. Letting $R\to\infty$ in \eqref{eq:uniform_localization} removes the stopping.
		
		It remains to freeze the innovations at the empirical minimizer. Write
		\[
		\bxi_{(i)}'(\bDelta^{(t-1)})=\bxi_{(i)}'(0)+\boldsymbol\zeta_{(i)}^{(t)},\qquad
		\bar{\boldsymbol\zeta}^{(t)}=\frac1m\sum_i I_i^{(t)}\boldsymbol\zeta_{(i)}^{(t)}.
		\]
		Both the conditional mean of $\bar{\boldsymbol\zeta}^{(t)}$ and the finite-population mean of $\bxi_{(i)}'(0)=-\nabla l(\bz_i;\btheta_n^*)$ are zero. Local Hessian bounds from Assumption 5, and the growth bound away from that neighborhood, give
		\[
		E(\|\bar{\boldsymbol\zeta}^{(t)}\|^2\mid\mathscr F_{t-1},\mathcal F_n)
		\le K_R\|\bDelta^{(t-1)}\|^2\quad\text{when }\|\bDelta^{(t-1)}\|\le R.
		\]
		The predictably stopped increments $\bar{\boldsymbol\zeta}^{(t)}1\{\tau_R>t-1\}$ remain martingale differences. Conditional orthogonality and the stopped moment bound imply
		\[
		E\left[\left\|\frac{\sqrt n}{T}\sum_{t=1}^{T}B_{t,T}(\widehat{\bA})\bar{\boldsymbol\zeta}^{(t)}1\{\tau_R>t-1\}\right\|^2\,\middle|\,\mathcal F_n\right]
		\le \frac{K_R n}{T^2}\left(1+\sum_{t=1}^{T-1}\eta_t\right)=o(1).
		\]
		After removing the stopping, the centered version of the linear-weight calculation therefore yields
		\begin{equation}\label{eq:empirical_centered_average}
			\sqrt n(\bar{\btheta}_T-\btheta_n^*)
			=\frac{\sqrt n}{T}\sum_{t=1}^{T}\bA^{-1}\bar{\bxi}^{(t)\prime}(0)+o_p(1).
		\end{equation}
		Here $\bar{\bxi}^{(t)\prime}(0)=m^{-1}\sum_i I_i^{(t)}\bxi_{(i)}'(0)$; replacing $\widehat{\bA}$ by $\bA$ is justified by \eqref{eq:empirical_root_expansion}. The empirical covariance of these frozen scores converges to $\bS$, by Assumptions 2 and 5. Their conditional Lindeberg condition can be checked directly: with $g_i(\btheta)=\nabla l(\bz_i;\btheta)$, Assumption 5 gives
		\[
		\max_{i\le n}\|g_i(\btheta_n^*)\|
		\le \max_{i\le n}\|g_i(\btheta^*)\|
		+\|\btheta_n^*-\btheta^*\|\left\{\max_{i\le n}\|\nabla^2l(\bz_i;\btheta^*)\|+L_2r\right\}.
		\]
		Finite second moments, $\btheta_n^*-\btheta^*=O_p(n^{-1/2})$, and $T\ge c_0n$ imply that this maximum divided by $\sqrt T$ is $o_p(1)$. Since a minibatch mean is bounded by the maximum score norm, the conditional Lindeberg condition follows. Thus the leading term in \eqref{eq:empirical_centered_average} contributes only the sampling covariance $\bA^{-1}\bS\bA^{-1}/(km)$. Its conditional characteristic function has a nonrandom limit, so it is asymptotically independent of the empirical-root fluctuation in \eqref{eq:empirical_root_expansion}. Finally, use the exact identity
		\[
		\sqrt n(\bar{\btheta}_T-\btheta^*)=
		\sqrt n(\btheta_n^*-\btheta^*)+\sqrt n(\bar{\btheta}_T-\btheta_n^*).
		\]
		This proves Theorem 1. If bounded $m=m_n$ or $k=k_n\ge c_0$ varies, the preceding argument applies along every subsequence on which $m$ is constant and $k$ converges in $[c_0,\infty]$, with $1/\infty=0$; the stated normalization then gives the same limiting law along every such subsequence.
		\hfill $\blacksquare$
	\end{proof}
	
	\subsection{Proof of Theorem 2}
	\begin{proof}
		We retain the linear-process calculation, with an independent privacy innovation $\boldsymbol n_t\sim N(0,\sigma_1^2\bI)$:
		\[
		\bDelta^{(t)}=(\bI-\eta_t\bA)\bDelta^{(t-1)}+\eta_t(\bar{\bxi}^{(t)}+\boldsymbol n_t).
		\]
		The sign of $\boldsymbol n_t$ is immaterial and can be chosen to match Algorithm 1. With the exact averaging weights defined in the proof of Theorem 1,
		\[
		\sqrt n\,\bar{\bDelta}^{(T)}=
		\underbrace{\sqrt n C_T(\bA)\bDelta^{(0)}}_{I^{(1)}}+
		\underbrace{\frac{\sqrt n}{T}\sum_{t=1}^{T}\bA^{-1}(\bar{\bxi}^{(t)}+\boldsymbol n_t)}_{I^{(2)}}+
		\underbrace{\frac{\sqrt n}{T}\sum_{t=1}^{T}\boldsymbol w_t^T(\bar{\bxi}^{(t)}+\boldsymbol n_t)}_{I^{(3)}}.
		\]
		As before, $I^{(1)}=O(\sqrt n/T)=o(1)$. Splitting each sampling innovation into its finite-population mean and its conditionally centered part gives
		\[
		I^{(3)}=O_p\left(\sqrt{n/T}(1+\sigma_1)T^{(\alpha-1)/2}+T^{\alpha-1}\right)=o_p(1).
		\]
		The leading term separates into
		\[
		I^{(2)}=\sqrt n\bA^{-1}\bar{\bxi}
		+\frac{\sqrt n}{T}\sum_{t=1}^{T}\bA^{-1}(\bar{\bxi}^{(t)}-\bar{\bxi})
		+\frac{\sqrt n}{T}\sum_{t=1}^{T}\bA^{-1}\boldsymbol n_t.
		\]
		The first two components were treated in Theorem 1; the third is independent of them and is exactly $N(0,\sigma_1^2\bA^{-2}n/T)$. This proves the three-component covariance formula for the linear process.
		
		For the general loss, use the empirical minimizer $\btheta_n^*$, $\widehat{\bA}$, and centered sampling innovations from the proof of Theorem 1. The actual recursion is
		\[
		\bDelta^{(t)}=\bDelta^{(t-1)}-\eta_t\nabla\mathcal L_n(\btheta_n^*+\bDelta^{(t-1)})
		+\eta_t\bar{\bxi}^{(t)\prime}+\eta_t\boldsymbol n_t.
		\]
		The conditional distance inequality \eqref{eq:conditional_distance_drift} now has an additional $p\sigma_1^2\eta_t^2$ term. Since $\sigma_1=O(1)$, the same supermartingale localization is uniform in $n$, and the stopped second-moment recursion gives
		\[
		E[\|\bDelta^{(t)}\|^2 1\{\tau_R>t\}\mid\mathcal F_n]\le K_R(1+\sigma_1^2)\eta_t
		\]
		after finitely many initial steps, in particular, the argument does not infer stability from compactness of the local regularity region.
		
		Define the same comparison process, now including privacy noise,
		\[
		\bDelta_1^{(0)}=\bDelta^{(0)},\qquad
		\bDelta_1^{(t)}=(\bI-\eta_t\widehat{\bA})\bDelta_1^{(t-1)}
		+\eta_t(\bar{\bxi}^{(t)\prime}+\boldsymbol n_t).
		\]
		Coupling the two processes by the same noises cancels their initial-condition and privacy-noise terms in the difference. The anchored Taylor bound and predictable stopping argument from Theorem 1 then apply without change, with constants multiplied by bounded factors depending on $1+\sigma_1^2$. Freezing the sampling innovations at $\btheta_n^*$ and using the centered linear-weight bound yields
		\[
		\sqrt n(\bar{\btheta}_T-\btheta_n^*)
		=\frac{\sqrt n}{T}\sum_{t=1}^{T}\bA^{-1}\bar{\bxi}^{(t)\prime}(0)
		+\frac{\sqrt n}{T}\sum_{t=1}^{T}\bA^{-1}\boldsymbol n_t+o_p(1).
		\]
		Add \eqref{eq:empirical_root_expansion}. Conditional on the data, the frozen sampling sum and the Gaussian sum are independent. The conditional sampling CLT has a nonrandom limiting characteristic function, so their Gaussian limits are also independent of the empirical-root limit. Their respective covariances are
		\[
		\bA^{-1}\bS\bA^{-1},\qquad
		\frac{1}{km}\bA^{-1}\bS\bA^{-1},\qquad
		\frac{\sigma_1^2}{k}\bA^{-2},
		\]
		as asserted. For varying $k,m,\sigma_1$, this is a Gaussian approximation with the displayed sequence of covariance matrices; equivalently, along any subsequence where $k\to k_0\in[c_0,\infty]$, $m\to m_0$, and $\sigma_1\to s_0$, the weak limit has the corresponding limiting covariances. These subsequences exhaust all possibilities because $m$ and $\sigma_1$ are bounded.
		\hfill $\blacksquare$
	\end{proof}
	\subsection{Proof of Corollary 1-3}
	The proof focuses mainly on checking the conditions of the existing DP-SGD framework.
	\begin{proof}[Proof of Corollary 1]
		We first consider the $(\varepsilon,\delta)$-DP framework shown in Corollary 1. To establish the existence of a suitable bounded noise standard deviation, choose any fixed $\sigma_1>0$. The privacy analysis uses the following consequence of Theorem 9 in \cite{wang2019subsampled}.
		\begin{lemma}[Consequence of Theorem 9 in \cite{wang2019subsampled}]
			\label{lem:subsampled_gaussian_dp}
			Under Assumption 3, let $q=m/n$ and fix $m$, $\sigma_1>0$, and an integer $r\geq2$. There exists a finite constant $C_r$, independent of $n$, the iteration, and its history, such that each noisy minibatch gradient in Algorithm 1 is $(r,C_rq^2)$-RDP for every fixed history of preceding noisy gradients.
		\end{lemma}
		Indeed, for each fixed history, Assumption 3 bounds the replacement sensitivity of the mean of $m$ gradients by $\Delta_g/m$. Before subsampling, the Gaussian mechanism therefore has order-$j$ RDP parameter at most $j\Delta_g^2/(2m^2\sigma_1^2)$ for $j>1$, by Proposition 7 of \cite{mironov2017renyi}. In Theorem 9 of \cite{wang2019subsampled}, all terms inside the logarithm beyond $1$ have powers $q^j$ with $2\leq j\leq r$ and finite coefficients independent of $n$ and the history. Using $q^j\leq q^2$ and $\log(1+x)\leq x$ gives the stated bound.
		
		Choose a fixed integer $r\geq2$ such that $\log(1/\delta)/(r-1)<\varepsilon/2$. By adaptive composition (Proposition 1 of \cite{mironov2017renyi}) and Lemma \ref{lem:subsampled_gaussian_dp}, the transcript of noisy gradients is $(r,\rho_T(r))$-RDP with
		\begin{equation*}
			\rho_T(r)\leq C_rTq^2=O(n^{l-2})\longrightarrow0.
		\end{equation*}
		Proposition 3 of \cite{mironov2017renyi} then gives $(\varepsilon,\delta)$-DP for all sufficiently large $n$, because
		\begin{equation*}
			\rho_T(r)+\frac{\log(1/\delta)}{r-1}\leq\varepsilon.
		\end{equation*}
		The estimator $\bar{\boldsymbol{\theta}}_T$ is a function of this transcript and the fixed initialization, so the same guarantee follows by postprocessing.
		
		Furthermore, the chosen $\sigma_1$ is $O(1)$, so the remainder in Theorem 2 is $o_p(1)$. When $T=\lceil n^{l}\rceil$ for $1<l<2$, we have $k=T/n\asymp n^{l-1}\stackrel{n\to\infty}{\to}\infty$. Thus, we have the convergence results: $\boldsymbol{\phi}_{sam}\stackrel{p}{\to}0$ and $\boldsymbol{\phi}_{privacy}\stackrel{p}{\to}0$ by Markov's inequality. Combining all the previous results, we have
		\begin{equation*}
			\sqrt{n}(\bar{\boldsymbol{\theta}}_T-\boldsymbol{\theta}^*)\stackrel{d}{\to}\boldsymbol{\phi}_{stat}\stackrel{d}{=}N(\boldsymbol{0},\boldsymbol{A}^{-1}\boldsymbol{S}\boldsymbol{A}^{-1}),
		\end{equation*}
		where we use Slutsky's theorem. The covariance of the leading privacy-noise component $n^{-1/2}\boldsymbol{\phi}_{privacy}$ is
		\begin{equation*}
			\frac{\sigma_1^2}{nk}\boldsymbol{A}^{-2}
			=\frac{\sigma_1^2}{T}\boldsymbol{A}^{-2}.
		\end{equation*}
		By Markov's inequality, this component is $O_p(T^{-1/2})=o_p(n^{-1/2})$.
		\hfill \( \blacksquare \)
	\end{proof}
	
	\begin{proof}[Proof of Corollary 2]
		The privacy analysis follows Theorem 9 in \cite{wang2019subsampled}, whose following consequence gives a uniform upper bound.
		\begin{lemma}[Consequence of Theorem 9 in \cite{wang2019subsampled}]
			\label{lem:rdp_sgd}
			Fix an integer $r\ge2$ and suppose $\sigma_1\ge\underline\sigma>0$. Write $q=m/n<1$. Conditional on any preceding history, each step of DP-SGD is $(r,\rho_t(r))$-RDP with
			\begin{equation*}
				\rho_t(r)\le2rq^2(e^{4\Delta_g^2/\sigma_1^2}-1)+K_rq^3,
			\end{equation*}
			where $K_r$ depends only on $r,\Delta_g,\underline\sigma$ and is independent of $n,t$ and the history.
		\end{lemma}
		For fixed history, the Gaussian minibatch mechanism has order-$j$ RDP parameter $j\Delta_g^2/(2m^2\sigma_1^2)$. The finite bound in the cited theorem, $\log(1+x)\le x$, and $m\ge1$ give the displayed inequality; its finitely many terms of degree at least three in $q$ are bounded by $K_rq^3$ uniformly under the stated lower bound on $\sigma_1$.
		\begin{lemma}[Proposition 1 in \cite{mironov2017renyi}]
			\label{lem:rdp_com}
			Let \( M = M_T \circ M_{T-1} \circ \cdots \circ M_1 \) be defined in an interactively compositional manner. Then, for any fixed \( \gamma \geq 1 \),
			\[
			\varepsilon_M(\gamma) \leq \sum_{i=1}^T \varepsilon_{M_i}(\gamma).
			\]
		\end{lemma}
		
		By Lemma \ref{lem:rdp_com}, it remains to bound the privacy cost of each step. Put $r=\lceil\gamma\rceil$, $q=m/n$, and $a_n=\varepsilon/(Tq^2)$, and choose $c_2^2\ge\max\{4c_1,32r\}$. The prescribed calibration gives $\sigma_1^2=c_2^2\Delta_g^2/a_n$. Hence $a_n<c_1$ and $\sigma_1\ge c_2\Delta_g/\sqrt{c_1}>0$, while the condition $\sigma_1=O(1)$ inherited from Theorem 2 implies $\liminf_n a_n>0$. Let $b_n=4\Delta_g^2/\sigma_1^2=4a_n/c_2^2\le1$. By Lemma \ref{lem:rdp_sgd} and $e^b-1\le2b$ for $0\le b\le1$,
		\begin{equation*}
			\rho_t(r)\le2rq^2(e^{b_n}-1)+K_rq^3
			\le\frac{16r}{c_2^2}a_nq^2+K_rq^3
			\le\frac{\varepsilon}{2T}+K_rq^3.
		\end{equation*}
		Since $q\to0$ and $\liminf_n a_n>0$, we have $K_rq^3=o(a_nq^2)=o(\varepsilon/T)$. Thus each step is $(r,\varepsilon/T)$-RDP for all sufficiently large $n$. Adaptive composition gives $(r,\varepsilon)$-RDP for the transcript, and monotonicity in the R\'enyi order and postprocessing give $(\gamma,\varepsilon)$-RDP for $\bar{\btheta}_T$.
		
		Furthermore, when $\sigma_1=O(1)$, the bias term is $o_p(1)$ by Theorem 2. When $T=\lceil n^{l}\rceil$ for $1<l<2$, we have $k=T/n\asymp n^{l-1}\stackrel{n\to\infty}{\to}\infty$. Thus, we have the convergence results: $\boldsymbol{\phi}_{sam}\stackrel{p}{\to}0$ and $\boldsymbol{\phi}_{privacy}\stackrel{p}{\to}0$ by Markov's inequality. Combining all the previous results, we have
		\begin{equation*}
			\sqrt{n}(\bar{\boldsymbol{\theta}}_T-\boldsymbol{\theta}^*)\stackrel{d}{\to}\boldsymbol{\phi}_{stat}\stackrel{d}{=}N(\boldsymbol{0},\boldsymbol{A}^{-1}\boldsymbol{S}\boldsymbol{A}^{-1}),
		\end{equation*}
		where we use Slutsky's theorem.
		\hfill \( \blacksquare \)
	\end{proof}
	
	\begin{proof}[Proof of Corollary 3]
		The privacy analysis directly follows from Corollary 5.4 in \cite{dong2022gaussian}, which is restated as Lemma \ref{lem:dong}.
		\begin{lemma}[Corollary 5.4 in \cite{dong2022gaussian}]
			\label{lem:dong}
			If \( m\sqrt{T}/n \to c \), then DP-SGD is asymptotically \(\mu\)-GDP with
			\[
			\mu = \sqrt{2}c \cdot \sqrt{e^{\sigma^{-2}} \cdot \Phi(1.5\sigma^{-1}) + 3\Phi(-0.5\sigma^{-1}) - 2},
			\]
			where \(\Phi(\cdot)\) denotes the cumulative distribution function of the standard normal distribution.
		\end{lemma}
		By Lemma \ref{lem:dong}, the estimator $\bar{\boldsymbol{\theta}}_T$ is asymptotically $\mu$-GDP. For any $c>0$, the condition of Corollary 3 is satisfied when the iteration satisfies $T=(cn/m)^2$. Thus, we have $k=T/n=(c/m)^2n\to\infty$ as $n\to\infty$. When $\mu$ is bounded away from zero, we have $\sigma_1=O(1)$. By Theorem 2, the bias term is $o_p(1)$. Thus, we have the convergence results: $\boldsymbol{\phi}_{sam}\stackrel{p}{\to}0$ and $\boldsymbol{\phi}_{privacy}\stackrel{p}{\to}0$ by Markov's inequality. Combining all the previous results, we have
		\begin{equation*}
			\sqrt{n}(\bar{\boldsymbol{\theta}}_T-\boldsymbol{\theta}^*)\stackrel{d}{\to}\boldsymbol{\phi}_{stat}\stackrel{d}{=}N(\boldsymbol{0},\boldsymbol{A}^{-1}\boldsymbol{S}\boldsymbol{A}^{-1}),
		\end{equation*}
		where we use Slutsky's theorem. We consider the privacy term in more detail. The variance of $n^{-1/2}\boldsymbol{\phi}_{privacy}$ satisfies
		\begin{equation*}
			\frac{\sigma_1^2}{nk}\boldsymbol{A}^{-2}\asymp\frac{\mu^2}{m^2T\mu^2}\asymp\frac{1}{\mu^2 n^2},
		\end{equation*}
		where we treat $\sigma$ as a fixed constant and use the relationship $\mu\asymp c$. By Markov's inequality, the cost of privacy is $O_p(1/(n\mu))$.
		\hfill \( \blacksquare \)
	\end{proof}
	
	Corollaries 1-3 involve specific constants or impose certain conditions on the privacy parameters and the number of iterations. Given a fixed number of iterations $T$ and privacy parameter $(\varepsilon,\delta)$, we outline the implementation details of DP-SGD under the $(\varepsilon,\delta)$-DP framework, where $(\varepsilon,\delta)$ are predefined privacy parameters. \cite{birrell2024differentially} provided tighter RDP guarantees of SRSWOR than the result by \cite{wang2019subsampled}. For finite-sample calibration with $q=m/n<1$, let $\rho_\alpha(\sigma_1)$ denote the full upper bound in Theorem 3.4 of \cite{birrell2024differentially}, including its remainder, with noise multiplier $m\sigma_1/\Delta_g$. To apply this bound under Assumption 3, for each fixed preceding history we center every gradient at $\nabla l(\bz_0;\btheta)$ for a fixed public reference point $\bz_0$ in the data domain. The centered gradients have norms at most $\Delta_g$, and this data-independent translation of the conditional noisy average preserves its privacy loss. In the notation of the cited theorem, the clipping bound is therefore $C=\Delta_g$, and the noise standard deviation of the averaged gradient is $C(m\sigma_1/\Delta_g)/m=\sigma_1$. We choose $\sigma_1$ numerically so that, for some $\alpha>1$,
	\begin{equation*}
		T\rho_\alpha(\sigma_1)+\log\bigg(\frac{\alpha-1}{\alpha}\bigg)
		-\frac{\log\delta+\log\alpha}{\alpha-1}\le\varepsilon.
	\end{equation*}
	Adaptive composition, the conversion lemma below, and postprocessing then give the target $(\varepsilon,\delta)$-DP guarantee. If $m=n$, we instead use the unsubsampled Gaussian bound $\rho_\alpha(\sigma_1)=\alpha\Delta_g^2/(2m^2\sigma_1^2)$ in the same calibration.
	
	\begin{lemma}[Proposition 1 in \cite{mironov2017renyi}]
		\label{lem:rd_compo}
		For any fixed $\gamma>1$, let $M_i$ be $(\gamma,\varepsilon_{M_i})$-RDP for $\varepsilon_{M_i}>0$ and $i=1,\dots,T$. Then, let \( M = M_T \circ M_{T-1} \circ \cdots \circ M_1 \) be defined in an interactively compositional way. Then, for any fixed \( \gamma \geq 1 \), the algorithm $M$ is $(\gamma,\varepsilon_M)$-RDP for
		\[
		\varepsilon_M=\sum_{i=1}^T \varepsilon_{M_i}.
		\]
	\end{lemma}
	Then, we use the transformation of the RDP to DP by the following lemma.
	\begin{lemma}[Theorem 20 in \cite{balle2020hypothesis}]
		If a mechanism \( M \) is \( (\gamma, \varepsilon) \)-RDP for $\gamma>1$, then it is 
		\[
		\bigg( \varepsilon + \log\bigg(\frac{\gamma - 1}{\gamma}\bigg) - \frac{\log \delta + \log \gamma}{\gamma - 1}, \delta \bigg)\text{-DP}
		\]
		for any \( 0 < \delta < 1 \).
	\end{lemma}
	
	\subsection{Proof of Proposition 1}
	\begin{proof}
		For each fixed preceding DP-SGD output $\bar{\btheta}_T$, Assumption 4 bounds the sensitivities of the vectorized covariance queries by $\Delta_A/n$ and $\Delta_S/n$. We apply the following bounds to these conditional queries; adaptive composition combines this additional covariance-release budget with that of DP-SGD. We first provide some necessary lemmas about privacy analysis.
		
		\begin{lemma}[The Gaussian Mechanism by \cite{dwork2014algorithmic}]
			\label{lem:ed_gaussian}
			Let $0<\varepsilon<1$ and $0<\delta<1$. For any algorithm \( f \) mapping a dataset \( X \) to \( \mathbb{R}^d \) such that \( \Delta_2(f) < \infty \), define
			\[
			M(X) := f(X) + (\xi_1, \xi_2, \dots, \xi_d),
			\]
			where \( \xi_1, \xi_2, \dots, \xi_d \) are i.i.d. samples drawn from \( \mathcal{N}(0, 2(\Delta_2(f) / \varepsilon)^2 \log(1.25 / \delta)) \). The mechanism \( M(X) \) achieves \((\varepsilon, \delta)\)-differential privacy.
		\end{lemma}
		
		\begin{lemma}[Composition Theorem in \cite{dwork2014algorithmic}]
			\label{lem:ed_compo}
			Let $M_i$ be $(\varepsilon_i,\delta_i)$-DP for $i=1,\dots,T$. Let \( M = M_T \circ M_{T-1} \circ \cdots \circ M_1 \) be defined in an interactively compositional way. Then, the algorithm $M$ is $(\sum_{i=1}^{T}\varepsilon_i,\sum_{i=1}^{T}\delta_i)$-DP.
		\end{lemma}
		
		By Lemma \ref{lem:ed_gaussian}, the estimates $\widetilde{\bA}$ and $\widetilde{\bS}$ are both $(\varepsilon/2,\delta/2)$-DP. By Lemma \ref{lem:ed_compo}, output $\widetilde{\bA}$ and $\widetilde{\bS}$ is $(\varepsilon,\delta)$-DP.
		
		We then consider the RDP. By the following Lemma \ref{lem:rd_gaussian}, the estimates $\widetilde{\bA}$ and $\widetilde{\bS}$ are $(\gamma,\varepsilon/2)$-RDP, separately, for $\sigma_2=\Delta_An^{-1}(\gamma/\varepsilon)^{1/2}$ and $\sigma_3=\Delta_Sn^{-1}(\gamma/\varepsilon)^{1/2}$. By Lemma \ref{lem:rd_compo}, output $\widetilde{\bA}$ and $\widetilde{\bS}$ is $(\gamma,\varepsilon)$-RDP.
		
		\begin{lemma}[Corollary 3 in \cite{mironov2017renyi}]
			\label{lem:rd_gaussian}
			For any algorithm \( f \) mapping a dataset \( X \) to \( \mathbb{R}^d \) such that \( \Delta_2(f) < \infty \), define
			\[
			M(X) := f(X) + (\xi_1, \xi_2, \dots, \xi_d),
			\]
			where \( \xi_1, \xi_2, \dots, \xi_d \) are i.i.d. samples drawn from \( \mathcal{N}(0, \sigma^2\Delta_2(f)^2 ) \). The mechanism \( M(X) \) achieves \((\gamma, \gamma/(2\sigma^2))\)-RDP.
		\end{lemma}
		
		For the $\mu$-GDP, by the Lemma \ref{lem:gd_gaussian}, the estimates $\widetilde{\bA}$ and $\widetilde{\bS}$ are $\mu/\sqrt{2}$-GDP, separately, for $\sigma_2=\Delta_An^{-1}(\sqrt{2}/\mu)$ and $\sigma_3=\Delta_Sn^{-1}(\sqrt{2}/\mu)$. By Lemma \ref{lem:gd_compo}, output $\widetilde{\bA}$ and $\widetilde{\bS}$ is $\mu$-GDP.
		
		\begin{lemma}[Theorem 2.7 in \cite{dong2022gaussian}]
			\label{lem:gd_gaussian}
			For any algorithm \( f \) mapping a dataset \( X \) to \( \mathbb{R}^d \) such that \( \Delta_2(f) < \infty \), define
			\[
			M(X) := f(X) + (\xi_1, \xi_2, \dots, \xi_d),
			\]
			where \( \xi_1, \xi_2, \dots, \xi_d \) are i.i.d. samples drawn from \( \mathcal{N}(0, \Delta_2(f)^2/\mu^2 ) \). The mechanism \( M(X) \) achieves \(\mu\)-GDP.
		\end{lemma}
		
		\begin{lemma}[Corollary 3.3 in \cite{dong2022gaussian}]
			\label{lem:gd_compo}
			Let $M_i$ be $\mu_i$-GDP for $i=1,\dots,T$. Then, let \( M = M_T \circ M_{T-1} \circ \cdots \circ M_1 \) be defined in an interactively compositional way. Then, the algorithm $M$ is $\sqrt{\sum_{i=1}^{T}\mu_i^2}$-GDP.
		\end{lemma}
		\hfill \( \blacksquare \)
	\end{proof}

	\subsection{Proof of Proposition 2 and Theorem 3}
	\begin{proof}[Proof of Proposition 2]
		We first establish the bounds for the raw noisy releases. Symmetrization cannot increase their operator-norm distances to the symmetric targets $\bA$ and $\bS$, so the same bounds hold after this postprocessing. We first consider the estimation of $\bA$. By the decomposition,
		\begin{equation*}
			\begin{split}
				\widetilde{\bA}-\bA&=\frac{1}{n}\sum_{i=1}^{n}\nabla^2 l(\bz_i;\bar{\btheta}_T)-\bA+\bxi^{\prime}\\
				&=\Big(\frac{1}{n}\sum_{i=1}^{n}\nabla^2 l(\bz_i;\bar{\btheta}_T)-\frac{1}{n}\sum_{i=1}^{n}\nabla^2 l(\bz_i;\btheta^*)\Big)+\Big(\frac{1}{n}\sum_{i=1}^{n}\nabla^2 l(\bz_i;\btheta^*)-\bA\Big)+\bxi^{\prime}.
			\end{split}
		\end{equation*}
		We consider the three terms one by one. For the first term, we have
		\begin{equation*}
			\begin{split}
				& \|\frac{1}{n}\sum_{i=1}^{n}\nabla^2 l(\bz_i;\bar{\btheta}_T)-\frac{1}{n}\sum_{i=1}^{n}\nabla^2 l(\bz_i;\btheta^*)\|_2\\
				\leq&\frac{1}{n}\sum_{i=1}^{n}\|\nabla^2 l(\bz_i;\bar{\btheta}_T)-\nabla^2 l(\bz_i;\btheta^*)\|_2\leq L_2\|\bar{\btheta}_T-\btheta^*\|_2=O_p(n^{-1/2}+\sigma_1T^{-1/2}),
			\end{split}
		\end{equation*}
		where we use the triangle inequality in the first inequality, we use Assumption 5 in the second inequality, and we use Theorem 2 in the last equality. By the fact $\E\{\nabla^2 l(\bz_i;\btheta^*)\}=\bA$, we have
		\begin{equation*}
			\begin{split}
				\E\|\frac{1}{n}\sum_{i=1}^{n}\nabla^2 l(\bz_i;\btheta^*)-\bA\|_2&\leq[\sum_{j=1}^{p}\sum_{k=1}^{p}\E\{\frac{1}{n}\sum_{i=1}^{n}\nabla^2 l(\bz_i;\btheta^*)-\bA\}_{j,k}^2]^{1/2}\leq\sqrt{\frac{L_4}{n}},
			\end{split}
		\end{equation*}
		where $\{\cdot\}_{j,k}$ represents the $(j,k)$-component of the matrix; we use Jensen's inequality in the first inequality, and we use Assumption 5 in the last equality. Thus, we have
		\begin{equation*}
			\|\widetilde{\bA}-\bA\|_2=O_p(n^{-1/2}+\sigma_1T^{-1/2}+\sigma_2).
		\end{equation*}
		
		We second consider the estimation of $\bS$. By the decomposition,
		\begin{equation*}
			\begin{split}
				\widetilde{\bS}-\bS=&\frac{1}{n}\sum_{i=1}^{n}\nabla l(\bz_i;\bar{\btheta}_T)\nabla l(\bz_i;\bar{\btheta}_T)^{\top}-\bS+\bxi^{\prime\prime}\\
				=&\frac{1}{n}\sum_{i=1}^{n}\nabla l(\bz_i;\bar{\btheta}_T)\nabla l(\bz_i;\bar{\btheta}_T)^{\top}-\frac{1}{n}\sum_{i=1}^{n}\nabla l(\bz_i;\btheta^*)\nabla l(\bz_i;\btheta^*)^{\top}\\
				&+\frac{1}{n}\sum_{i=1}^{n}\nabla l(\bz_i;\btheta^*)\nabla l(\bz_i;\btheta^*)^{\top}-\bS+\bxi^{\prime\prime}.
			\end{split}
		\end{equation*}
		We consider the three terms one by one. Put $g_i(\btheta)=\nabla l(\bz_i;\btheta)$, $\boldsymbol d=\bar\btheta_T-\btheta^*$, and
		\begin{equation*}
			D_n=\frac1n\sum_{i=1}^n\|g_i(\bar\btheta_T)-g_i(\btheta^*)\|_2^2,
			\qquad G_n=\frac1n\sum_{i=1}^n\|g_i(\btheta^*)\|_2^2=O_p(1).
		\end{equation*}
		By Theorem 2, $\boldsymbol d=O_p(n^{-1/2}+\sigma_1T^{-1/2})$, and the segment joining $\bar\btheta_T$ and $\btheta^*$ lies in $\Theta$ with probability tending to one. On this event, the integral Taylor formula and Assumption 5 give
		\begin{equation*}
			D_n\le 2\|\boldsymbol d\|_2^2\left\{\frac1n\sum_{i=1}^n\|\nabla^2l(\bz_i;\btheta^*)\|_2^2+L_2^2\|\boldsymbol d\|_2^2\right\}
			=O_p(\|\boldsymbol d\|_2^2).
		\end{equation*}
		Consequently, Cauchy--Schwarz gives the required evaluation-error bound,
		\begin{equation*}
			\begin{split}
				&\left\|\frac1n\sum_{i=1}^n\{g_i(\bar\btheta_T)g_i(\bar\btheta_T)^\top-g_i(\btheta^*)g_i(\btheta^*)^\top\}\right\|_2\\
				&\qquad\le 2\sqrt{D_nG_n}+D_n
				=O_p(n^{-1/2}+\sigma_1T^{-1/2}).
			\end{split}
		\end{equation*}
		By the fact $\E\{\nabla l(\bz_i;\btheta^*)\nabla l(\bz_i;\btheta^*)^{\top}\}=\bS$, we have
		\begin{equation*}
			\begin{split}
				&\E\|\frac{1}{n}\sum_{i=1}^{n}\nabla l(\bz_i;\btheta^*)\nabla l(\bz_i;\btheta^*)^{\top}-\bS\|_2\\
				\leq&[\sum_{j=1}^{p}\sum_{k=1}^{p}\E\{\frac{1}{n}\sum_{i=1}^{n}\nabla l(\bz_i;\btheta^*)\nabla l(\bz_i;\btheta^*)^{\top}-\bS\}_{j,k}^2]^{1/2}\leq\frac{C}{\sqrt n},
			\end{split}
		\end{equation*}
		where $C<\infty$ follows from the fourth score moment in Assumption 2 and $\{\cdot\}_{j,k}$ denotes a matrix entry. Thus, we have
		\begin{equation*}
			\|\widetilde{\bS}-\bS\|_2=O_p(n^{-1/2}+\sigma_1T^{-1/2}+\sigma_3).
		\end{equation*}
		
		For the unthresholded symmetrized estimate, $\widetilde{\bA}^{-1}=O_p(1)$ also holds without assuming $\sigma_2\to0$. Indeed, from any subsequence we can extract one along which $\sigma_2\to s\in[0,\infty]$. If $s=0$, then $\widetilde{\bA}\stackrel p\to\bA$; if $0<s<\infty$, then $\widetilde{\bA}\stackrel d\to\bA+s\bG$; and if $s=\infty$, then $\widetilde{\bA}/\sigma_2\stackrel d\to\bG$, where $\bG$ is the symmetrization of a matrix with independent standard Gaussian entries. Both $\bG$ and $\bA+s\bG$ are nonsingular almost surely, so the inverses are tight in every case; when $s=\infty$, $\widetilde{\bA}^{-1}=\sigma_2^{-1}(\widetilde{\bA}/\sigma_2)^{-1}=O_p(\sigma_2^{-1})$. Put $\bB=\bA^{-1}$ and $\widetilde{\bB}=\widetilde{\bA}^{-1}$. The inverse identity gives $\widetilde{\bB}-\bB=-\widetilde{\bB}(\widetilde{\bA}-\bA)\bB$. Combining the preceding bounds with
		\begin{equation*}
			\begin{split}
				\widetilde{\bB}\widetilde{\bS}\widetilde{\bB}-\bB\bS\bB
				&=\widetilde{\bB}(\widetilde{\bS}-\bS)\widetilde{\bB}
				+(\widetilde{\bB}-\bB)\bS\widetilde{\bB}\\
				&\quad+\bB\bS(\widetilde{\bB}-\bB),
			\end{split}
		\end{equation*}
		we obtain the following bound, which also follows from Lemma C.1 in \cite{chen2020statistical} when $\sigma_2=o(1)$:
		\begin{equation*}
			\|\widetilde{\bA}^{-1}\widetilde{\bS}\widetilde{\bA}^{-1}-\bA^{-1}\bS\bA^{-1}\|_2\lesssim O_p(n^{-1/2}+\sigma_1T^{-1/2}+\sigma_2+\sigma_3).
		\end{equation*}
		\hfill \( \blacksquare \)
	\end{proof}
	
	\begin{proof}[Proof of Theorem 3]
		By Corollary 1 (Corollary 2, or Corollary 3), we have
		\begin{equation*}
			\sqrt{n}(\bar{\theta}_{j,T}-\theta^*_j)\stackrel{d}{\to}N(0,\{\bA^{-1}\bS\bA^{-1}\}_{j,j}).
		\end{equation*}
		By Theorem 2, $\sigma_1=O(1)$ and $T\to\infty$, so $\sigma_1T^{-1/2}\to0$. The additional assumptions in Theorem 3 give $\sigma_2,\sigma_3\to0$. The first two bounds in Proposition 2 therefore imply $\widetilde{\bA}\stackrel p\to\bA$ and $\widetilde{\bS}\stackrel p\to\bS$. Since $\bA$ is positive definite, $\widetilde{\bA}$ is invertible with probability tending to one and $\widetilde{\bA}^{-1}\stackrel p\to\bA^{-1}$. Thus, we have
		\begin{equation*}
			\|\widetilde{\bA}^{-1}\widetilde{\bS}\widetilde{\bA}^{-1}-\bA^{-1}\bS\bA^{-1}\|_2=o_p(1).
		\end{equation*}
		We have
		\begin{equation*}
			\widetilde{\bA}^{-1}\widetilde{\bS}\widetilde{\bA}^{-1}\stackrel{p}{\to}\bA^{-1}\bS\bA^{-1}.
		\end{equation*}
		Consequently, $\widetilde V_{j,j}\stackrel p\to V_{j,j}>0$ for each coordinate. In particular, the estimated coordinate variance is positive with probability tending to one, and Slutsky's theorem gives
		\begin{equation*}
			\frac{\sqrt n(\bar\theta_{T,j}-\theta_j^*)}{\sqrt{\widetilde V_{j,j}}}\stackrel d\longrightarrow N(0,1).
		\end{equation*}
		Hence, for any $j=1,\dots,p$ and $\alpha\in(0,1)$, we have
		\begin{equation*}
			\lim_{n\to\infty}\mathbb{P}\{\theta_j^*\in\text{CI}_{\alpha}(\theta_j)\}=1-\alpha.
		\end{equation*}
		
		\hfill \( \blacksquare \)
	\end{proof}
	
	\subsection{Proof of Proposition 3 and Theorem 4}
	\begin{lemma}[Uniform linearization]\label{lem:uniform_linearization}
		Suppose the conditions of Theorem 2 hold. Let $\btheta_n^*$ be the empirical minimizer constructed in the proof of Theorem 1, and let $\bar{\bxi}^{(t)\prime}(0)$ denote the centered sampling innovation evaluated at $\btheta_n^*$. Use the signed Gaussian innovations $\boldsymbol n_t$ from the proof of Theorem 2. Define, with empty sums equal to zero,
		\begin{equation*}
			R_{n,s}=\sum_{t=1}^{s}(\btheta^{(t)}-\btheta_n^*)
			-\bA^{-1}\sum_{t=1}^{s}\{\bar{\bxi}^{(t)\prime}(0)+\boldsymbol n_t\},
			\qquad 0\leq s\leq T.
		\end{equation*}
		Then
		\begin{equation}\label{eq:uniform_linearization_rate}
			\frac1{\sqrt T}\max_{0\leq s\leq T}\|R_{n,s}\|
			=O_p\!\left(T^{(\alpha-1)/4}+T^{1/2-\alpha}
			+n^{-1/2}T^{(1-\alpha)/2}\right).
		\end{equation}
		Consequently,
		\begin{equation*}
			\frac{\sqrt n}{T}\max_{s\leq T}\|R_{n,s}\|=o_p(1)
			\quad\text{for every }T/n\geq c_0.
		\end{equation*}
		Moreover,
		\begin{equation*}
			\frac1{\sqrt T}\max_{s\leq T}\|R_{n,s}\|=o_p(1)
			\quad\text{whenever }T^{1-\alpha}/n\to0,
		\end{equation*}
		in particular when $T=O(n^2)$.
	\end{lemma}
	\begin{proof}
		Work on the high-probability data events and use the localization from the proof of Theorem 2. Write $e_t=\btheta^{(t)}-\btheta_n^*$, $d_n=\|\btheta_n^*-\btheta^*\|=O_p(n^{-1/2})$, $H_n=\nabla^2\mathcal L_n(\btheta_n^*)$, and $h_n(e)=\nabla\mathcal L_n(\btheta_n^*+e)$. Let $z_t(e)$ be the centered minibatch innovation, so that
		\begin{equation*}
			e_t=e_{t-1}-\eta_t h_n(e_{t-1})+\eta_t\{z_t(e_{t-1})+\boldsymbol n_t\},
			\qquad z_t(0)=\bar{\bxi}^{(t)\prime}(0).
		\end{equation*}
		Assumption 3 gives $\|z_t(e)\|\leq\Delta_g$ for every $e$, since any minibatch mean and the full-sample mean differ by at most the maximal pairwise gradient difference. Thus the conditional fourth moments of $z_t(e_{t-1})+\boldsymbol n_t$ are uniformly bounded. On any fixed localization ball $\|e\|\leq R$, the preceding curvature and growth arguments give
		\begin{equation*}
			\langle e,h_n(e)\rangle\geq c_R\|e\|^2,
			\qquad \|h_n(e)\|\leq K_R\|e\|.
		\end{equation*}
		Set $\tau_R=\inf\{t:\|e_t\|>R\}$ and $u_{q,t}=E[\|e_t\|^q1\{\tau_R>t\}\mid\mathcal F_n]$. The second-moment bound above gives $u_{2,t}\leq K_R\eta_t$. Expanding the fourth power of the recursion, using the centered innovations and their bounded fourth moments, gives, for all sufficiently large $t$,
		\begin{equation*}
			u_{4,t}\leq(1-c_R\eta_t)u_{4,t-1}
			+K_R\eta_t^2u_{2,t-1}+K_R\eta_t^4,
			\qquad u_{4,t}\leq K_R\eta_t^2.
		\end{equation*}
		The second inequality follows by iteration, since $t^{-1}=o(\eta_t)$; finitely many initial terms are absorbed into $K_R$.
		
		To control all endpoints, partition the time indices into consecutive blocks beginning at $b$ and having length $\lceil b^\alpha\rceil$, truncating the last block at $T$. On a block, $\eta_t\asymp b^{-\alpha}$ and $\sum\eta_t\leq K$. Apply the fourth-moment martingale maximal inequality to the increments
		$\eta_t\{z_t(e_{t-1})+\boldsymbol n_t\}1\{\tau_R>t-1\}$.
		Their maximal partial sum has conditional fourth moment at most
		\begin{equation*}
			K_R\left\{\left(\sum_{\text{block}}\eta_t^2\right)^2
			+\sum_{\text{block}}\eta_t^4\right\}
			\leq K_Rb^{-2\alpha}.
		\end{equation*}
		The drift bound, discrete Gronwall inequality, and $u_{4,b}\leq K_Rb^{-2\alpha}$ therefore imply
		\begin{equation*}
			E\left[\max_{b\leq t\leq b+\lceil b^\alpha\rceil}
			\{\|e_t\|^4 1\{\tau_R>t\}\}\,\middle|\,\mathcal F_n\right]
			\leq K_Rb^{-2\alpha}.
		\end{equation*}
		Here the martingale is stopped predictably, including a possible exit increment; the moments at the block start are killed at $\tau_R$. Summing the block bounds and using $\sum_{\text{blocks}}b^{2\alpha}\leq KT^{1+\alpha}$ gives
		\begin{equation*}
			E\left[1\{\tau_R>T\}\max_{1\leq s\leq T}
			\|e_s/\eta_s\|^4\,\middle|\,\mathcal F_n\right]
			\leq K_RT^{1+\alpha}.
		\end{equation*}
		Thus the maximal endpoint term divided by $\sqrt T$ is $O_p(T^{(\alpha-1)/4})$ on $\{\tau_R>T\}$.
		
		Put $r_n(e)=h_n(e)-H_ne$ and $\zeta_t=z_t(e_{t-1})-z_t(0)$. Summation by parts gives, for $1\leq s\leq T$, the exact identity
		\begin{equation*}
			\begin{split}
				H_n\sum_{t=1}^{s}e_t
				={}&\sum_{t=1}^{s}\{z_t(0)+\boldsymbol n_t\}
				+\sum_{t=1}^{s}\zeta_t-\sum_{t=1}^{s}r_n(e_{t-1})+D_s,\\
				D_s={}&\frac{e_0}{\eta_1}-\frac{e_s}{\eta_s}
				+\sum_{t=1}^{s-1}(\eta_{t+1}^{-1}-\eta_t^{-1})e_t
				+H_n(e_s-e_0).
			\end{split}
		\end{equation*}
		Besides the endpoint bound, the stopped second moments and
		$\eta_{t+1}^{-1}-\eta_t^{-1}=O(t^{\alpha-1})$ give
		\begin{equation*}
			\frac1{\sqrt T}\sum_{t=1}^{T-1}
			(\eta_{t+1}^{-1}-\eta_t^{-1})\|e_t\|
			=O_p(T^{(\alpha-1)/2})
			\quad\text{on }\{\tau_R>T\}.
		\end{equation*}
		The initial and index-shift terms contribute $O_p(T^{-1/2})$ there. The anchored Taylor bound from Theorem 1 yields
		\begin{equation*}
			\|r_n(e)\|\leq K_R(\|e\|^2+d_n\|e\|),\qquad
			\frac1{\sqrt T}\sum_{t=1}^{T}\|r_n(e_{t-1})\|
			=O_p\!\left(T^{1/2-\alpha}+n^{-1/2}T^{(1-\alpha)/2}\right)
		\end{equation*}
		on $\{\tau_R>T\}$. Finally, the same mean-square innovation bound used in Theorem 1 and Doob's inequality gives
		\begin{equation*}
			E\left[\max_{s\leq T}\left\|\frac1{\sqrt T}
			\sum_{t=1}^{s}\zeta_t1\{\tau_R>t-1\}\right\|^2
			\,\middle|\,\mathcal F_n\right]
			\leq\frac{K_R}{T}\left(1+\sum_{t=1}^{T-1}\eta_t\right)
			=O(T^{-\alpha}).
		\end{equation*}
		The frozen-noise partial sums divided by $\sqrt T$ have a tight maximum by Doob's inequality. Assumption 5 and the empirical-root expansion give $\|H_n-\bA\|=O_p(n^{-1/2})$, so replacing $H_n^{-1}$ by $\bA^{-1}$ adds $O_p(n^{-1/2})$. Combining the bounds proves \eqref{eq:uniform_linearization_rate} on $\{\tau_R>T\}$. Letting $R\to\infty$ in the uniform exit bound \eqref{eq:uniform_localization} removes the stopping. Multiplying the rate by $\sqrt{n/T}$ proves the first consequence; the second follows from $\alpha\in(1/2,1)$ and $T^{1-\alpha}/n\to0$. If $T=O(n^2)$, then $T^{1-\alpha}/n=O(n^{1-2\alpha})\to0$.
		\hfill $\blacksquare$
	\end{proof}
	
	\begin{proof}[Proof of Proposition 3]
		We retain the decomposition into the empirical-estimation, sampling, and privacy components, using Lemma \ref{lem:uniform_linearization} to make the approximation uniform. Put $h_T(r)=\lfloor rT\rfloor$, and define
		\begin{equation*}
			\bar{\bDelta}^{(T)}(r)=\frac1T\sum_{t=1}^{h_T(r)}(\btheta^{(t)}-\btheta^*).
		\end{equation*}
		Then, uniformly for $r\in[0,1]$,
		\begin{equation*}
			\begin{split}
				\sqrt n\bar{\bDelta}^{(T)}(r)
				={}&\frac{h_T(r)}T\sqrt n(\btheta_n^*-\btheta^*)
				+\frac{\sqrt{n/T}}{\sqrt T}\sum_{t=1}^{h_T(r)}\bA^{-1}\bar{\bxi}^{(t)\prime}(0)\\
				&+\frac{\sqrt{n/T}}{\sqrt T}\sum_{t=1}^{h_T(r)}\bA^{-1}\boldsymbol n_t+o_p(1).
			\end{split}
		\end{equation*}
		The empirical-root expansion \eqref{eq:empirical_root_expansion} gives
		$\sqrt n(\btheta_n^*-\btheta^*)\stackrel d\to\bV^{1/2}\bphi$.
		Since $\sup_r|h_T(r)/T-r|\leq T^{-1}$, its contribution converges to $r\bV^{1/2}\bphi$ uniformly.
		
		Conditional on $\mathcal F_n$, the frozen sampling innovations are independent and identically distributed across iterations, have mean zero, and have exact SRSWOR covariance
		\begin{equation*}
			\operatorname{Var}\{\bar{\bxi}^{(t)\prime}(0)\mid\mathcal F_n\}
			=\frac{n-m}{m(n-1)}\bS_n^*,\qquad
			\bS_n^*=\frac1n\sum_{i=1}^{n}\nabla l(\bz_i;\btheta_n^*)
			\nabla l(\bz_i;\btheta_n^*)^{\top}\stackrel p\to\bS.
		\end{equation*}
		The covariance convergence was established in the proof of Theorem 1. Assumption 3 also gives $\|\bar{\bxi}^{(t)\prime}(0)\|\leq\Delta_g$, so the conditional Lindeberg condition holds. Applying the functional central limit theorem \citep{mcleish1978extended}, along the subsequences specified in the main text, gives
		\begin{equation*}
			\frac1{\sqrt T}\sum_{t=1}^{h_T(r)}\bA^{-1}\bar{\bxi}^{(t)\prime}(0)
			\ \Big|\ \mathcal F_n\ \stackrel d\longrightarrow
			(\bV/m)^{1/2}\bW_1(r)
			\quad\text{in }D([0,1],\mathbb R^p).
		\end{equation*}
		The independent Gaussian privacy innovations similarly give
		\begin{equation*}
			\frac1{\sqrt T}\sum_{t=1}^{h_T(r)}\bA^{-1}\boldsymbol n_t
			\stackrel d\longrightarrow\sigma_1\bA^{-1}\bW_2(r).
		\end{equation*}
		The conditional sampling laws converge in probability to a nonrandom Gaussian law; hence the two process limits are jointly independent of the empirical-root limit, as well as of each other. This establishes the joint convergence needed below. Combining the three components and the uniform remainder proves Proposition 3. This uses only $T/n\geq c_0$, since $\sqrt{n/T}$ times the bound in \eqref{eq:uniform_linearization_rate} tends to zero under that condition.
		\hfill $\blacksquare$
	\end{proof}
	
	\begin{proof}[Proof of Theorem 4]
		It remains to consider the random scaling matrix. Put $h_T(r)=\lfloor rT\rfloor$ and define the centered process
		\begin{equation*}
			B_T(r)=\sqrt{\frac mT}\left\{
			\sum_{t=1}^{h_T(r)}(\btheta^{(t)}-\btheta_n^*)
			-\frac{h_T(r)}T\sum_{t=1}^{T}(\btheta^{(t)}-\btheta_n^*)\right\}
			=\sqrt{\frac mT}\sum_{t=1}^{h_T(r)}(\btheta^{(t)}-\bar{\btheta}_T).
		\end{equation*}
		All three corollaries imply $T=O(n^2)$: Corollaries 1 and 2 use $1<l<2$, while Corollary 3 has $m\sqrt T/n\to c$ with bounded positive integer $m$. Consequently, the stronger conclusion of Lemma \ref{lem:uniform_linearization} applies. With
		\begin{equation*}
			M_T(r)=\frac1{\sqrt T}\sum_{t=1}^{h_T(r)}
			\bA^{-1}\{\bar{\bxi}^{(t)\prime}(0)+\boldsymbol n_t\},
		\end{equation*}
		it gives directly
		\begin{equation*}
			\sup_{r\in[0,1]}\left\|B_T(r)-\sqrt m
			\left\{M_T(r)-\frac{h_T(r)}T M_T(1)\right\}\right\|=o_p(1).
		\end{equation*}
		Thus the joint functional convergence proved above yields
		\begin{equation*}
			B_T(r)\stackrel d\longrightarrow
			\bV^{1/2}\{\bW_1(r)-r\bW_1(1)\}
			+\sqrt m\sigma_1\bA^{-1}\{\bW_2(r)-r\bW_2(1)\}.
		\end{equation*}
		This centered approximation is at the $\sqrt T$ scale and does not rescale the remainder in Proposition 3. The random scaling matrix satisfies the exact identity
		\begin{equation*}
			\widehat{\bV}=\frac1T\sum_{t=1}^{T}B_T(t/T)^{\otimes2}.
		\end{equation*}
		Since the limiting paths are continuous, the continuous mapping theorem and the Riemann sums give
		\begin{equation*}
			\widehat{\bV}\stackrel d\longrightarrow
			\int_0^1\left[\bV^{1/2}\{\bW_1(r)-r\bW_1(1)\}
			+\sqrt m\sigma_1\bA^{-1}\{\bW_2(r)-r\bW_2(1)\}\right]^{\otimes2}\,\mathrm dr.
		\end{equation*}
		The two independent Brownian bridges combine coordinatewise into a Brownian bridge with the summed variance, so
		\begin{equation*}
			\widehat{V}_{j,j}\stackrel{d}{\to}\{V_{j,j}+(A^{-2})_{j,j}\sigma^2_1 m\}\int_{0}^{1}\{W_1(r)-rW_1(1)\}^2\text{d}r,
		\end{equation*}
		where $V_{j,j}$ is the $(j,j)$-th component of the matrix $\bA^{-1}\bS\bA^{-1}$, $(A^{-2})_{j,j}$ is the $(j,j)$-th component of the matrix $\bA^{-2}$, and $W_1(r)$ is a standard Wiener process. Furthermore, by the conditions in Corollary 1 (Corollary 2 or Corollary 3), we have $k\to\infty$ and $1/k=o(1)$. Thus, combining the result from Theorem 2, we have
		\begin{equation*}
			\sqrt{n}(\bar{\theta}_{T,j}-\theta^*_j)V_{j,j}^{-1/2}\stackrel{d}{\to}Z.
		\end{equation*}
		Thus, we have
		\begin{equation*}
			\sqrt{n}(\bar{\theta}_{T,j}-\theta^*_j)\widehat{V}_{j,j}^{-1/2}\stackrel{d}{\to}Z\cdot\bigg[\int_{0}^{1}\{W(r)-rW(1)\}^2\text{d}r\bigg]^{-1/2}\cdot\bigg\{\frac{V_{j,j}}{V_{j,j}+m\sigma_1^2(A^{-2})_{j,j}}\bigg\}^{1/2},
		\end{equation*}
		where $V_{j,j}>0$ by the stated coordinate restriction, and the standardized empirical-root limit $Z$ is independent of the Brownian bridge by the joint convergence established in the proof of Proposition 3.
		
		The critical value relies on the key observation that $Z\times\bigg[\int_{0}^{1}\{W_1(r)-rW_1(1)\}^2\text{d}r\bigg]^{-1/2}$ and $W_1(1)\times\bigg[\int_{0}^{1}\{W_1(r)-rW_1(1)\}^2\text{d}r\bigg]^{-1/2}$ have the same cumulative distribution function. Notice that $W_1(1)$ has the same distribution as $Z$, and its covariance with $W_1(r)-rW_1(1)$ is
		\begin{equation*}
			\text{Cov}\{W_1(1),W_1(r)-rW_1(1)\}=r-r=0.
		\end{equation*}
		By the fact that the uncorrelatedness of jointly normal random variables implies independence, we have $W_1(1)$ is independent of $\int_{0}^{1}\{W_1(r)-rW_1(1)\}^2\text{d}r$. Thus, the critical value can be found in Table 1 in \cite{abadir1997two}.

		By the condition $\sigma_1=o(1)$, $m$ is bounded, we have
		\begin{equation*}
			\widehat{V}_{j,j}\stackrel{d}{\to}V_{j,j}\int_{0}^{1}\{W_1(r)-rW_1(1)\}^2\text{d}r,
		\end{equation*}
		and thus
		\begin{equation*}
			\sqrt{n}(\bar{\theta}^{(T)}_j-\theta^*_j)\widehat{V}_{j,j}^{-1/2}\stackrel{d}{\to}Z\times\bigg[\int_{0}^{1}\{W_1(r)-rW_1(1)\}^2\text{d}r\bigg]^{-1/2}.
		\end{equation*}
		Thus, the confidence interval is asymptotically valid.
		
		\hfill \( \blacksquare \)
	\end{proof}
	\subsection{Lemma}
	
	\begin{lemma}\label{lem:linear_algebra}
		Let $\bA$ be symmetric positive definite, $\eta_t=\eta t^{-\alpha}$ for $t\ge1$, $\eta>0$, and $0<\alpha<1$. Define
		\[
		P_{t,j}=\prod_{s=j+1}^{t}(\bI-\eta_s\bA),\quad P_{j,j}=\bI,\quad
		B_{j,T}=\eta_j\sum_{t=j}^{T}P_{t,j},\quad
		C_T=\frac1T\sum_{t=1}^{T}P_{t,0}.
		\]
		Then
		\[
		\|C_T\|=O(T^{-1}),\qquad
		\max_{1\le j\le T}\|B_{j,T}-\bA^{-1}\|\le K,\qquad
		\frac1T\sum_{j=1}^{T}\|B_{j,T}-\bA^{-1}\|=O(T^{\alpha-1}).
		\]
		The constants can be chosen uniformly when the eigenvalues of $\bA$ lie in a fixed compact subset of $(0,\infty)$.
	\end{lemma}
	\begin{proof}
		Choose a fixed $J$ so that $\eta_s\lambda_{\max}(\bA)\le1$ for $s\ge J$, uniformly over the stated set of matrices. The finitely many earlier factors are uniformly bounded. For $t\ge j\ge J$,
		\[
		\|P_{t,j}\|\le\exp\left(-\lambda_{\min}(\bA)\sum_{s=j+1}^{t}\eta_s\right).
		\]
		In particular, $\sum_{t=1}^{\infty}\|P_{t,0}\|<\infty$ uniformly, which proves the bound for $C_T$.
		
		Since $\bA\eta_{t+1}P_{t,j}=P_{t,j}-P_{t+1,j}$, telescoping gives the exact identity
		\[
		B_{j,T}-\bA^{-1}
		=\sum_{t=j}^{T}(\eta_j-\eta_{t+1})P_{t,j}-\bA^{-1}P_{T+1,j}.
		\]
		For $j\ge J$, split the first sum at $2j$. On $j\le t\le2j$,
		$\eta_j-\eta_{t+1}\le K(t+1-j)j^{-\alpha-1}$ and
		$\|P_{t,j}\|\le\exp\{-c(t-j)j^{-\alpha}\}$, so the sum of the norms on this range is at most $Kj^{\alpha-1}$. On $t>2j$, the bound
		\[
		\|P_{t,j}\|\le K\exp\{-c[(t+1)^{1-\alpha}-(j+1)^{1-\alpha}]\}
		\]
		makes the remaining contribution exponentially small in $j^{1-\alpha}$, and hence bounded by $Kj^{\alpha-1}$. The finitely many $j<J$ have uniformly bounded contributions. Finally,
		\[
		\sum_{j=1}^{T}\|P_{T+1,j}\|
		\le K+K\sum_{j=J}^{T}\exp\{-c(T+1-j)(T+1)^{-\alpha}\}
		=O(T^{\alpha}).
		\]
		Together with $\sum_{j=1}^{T}j^{\alpha-1}=O(T^{\alpha})$, the telescoping identity proves the average-norm bound and also the uniform bound on individual weights.
		\hfill $\blacksquare$
	\end{proof}
	
	\subsection{Proof of Theorem 5}
	\begin{proof}
		The procedure is similar to the proof of Theorem 2. We work on the high-probability data events constructed there. In particular, the empirical minimizer $\boldsymbol\theta_n^*$ exists, and deterministic constants $\rho,\lambda_0,L_0>0$ satisfy
		\[
		\lambda_0\boldsymbol I\preceq\nabla^2\mathcal L_n(\boldsymbol\theta)\preceq L_0\boldsymbol I,
		\qquad \|\boldsymbol\theta-\boldsymbol\theta_n^*\|\le\rho.
		\]
		Put $\Delta=\Delta_g$, the sensitivity bound on the original gradients in Assumption 3. A sufficient fixed threshold is
		\begin{equation}\label{eq:clipping_threshold_sufficient}
			\tau>\max\left\{\Delta+L_0\rho,\;2\Delta+\frac{4\Delta^2}{\lambda_0\rho}\right\}.
		\end{equation}
		Assumption 3 gives $\|\nabla l(\bz_i;\boldsymbol\theta)-\nabla\mathcal L_n(\boldsymbol\theta)\|\le\Delta<\tau/2$ for every $i$ and $\boldsymbol\theta\in\mathbb R^p$.
		
		For notational simplicity, write $\boldsymbol G_t=\nabla\mathcal L_n(\boldsymbol\theta^{(t-1)})$ and $\boldsymbol g_{i,t}=\nabla l(\boldsymbol z_i;\boldsymbol\theta^{(t-1)})$. We define the clipped innovation by
		$$
		\boldsymbol{\xi}^{(t)\prime}_{(i)}=\Pi_\tau(\boldsymbol g_{i,t})-\boldsymbol G_t,
		\qquad
		\bar{\boldsymbol{\xi}}^{(t)\prime}=\frac{1}{m}\sum_{i=1}^n I_i^{(t)}\boldsymbol{\xi}^{(t)\prime}_{(i)}.
		$$
		Then the clipped minibatch gradient can be written as
		$$
		\frac{1}{m}\sum_{i=1}^n I_i^{(t)}\Pi_\tau(\boldsymbol g_{i,t})=\boldsymbol G_t+\bar{\boldsymbol{\xi}}^{(t)\prime}.
		$$
		Therefore, the clipped DP-SGD update is
		$$
		\boldsymbol{\theta}^{(t)}=\boldsymbol{\theta}^{(t-1)}-\eta_t\boldsymbol G_t-\eta_t\bar{\boldsymbol{\xi}}^{(t)\prime}-\eta_t\boldsymbol n_t,
		$$
		where $\boldsymbol n_t$ is the Gaussian noise added to the gradient in Algorithm 1. We introduce the notation $\boldsymbol\Delta_{tr}^{(t)}=\boldsymbol\theta^{(t)}-\boldsymbol\theta_n^*$ and
		\[
		\boldsymbol F_n(\boldsymbol\theta)=\frac1n\sum_{i=1}^n\Pi_\tau\{\nabla l(\bz_i;\boldsymbol\theta)\}.
		\]
		Here $\Pi_\tau$ is the projection onto the Euclidean ball of radius $\tau$, and $K$ denotes a generic positive constant. Since the expectation is over the sampling indicators, we have
		$$
		\mathbb E[\bar{\boldsymbol{\xi}}^{(t)\prime}\mid\mathscr F_{t-1},\mathcal F_n]=\frac{1}{n}\sum_{i=1}^n\{\Pi_\tau(\boldsymbol g_{i,t})-\boldsymbol G_t\}.
		$$
		In particular, the clipped innovation need not be centered. The clipped minibatch norm is at most $\tau$, so, for $X_t=\|\boldsymbol\Delta_{tr}^{(t)}\|^2$, the conditional distance recursion is
		\[
		E(X_t\mid\mathscr F_{t-1},\mathcal F_n)
		\le X_{t-1}-2\eta_t\langle\boldsymbol\Delta_{tr}^{(t-1)},\boldsymbol F_n(\boldsymbol\theta^{(t-1)})\rangle+K\eta_t^2,
		\]
		where $K$ is uniform in $n$ because $\sigma_1=O(1)$. It remains to lower-bound the drift term.
		
		Let $a_t=\|\boldsymbol G_t\|_2$ and $\boldsymbol h_{i,t}=\boldsymbol g_{i,t}-\boldsymbol G_t$. We have $\|\boldsymbol h_{i,t}\|_2\le\Delta$ for all $i$. If $\|\boldsymbol\Delta_{tr}^{(t-1)}\|\le\rho$, then $\|\boldsymbol g_{i,t}\|\le\Delta+L_0\rho<\tau$, so clipping is inactive. If any gradient is clipped, then $a_t>\tau-\Delta>\Delta$. Define
		$$
		\rho_{i,t}=\min\left(1,\frac{\tau}{\|\boldsymbol g_{i,t}\|_2}\right),
		$$
		taking $\rho_{i,t}=1$ at a zero gradient. Then $\Pi_\tau(\boldsymbol g_{i,t})=\rho_{i,t}\boldsymbol g_{i,t}$. When at least one gradient is clipped,
		\[
		r_{-,t}:=\frac{\tau}{a_t+\Delta}\le\rho_{i,t}\le r_{+,t}:=\min\{1,\tau/(a_t-\Delta)\},
		\qquad \frac{r_{+,t}-r_{-,t}}{r_{-,t}}\le\frac{2\Delta}{a_t-\Delta}.
		\]
		Using $\sum_i\boldsymbol h_{i,t}=0$ and convexity, we obtain
		\[
		\langle\boldsymbol\Delta_{tr}^{(t-1)},\boldsymbol F_n(\boldsymbol\theta^{(t-1)})\rangle
		\ge r_{-,t}\left\{\langle\boldsymbol\Delta_{tr}^{(t-1)},\boldsymbol G_t\rangle
		-\frac{2\Delta^2\|\boldsymbol\Delta_{tr}^{(t-1)}\|}{a_t-\Delta}\right\}.
		\]
		Outside the radius-$\rho$ ball, local curvature and convexity along rays from $\boldsymbol\theta_n^*$ give
		\[
		\langle\boldsymbol\Delta_{tr}^{(t-1)},\boldsymbol G_t\rangle
		\ge\lambda_0\rho\|\boldsymbol\Delta_{tr}^{(t-1)}\|.
		\]
		Moreover, in the clipping case, \eqref{eq:clipping_threshold_sufficient} gives $a_t-\Delta>\tau-2\Delta>4\Delta^2/(\lambda_0\rho)$. Combining this with the no-clipping case, we have globally
		\begin{equation}\label{eq:clipped_radial_drift}
			\langle\boldsymbol\Delta_{tr}^{(t-1)},\boldsymbol F_n(\boldsymbol\theta^{(t-1)})\rangle
			\ge\frac12\min\left\{1,\frac{\tau}{a_t+\Delta}\right\}
			\langle\boldsymbol\Delta_{tr}^{(t-1)},\boldsymbol G_t\rangle\ge0.
		\end{equation}
		On every fixed ball $\|\boldsymbol\Delta_{tr}^{(t-1)}\|\le R$, the growth bound in Assumption 2(i) bounds $a_t$ on the stated data events. Thus the multiplier is bounded away from zero, and the same ray argument yields a lower bound $c_R\|\boldsymbol\Delta_{tr}^{(t-1)}\|^2$ for a deterministic $c_R>0$.
		
		The conditional distance recursion now shows that $X_t+K\sum_{s>t}\eta_s^2$ is a nonnegative conditional supermartingale. Its maximal inequality yields
		\[
		P\left(\sup_{t\ge0}\|\boldsymbol\Delta_{tr}^{(t)}\|>R\mid\mathcal F_n\right)
		\le\frac{X_0+K\sum_{s\ge1}\eta_s^2}{R^2}.
		\]
		Define $\tau_R=\inf\{t\ge0:\|\boldsymbol\Delta_{tr}^{(t)}\|>R\}$. As in the proof of Theorem 2, the killed second moments $u_t=E[X_t1\{\tau_R>t\}\mid\mathcal F_n]$ satisfy, for all sufficiently large $t$,
		\[
		u_t\le(1-2c_R\eta_t)u_{t-1}+K\eta_t^2,
		\qquad u_t\le K_R\eta_t.
		\]
		The finitely many initial steps are bounded separately.
		
		It remains to consider the asymptotic normality of the corresponding linear processes. Define the centered, unclipped sampling innovations along the actual clipped path by
		\[
		\boldsymbol\xi_{(i)}^{(t)}=\boldsymbol g_{i,t}-\boldsymbol G_t,
		\qquad \bar{\boldsymbol\xi}^{(t)}=\frac1m\sum_{i=1}^n I_i^{(t)}\boldsymbol\xi_{(i)}^{(t)}.
		\]
		With $\widehat{\boldsymbol A}=\nabla^2\mathcal L_n(\boldsymbol\theta_n^*)$, define the two linear comparison processes
		\[
		\begin{split}
			\boldsymbol\Delta^{(t)}
			&=(\boldsymbol I-\eta_t\widehat{\boldsymbol A})\boldsymbol\Delta^{(t-1)}
			-\eta_t\{\bar{\boldsymbol\xi}^{(t)}+\boldsymbol n_t\},\\
			\boldsymbol\Delta_1^{(t)}
			&=(\boldsymbol I-\eta_t\widehat{\boldsymbol A})\boldsymbol\Delta_1^{(t-1)}
			-\eta_t\{\bar{\boldsymbol\xi}^{(t)\prime}+\boldsymbol n_t\},
		\end{split}
		\]
		with $\boldsymbol\Delta_1^{(0)}=\boldsymbol\Delta^{(0)}=\boldsymbol\Delta_{tr}^{(0)}$. Both innovations are evaluated along the actual clipped trajectory, rather than at the linear comparison iterates. We use $\bar{\boldsymbol\Delta}^{(T)}=T^{-1}\sum_{t=1}^T\boldsymbol\Delta^{(t)}$ and $\bar{\boldsymbol\Delta}_1^{(T)}=T^{-1}\sum_{t=1}^T\boldsymbol\Delta_1^{(t)}$.
		
		We couple the two comparison processes by using the same sampling indicators $\{I_i^{(t)}\}$ and the same Gaussian noises $\boldsymbol n_t$. By the definition of $\Pi_\tau$, we have $\boldsymbol\xi_{(i)}^{(t)\prime}=\boldsymbol\xi_{(i)}^{(t)}+\boldsymbol d_{i,t}$, where
		\[
		\boldsymbol d_{i,t}=\Pi_\tau(\boldsymbol g_{i,t})-\boldsymbol g_{i,t}.
		\]
		Consequently,
		$$
		\bar{\boldsymbol\xi}^{(t)\prime}-\bar{\boldsymbol\xi}^{(t)}=\frac1m\sum_{i=1}^n I_i^{(t)}\boldsymbol d_{i,t}.
		$$
		Let $\boldsymbol\delta^{(T)}=\bar{\boldsymbol\Delta}_1^{(T)}-\bar{\boldsymbol\Delta}^{(T)}$. Since the two processes have the same initial value and use the same Gaussian noises, the initial-condition term and the privacy-noise term cancel in the difference. By the same linear representation used in the proof of Theorem~2, we obtain
		\[
		\sqrt n\,\boldsymbol\delta^{(T)}
		=-\frac{\sqrt n}{T}\sum_{t=1}^T B_{t,T}(\widehat{\boldsymbol A})
		\{\bar{\boldsymbol\xi}^{(t)\prime}-\bar{\boldsymbol\xi}^{(t)}\}.
		\]
		Since the exact averaging weights satisfy $\|B_{t,T}(\widehat{\boldsymbol A})\|\le K$, we have
		\[
		\|\sqrt n\,\boldsymbol\delta^{(T)}\|
		\le K\frac{\sqrt n}{T}\sum_{t=1}^T\frac1m\sum_{i=1}^n I_i^{(t)}\|\boldsymbol d_{i,t}\|.
		\]
		If $\|\boldsymbol\Delta_{tr}^{(t-1)}\|\le\rho$, clipping is inactive and $\boldsymbol d_{i,t}=\boldsymbol0$ for all $i$. On a fixed localization ball, the growth bound and $\|\boldsymbol g_{i,t}-\boldsymbol G_t\|\le\Delta$ bound every original gradient. Hence, on $\{\tau_R>T\}$,
		\[
		\|\boldsymbol d_{i,t}\|\le C_R1\{\|\boldsymbol\Delta_{tr}^{(t-1)}\|>\rho\}.
		\]
		Thus,
		\[
		\|\sqrt n\,\boldsymbol\delta^{(T)}\|
		\le K_R\frac{\sqrt n}{T}\sum_{t=1}^T1\{\|\boldsymbol\Delta_{tr}^{(t-1)}\|>\rho\}.
		\]
		It remains to show that the right-hand side is $o_p(1)$. By Markov's inequality and the stopped moment bound,
		\begin{align*}
			&\frac{\sqrt n}{T}\sum_{t=1}^T
			P\{\|\boldsymbol\Delta_{tr}^{(t-1)}\|>\rho,\ \tau_R>t-1\mid\mathcal F_n\}\\
			&\quad\le\frac{\sqrt n}{\rho^2T}\sum_{t=1}^T
			E[\|\boldsymbol\Delta_{tr}^{(t-1)}\|^2 1\{\tau_R>t-1\}\mid\mathcal F_n]\\
			&\quad\le\frac{K_R\sqrt n}{T}\left(1+\sum_{t=1}^{T-1}\eta_t\right)=o(1).
		\end{align*}
		Since $\eta_t=\eta t^{-\alpha}$ with $\alpha>1/2$ and $T=kn$ with $k$ bounded away from zero,
		$$
		\frac{\sqrt n}{T}\sum_{t=1}^{T-1}\eta_t=O(\sqrt n\,T^{-\alpha})=o(1),
		$$
		and the finitely many initial terms contribute $O(\sqrt n/T)=o(1)$. Conditional Markov's inequality therefore gives $\sqrt n\,\boldsymbol\delta^{(T)}=o_p(1)$ on $\{\tau_R>T\}$. Letting $R\to\infty$ in the exit bound removes the stopping. Consequently, $\sqrt n(\bar{\boldsymbol\Delta}_1^{(T)}-\bar{\boldsymbol\Delta}^{(T)})=o_p(1)$.
		
		It remains to connect the actual clipped DP-SGD iterates with the linear comparison process. Their updates use the same clipped innovations, so the difference is driven only by
		\[
		\nabla\mathcal L_n(\boldsymbol\theta_n^*+\boldsymbol\Delta_{tr}^{(t-1)})
		-\widehat{\boldsymbol A}\boldsymbol\Delta_{tr}^{(t-1)}.
		\]
		By the anchored Taylor bound and the same nonlinear-to-linear approximation used in the proof of Theorem~2, now using the stopped moment bound established above, we have
		\[
		\sqrt n\{\bar{\boldsymbol\theta}_T-\boldsymbol\theta_n^*-\bar{\boldsymbol\Delta}_1^{(T)}\}=o_p(1).
		\]
		The centered sampling innovations in the first linear process can be frozen at $\boldsymbol\theta_n^*$ by the same predictably stopped argument as in Theorem 2. Clipping is inactive there, so the frozen sampling scores and the privacy-noise term are unchanged. Combining their conditional CLT with the empirical-root expansion from Theorem 2 gives the same three independent limiting components. Therefore, gradient clipping does not change the first-order asymptotic distribution. Therefore, by Slutsky's theorem,
		$$
		\sqrt n(\bar{\boldsymbol\theta}_T-\boldsymbol\theta^*)\stackrel{d}{\to}\boldsymbol\phi_{\mathrm{stat}}+\boldsymbol\phi_{\mathrm{sam}}+\boldsymbol\phi_{\mathrm{privacy}},
		$$
		where the three terms are defined as in Theorem~2.
		\hfill \( \blacksquare \)
	\end{proof}

	\section{Theoretical Details of Examples in Section 5}
	
	\subsection{Ordinary Linear Regression}
	
	By taking the expectation, we have that the population loss function is
	\begin{equation*}
		\begin{split}
			\mathcal{L}(\btheta):=&\mathbb{E}(y_i-\bx_i^{\top}\btheta)^2/2=\mathbb{E}\{e_i+\bx_i^{\top}(\btheta^*-\btheta)\}^2/2\\
			=&\mathbb{E}\{e_i^2+2e_i\bx_i^{\top}(\btheta^*-\btheta)+(\btheta^*-\btheta)^{\top}\bx_i\bx_i^{\top}(\btheta^*-\btheta)\}/2\\
			=&\sigma^2/2+(\btheta^*-\btheta)^{\top}\bSigma(\btheta^*-\btheta)/2.
		\end{split}
	\end{equation*}
	We first check strong convexity:
	\begin{equation*}
		\begin{split}
			&\mathcal{L}(\btheta^{\prime})-\mathcal{L}(\btheta)-\langle\nabla \mathcal{L}(\btheta),\btheta^{\prime}-\btheta\rangle\\
			=&(\btheta^*-\btheta^{\prime})^{\top}\bSigma(\btheta^*-\btheta^{\prime})/2-(\btheta^*-\btheta)^{\top}\bSigma(\btheta^*-\btheta)/2-\langle\bSigma(\btheta-\btheta^*),\btheta^{\prime}-\btheta\rangle\\
			=&(\btheta-\btheta^{\prime})^{\top}\bSigma(\btheta-\btheta^{\prime})/2\geq\lambda_{\min}(\bSigma)/2\|\btheta-\btheta^{\prime}\|_2^2,
		\end{split}
	\end{equation*}
	thus the strong convexity holds for $\lambda=\lambda_{\min}(\bSigma)>0$. The Hessian matrix exists and equals $\mathbb{E}(\bSigma)=\bSigma$.
	
	Then, we check Assumption 2. First, the sample loss function $l(\bx_i;\btheta)=(y_i-\bx_i^{\top}\btheta)^2/2$ is continuously differentiable and satisfies
	\begin{equation*}
		\begin{split}
			\|\nabla l(\bx_i;\btheta)\|_2=|y_i-\bx_i^{\top}\btheta|\cdot\|\bx_i\|_2=|e_i+\bx_i^{\top}(\btheta^*-\btheta)|\cdot\|\bx_i\|_2,
		\end{split}
	\end{equation*}
	which is integrable. The innovation $\bxi(\btheta)$ satisfies
	\begin{equation*}
		\begin{split}
			\bxi(\btheta)&=\nabla \mathcal{L}(\btheta)-\nabla l(\bx_i;\btheta)=(\bSigma-\bx_i\bx_i^{\top})(\btheta-\btheta^*)+\bx_i e_i.
		\end{split}
	\end{equation*}
	Thus, the expectation satisfies $\bE\bxi(\btheta)=\boldsymbol{0}$ and
	\begin{equation*}
		\begin{split}
			\mathbb{E}\bxi(\btheta)\bxi(\btheta)^{\top}
			&=\mathbb{E}(\bx_i\bx_i^{\top}e_i^2)\\
			&\quad+\mathbb{E}\big[(\bSigma-\bx_i\bx_i^{\top})(\btheta-\btheta^*)(\btheta-\btheta^*)^{\top}(\bSigma-\bx_i\bx_i^{\top})\big].
		\end{split}
	\end{equation*}
	The conditions in Assumption 2 are satisfied. The strengthened global condition in Assumption 2(i) holds with $B(\bz)=\|\bx\|^2+|y|\|\bx\|$, since
	$\|\nabla l(\bz;\btheta)\|\le |y|\|\bx\|+\|\bx\|^2\|\btheta\|$ for every $\btheta\in\mathbb R^p$. The bounded-support design conditions ensure $E B(\bZ)^2<\infty$. Moreover, $\nabla^2l(\bz;\btheta)=\bx\bx^\top$ does not depend on $\btheta$, so Assumption 5 holds with $L_2=0$ and a finite $L_4$.
	
	We then consider sensitivity at a fixed parameter satisfying $\|\btheta\|_2\le C_0$. The gradient sensitivity satisfies
	\begin{equation*}
		\begin{split}
			\Delta_g=\max_{\bz,\bz^{\prime}}\|\nabla l(\bz;\btheta)-\nabla l(\bz^{\prime};\btheta)\|_2&= \max_{\bz,\bz^{\prime}}\|(y\bx-y^{\prime}\bx^{\prime})-(\bx\bx^{\top}-\bx^{\prime}\bx^{\prime\top})\btheta\|_2\\
			&\leq\max_{\bz,\bz^{\prime}}\big\{\|y\bx-y^{\prime}\bx^{\prime}\|_2+\|(\bx\bx^{\top}-\bx^{\prime}\bx^{\prime\top})\btheta\|_2\big\}\\
			&\leq 2C_y\cdot\sqrt{p}C_x+2pC_x^2\cdot C_0=2(C_y+\sqrt{p}C_xC_0)\sqrt{p}C_x.
		\end{split}
	\end{equation*}
	The sensitivity of the variance matrix $\widehat{\bSigma}$ satisfies
	\begin{equation*}
		\Delta_{\Sigma}=\max_{\bz,\bz^{\prime}}\bigg\|\frac{\bx\bx^{\top}}{n}-\frac{\bx^{\prime}\bx^{\prime\top}}{n}\bigg\|_F\leq\frac{2pC_x^2}{n}
	\end{equation*}
	and the sensitivity of the variance $\widehat{\sigma}^2$ satisfies
	\begin{equation*}
		\Delta_{\sigma}=\max_{\bz,\bz^{\prime}}\bigg\|\frac{(y-\bx^{\top}\btheta)^2}{n}-\frac{(y^{\prime}-\bx^{\prime\top}\btheta)^2}{n}\bigg\|_2\leq \frac{2(C_y+\sqrt{p}C_xC_0)^2}{n}.
	\end{equation*}
	
	\subsection{Logistic Regression}
	
	We use the negative log-likelihood function as the sample loss function,
	\begin{equation*}
		\begin{split}
			l(\bz_i;\btheta):&=-y_i\log[\{1+\exp(-\bx^{\top}\btheta)\}^{-1}]-(1-y_i)\log[1-\{1+\exp(-\bx^{\top}\btheta)\}^{-1}]\\
			&=-[y_i\bx_i^{\top}\btheta-\log\{1+\exp(\bx^{\top}\btheta)\}].
		\end{split}
	\end{equation*}
	The corresponding population function
	\begin{equation*}
		\begin{split}
			\mathcal{L}(\btheta)&=\int \sum_{y_i=0,1}l(\bx_i,y_i;\btheta)\Pr(y_i\mid\boldsymbol{x},\btheta^*)f(\bx)d\bx\\
			&=-\int\{1+\exp(-\bx^{\top}\btheta^*)\}^{-1}\bx^{\top}f(\bx)d\bx\cdot\btheta +\mathbb{E}[\log\{1+\exp(\bx^{\top}\btheta)\}].
		\end{split}
	\end{equation*}
	Take the second-order derivative,
	\begin{equation*}
		\nabla^2\mathcal{L}(\btheta)=\nabla^2 \mathbb{E}[\log\{1+\exp(\bx^{\top}\btheta)\}]=\mathbb{E}\bigg[\frac{\exp(\bx^{\top}\btheta)}{\{1+\exp(\bx^{\top}\btheta)\}^2}\bx\bx^{\top}\bigg]. 
	\end{equation*}
	Thus, under the uniform positive-curvature condition in the main paper, $\mathcal{L}(\btheta)$ is strongly convex on $\Theta$ with $\lambda=\inf_{\btheta\in\Theta}\lambda_{\min}(\nabla^2\mathcal{L}(\btheta))>0$.
	
	Then, we check Assumption 2. The innovation sequence satisfies
	\begin{equation*}
		\begin{split}
			\bxi(\btheta)=&\nabla \mathcal{L}(\btheta)-\nabla l(\bz_i;\btheta)\\
			=&-\int\{1+\exp(-\bx^{\top}\btheta^*)\}^{-1}\bx f(\bx)d\bx+\mathbb{E}[\{1+\exp(-\bx^{\top}\btheta)\}^{-1}\bx]\\
			&-\{-y_i\bx_i+ \{1+\exp(-\bx^{\top}_i\btheta)\}^{-1}\bx_i\}.
		\end{split}
	\end{equation*}
	The expectation satisfies $\mathbb{E}\bxi(\btheta)=\boldsymbol{0}$ by the fact that $\E(y_i\mid\boldsymbol{x}_i)=\Pr(y_i=1\mid\bx_i)=\{1+\exp(-\bx^{\top}\btheta^*)\}^{-1}$. Furthermore, we have
	\begin{equation*}
		\begin{split}
			\mathbb{E}\bxi(\btheta)\bxi(\btheta)^{\top}&=\mathbb{V}\{\bxi(\btheta)\}=\mathbb{V}[\E\{\bxi(\btheta)\mid\bx_i\}]+\mathbb{E}[\mathbb{V}\{\bxi(\btheta)\mid\bx_i\}]\\
			&=\mathbb{V}[\{1+\exp(-\bx^{\top}_i\btheta^*)\}^{-1}\bx_i-\{1+\exp(-\bx^{\top}_i\btheta)\}^{-1}\bx_i]+\mathbb{E}[\mathbb{V}\{y_i\bx_i\mid\bx_i\}]\\
			&=\mathbb{E}\bigg[\frac{\exp(\bx^{\top}\btheta^*)}{\{1+\exp(\bx_i^{\top}\btheta^*)\}^2}\bx_i\bx_i^{\top}\bigg]\\
			&+\mathbb{V}[\{\{1+\exp(-\bx^{\top}_i\btheta)\}^{-1}-\{1+\exp(-\bx^{\top}_i\btheta^*)\}^{-1}\}\bx_i\}].
		\end{split}
	\end{equation*}
	Thus, the innovation satisfies Assumption 2. For the strengthened part (i), the logistic loss is convex and continuously differentiable on $\mathbb R^p$ and $\|\nabla l(\bz;\btheta)\|\le\|\bx\|$, so one can take $B(\bz)=\|\bx\|$. Under the bounded-design condition $\|\bx\|_\infty\le C_x$, its Hessian is bounded and satisfies
	$\|\nabla^2l(\bz;\btheta)-\nabla^2l(\bz;\btheta^*)\|\le \|\bx\|^3\|\btheta-\btheta^*\|/4$; hence Assumption 5 also holds.

	We then consider the sensitivity parameters of the proposed method. Then, the $\ell_2$ sensitivities of the gradient satisfies
	\begin{equation*}
		\begin{split}
			\Delta_g&=\max_{\bz,\bz^{\prime}}\|\nabla l(\bz;\btheta)-\nabla l(\bz^{\prime};\btheta)\|_2\\
			&= \max_{\bz,\bz^{\prime}}\|\{-y\bx+ \{1+\exp(-\bx^{\top}\btheta)\}^{-1}\bx\}-\{-y^{\prime}\bx^{\prime}+ \{1+\exp(-\bx^{\prime\top}\btheta)\}^{-1}\bx^{\prime}\}\|_2\\
			&\leq\max_{\bz,\bz^{\prime}}(\|\bx\|_2+\|\bx^{\prime}\|_2)\leq2\sqrt{p}C_x,
		\end{split}
	\end{equation*}
	where we use the fact that $\{1+\exp(-\bx^{\top}\btheta)\}^{-1}\in(0,1)$. The sensitivity of the Hessian estimator $\widehat{\bA}$ satisfies
	\begin{equation*}
		\begin{split}
			\Delta_{\widehat A}&=\max_{\bz,\bz^{\prime}}\frac{1}{n}\bigg\|\frac{\exp(\bx^{\top}\btheta)}{\{1+\exp(\bx^{\top}\btheta)\}^2}\bx\bx^{\top}-\frac{\exp(\bx^{\prime\top}\btheta)}{\{1+\exp(\bx^{\prime\top}\btheta)\}^2}\bx^{\prime}\bx^{\prime\top}\bigg\|_F\\
			&\leq\max_{\bz,\bz^{\prime}}\frac{\|\bx\bx^{\top}\|_F+\|\bx^{\prime}\bx^{\prime\top}\|_F}{4n}\leq\frac{pC_x^2}{2n},
		\end{split}
	\end{equation*}
	where we use the fact $\exp(\bx^{\top}\btheta)\{1+\exp(\bx^{\top}\btheta)\}^{-2}\in(0,1/4]$ for any $\bx$ and $\btheta$.

	\section{Additional Numerical Results for SGD with Random Draws}
	
	In this section, we use some numerical studies to illustrate the performance of SGD with random draws.
	
	\subsection{Variance Inflation Resulting from Random Sampling (Theorem 1)}
	
	We conduct a simulation study to validate the theoretical findings presented in Theorem 1. As established in Theorem 1, both the number of iterations ($T=k\cdot n$) and the batch size ($m$) contribute nontrivially to the asymptotic variance of the SGD estimator. Specifically, for fixed values of $k$ and $m$, the distribution of the SGD estimator with randomly selected samples converges to a normal distribution, where the asymptotic variance is given by the variance of the full sample estimator multiplied by $\{1+(km)^{-1}\}$.
	We consider the linear regression models, where $\{x_i,y_i\}_{i=1}^{n}$ are i.i.d. samples with $x_i\sim N(0,1)$, $e_i\sim N(0,1)$ and $y_i=\theta\cdot x_i+e_i$. The regression coefficient $\theta^*=1$ is the true parameter of the model. We consider three different choices of batch sizes and a number of iterations across various sample sizes. The initial value is set to $\theta^{(0)}=0$ and the step size is set to $\eta_t=t^{-0.501}$, following the choice in \cite{chen2020statistical}. The procedures are replicated $5000$ times.
	
\begin{figure}[ht]
		\centering
		\includegraphics[width=\linewidth]{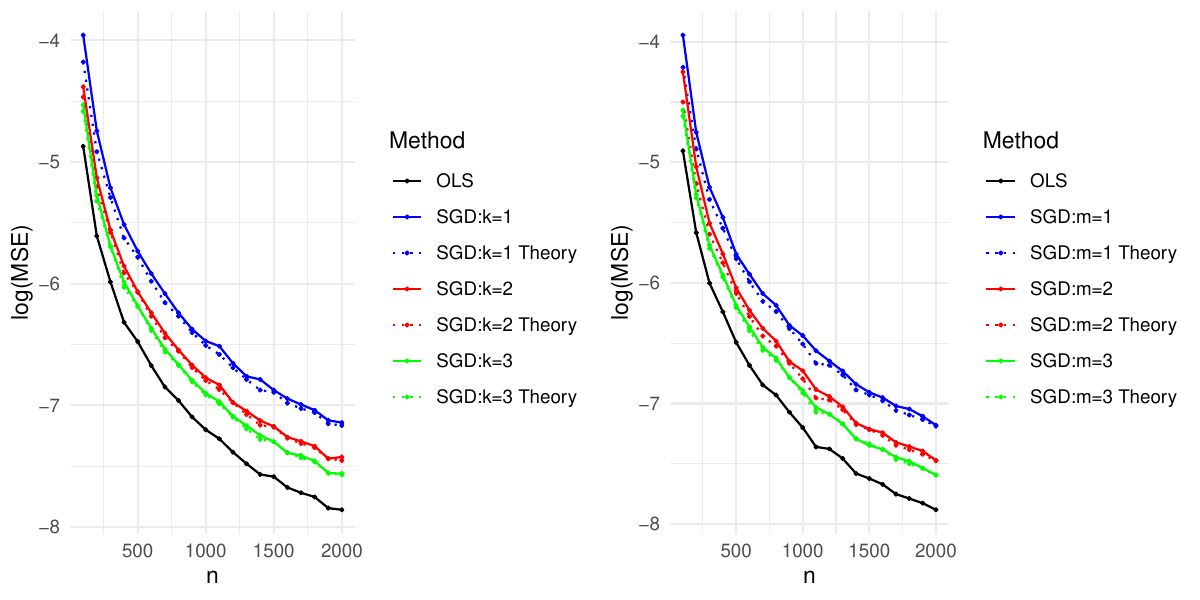}
		\caption{Logarithm of Empirical Mean Squared Error (MSE) for OLS and SGD with various iterations ($k$) and batch sizes ($m$), compared to theoretical results based on OLS under varying sample size ($n$).}
		\label{fig:thm1_example}
	\end{figure}

	The results are shown in Figure \ref{fig:thm1_example}, where we report the logarithm of the empirical mean squared error (MSE) for both the random SGD estimator and the ordinary least squares (OLS) estimator, with the latter serving as the benchmark full-sample estimator. Theoretical results, presented by dotted lines in Figure \ref{fig:thm1_example}, were computed by multiplying the empirical MSE of the OLS estimator by the factor $\{1+(km)^{-1}\}$. As shown in Figure \ref{fig:thm1_example}, the theoretical values align closely with the empirical results across all settings of batch sizes, iteration counts, and sample sizes.
	
	\subsection{Comparison between Plug-in and Random Scale (Theorem 4)}
	
	We perform a simulation study of the nonprivate analog of Theorem 4 for SGD with random draws. For $\sigma_1=0$ and $T/n\to\infty$, the corresponding limiting pivot is
	\begin{equation*}
		\sqrt{n}(\bar{\theta}_{T,j}-\theta^*_j)\widehat{V}_{j,j}^{-1/2}\stackrel{d}{\to}Z\cdot\bigg[\int_{0}^{1}\{W(r)-rW(1)\}^2\text{d}r\bigg]^{-1/2},
	\end{equation*}
	where $Z$ follows a standard normal distribution and is independent of the standard Wiener process $W(\cdot)$. For any $\alpha\in(0,1)$, let $c_{1-\alpha/2}$ be the $1-\alpha/2$ quantile of $Z\cdot[\int_{0}^{1}\{W(r)-rW(1)\}^2\text{d}r]^{-1/2}$. The proposed confidence interval is constructed by
	\begin{equation*}
		\text{CI}_{\alpha}(\theta_j):=\big(\bar{\theta}_{T,j}-c_{1-\alpha/2}(\widehat{V}_{j,j}/n)^{1/2},\bar{\theta}_{T,j}+c_{1-\alpha/2}(\widehat{V}_{j,j}/n)^{1/2}\big).
	\end{equation*}
	
	We are interested in comparing two inference methods: plug-in and random scale. We consider the linear regression models, where $\{\bx_i,y_i\}_{i=1}^{n}$ are i.i.d. samples with $\bx_i\sim N(\boldsymbol{0},\boldsymbol{I})$, $e_i\sim N(0,1)$ and $y_i=\btheta^{\top} \bx_i+e_i$. Each component of the true regression coefficient $\btheta$ is uniformly distributed in the interval $[0,1]$. We consider different choices of the number of iterations $T$. The initial value is set to $\theta^{(0)}=0$ and the step size is set to $\eta_t=t^{-0.501}$, following the choice in \cite{chen2020statistical}. The procedures are replicated $2000$ times. The performance is measured by two statistics: the coverage rate and the average length of the $95\%$ confidence interval.
	
	\begin{figure}[htbp]
		\centering
		% First subfigure
		\begin{subfigure}[b]{0.48\textwidth}
			\centering
			\includegraphics[width=\textwidth]{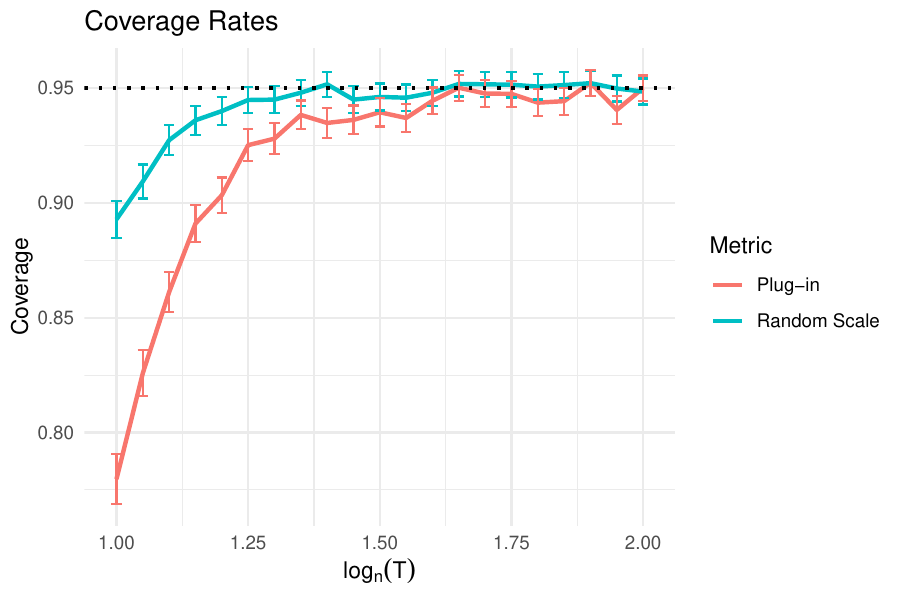}
		\end{subfigure}
		\hfill
		% Second subfigure
		\begin{subfigure}[b]{0.48\textwidth}
			\centering
			\includegraphics[width=\textwidth]{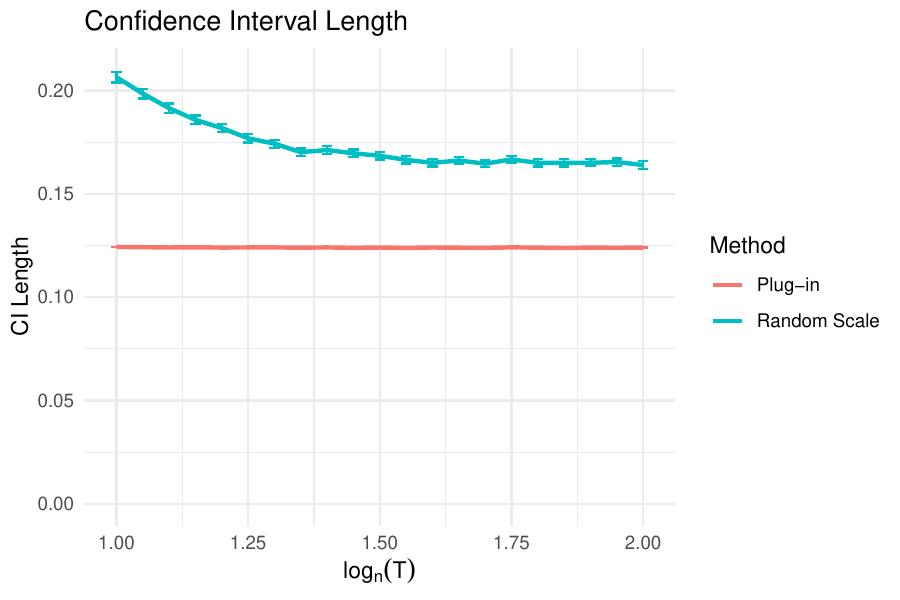}
		\end{subfigure}
		\caption{Empirical coverage and confidence interval lengths for the plug-in and random-scale methods across varying numbers of iterations $T$.}
		\label{fig:pi.vs.rs}
	\end{figure}
	
	The results are shown in Figure \ref{fig:pi.vs.rs}, where we report the empirical coverage and confidence interval lengths for the plug-in and random-scale methods across varying numbers of iterations $T$. The performance of the plug-in method is strongly related to the convergence of SGD with random draws, as shown in Figure~\ref{fig:thm1_example}. When the iterations $T$ satisfy $T\geq n^{1.5}$, the empirical coverage of the plug-in method approaches the nominal level of $95\%$. It is noteworthy that for sufficiently large $T$, the confidence interval produced by the plug-in method has approximately the same length as that of the full-sample estimator. The convergence behavior of the plug-in method can be attributed to the fact that when $T\geq n^{1.5}$, the RMSE of the SGD is very close to that of the full-sample estimator. Compared to the plug-in method, the random scale method achieves empirical coverage that is closer to the nominal level of $95\%$, even with a smaller number of iterations ($T\asymp n^{1.5}$). However, the average confidence interval length for the random scale method is slightly greater than that of the plug-in method. The numerical results demonstrate that the random scale method provides a fast and robust approach for inference, aligning with the findings of \cite{lee2022fast} regarding SGD with cyclic draws.

\end{document}